\pdfoutput=1
\documentclass[11pt]{article}
\usepackage{float}

\usepackage{acl}

\usepackage[utf8]{inputenc}
\usepackage{pgfplots}
\usepackage{dsfont}
\DeclareUnicodeCharacter{2212}{−}
\usepgfplotslibrary{groupplots,dateplot}
\usetikzlibrary{patterns,shapes.arrows}
\pgfplotsset{compat=newest}

\usepackage{tikzscale}
\usepackage{amsmath}
\newcommand{\probP}{\text{I\kern-0.15em P}}

\usepackage{multirow, colortbl}
\usepackage{color}
\usepackage{array}
\usepackage{multirow}
\usepackage{pifont}
\usepackage{tabularx,booktabs}
\usepackage{makecell}
\usepackage{graphicx}
\usepackage{subcaption}

\usepackage[normalem]{ulem}
\useunder{\uline}{\ul}{}

\usepackage{graphicx}

\usepackage{pifont}
\newcommand{\cmark}{\ding{51}} % Check mark
\definecolor{ablation6}{HTML}{fcefed}
\definecolor{ablation_tie}{HTML}{fce3e1}

\definecolor{ablation5}{HTML}{fcd8d4}
\definecolor{ablation4}{HTML}{FBC3BC}
\definecolor{ablation3}{HTML}{F7A399}
\definecolor{ablation2}{HTML}{F38375}
\definecolor{ablation1}{HTML}{EF6351}

\usepackage[]{algpseudocode}
\usepackage[]{algorithm}
\usepackage{float}
\algtext*{EndFor}%
\algtext*{EndProcedure}%

\usepackage{booktabs}
\usepackage[normalem]{ulem}
\useunder{\uline}{\ul}{}

\usepackage{times}
\usepackage{authblk}
\usepackage{latexsym}
\usepackage{adjustbox}
\usepackage{caption}

\usepackage[T1]{fontenc}
\usepackage{amsmath}
\usepackage{amssymb}
\usepackage{booktabs}
\usepackage{multirow}
\usepackage{scalerel,xparse}
\usepackage[utf8]{inputenc}
\usepackage{cleveref}
\usepackage{authblk}
\usepackage{microtype}
\usepackage{enumitem}
\crefformat{section}{\S#2#1#3}
\crefformat{subsection}{\S#2#1#3}
\crefformat{subsubsection}{\S#2#1#3}

\title{Mitigating Semantic Leakage in Cross-lingual Embeddings \\ via Orthogonality Constraint}

\author{\textbf{Dayeon Ki}$^{1*}$ \hspace{0.5cm} \textbf{Cheonbok Park}$^{2}$  \hspace{0.5cm} \textbf{Hyunjoong Kim}$^{2}$ \vspace{0.1cm}  \\
  $^{1}$University of Maryland \hspace{0.5cm}
  $^{2}$NAVER Cloud \hspace{0.5cm} \\
  \texttt{dayeonki@umd.edu}
}

\begin{document}
\maketitle

\begingroup
\renewcommand\thefootnote{}\footnote{*Work done during internship at NAVER Cloud.}
\addtocounter{footnote}{-1}
\endgroup

\begin{abstract}
Accurately aligning contextual representations in cross-lingual sentence embeddings is key for effective parallel data mining. A common strategy for achieving this alignment involves disentangling semantics and language in sentence embeddings derived from multilingual pre-trained models. However, we discover that current disentangled representation learning methods suffer from \textit{semantic leakage}—a term we introduce to describe when a substantial amount of language-specific information is unintentionally leaked into semantic representations. 
This hinders the effective disentanglement of semantic and language representations, making it difficult to retrieve embeddings that distinctively represent the meaning of the sentence.
To address this challenge, we propose a novel training objective, ORthogonAlity Constraint LEarning (\textsc{ORACLE}), tailored to enforce orthogonality between semantic and language embeddings. \textsc{ORACLE} builds upon two components: intra-class clustering and inter-class separation.
Through experiments on cross-lingual retrieval and semantic textual similarity tasks, we demonstrate that training with the \textsc{ORACLE} objective effectively reduces semantic leakage and enhances semantic alignment within the embedding space.\footnote{We release our code and dataset at \url{https://github.com/dayeonki/oracle}.}

% These components aim to distil language-agnostic semantic representations from sentence embeddings.
% aiming to effectively separate semantics and language representations learned by multilingual encoders. 

\end{abstract}

\section{Introduction}

Parallel datasets play a pivotal role in enhancing neural machine translation (NMT) performance \cite{michel-neubig-2018-mtnt}. However, acquiring high-quality parallel texts is challenging, especially for lower-resourced languages where monolingual data is more abundant \cite{niu-etal-2018-bi}. In this context, effective approaches for mining parallel data are essential for applying NMT in practical scenarios \cite{Artetxe_2019}.

% To obtain cross-lingual sentence embeddings, earlier studies have fine-tuned pre-trained multilingual language models with vast amounts of parallel datasets \cite{artetxe-schwenk-2019-massively, feng-etal-2022-language} or with natural language inference (NLI) datasets \cite{gao-etal-2021-simcse, wang-etal-2022-english}. These methods underscore the necessity of language agnostic representations for effective parallel mining.

\begin{figure}[htbp]
    \centering
    \begin{subfigure}[b]{0.22\textwidth}
        \centering
        \includegraphics[width=\textwidth]{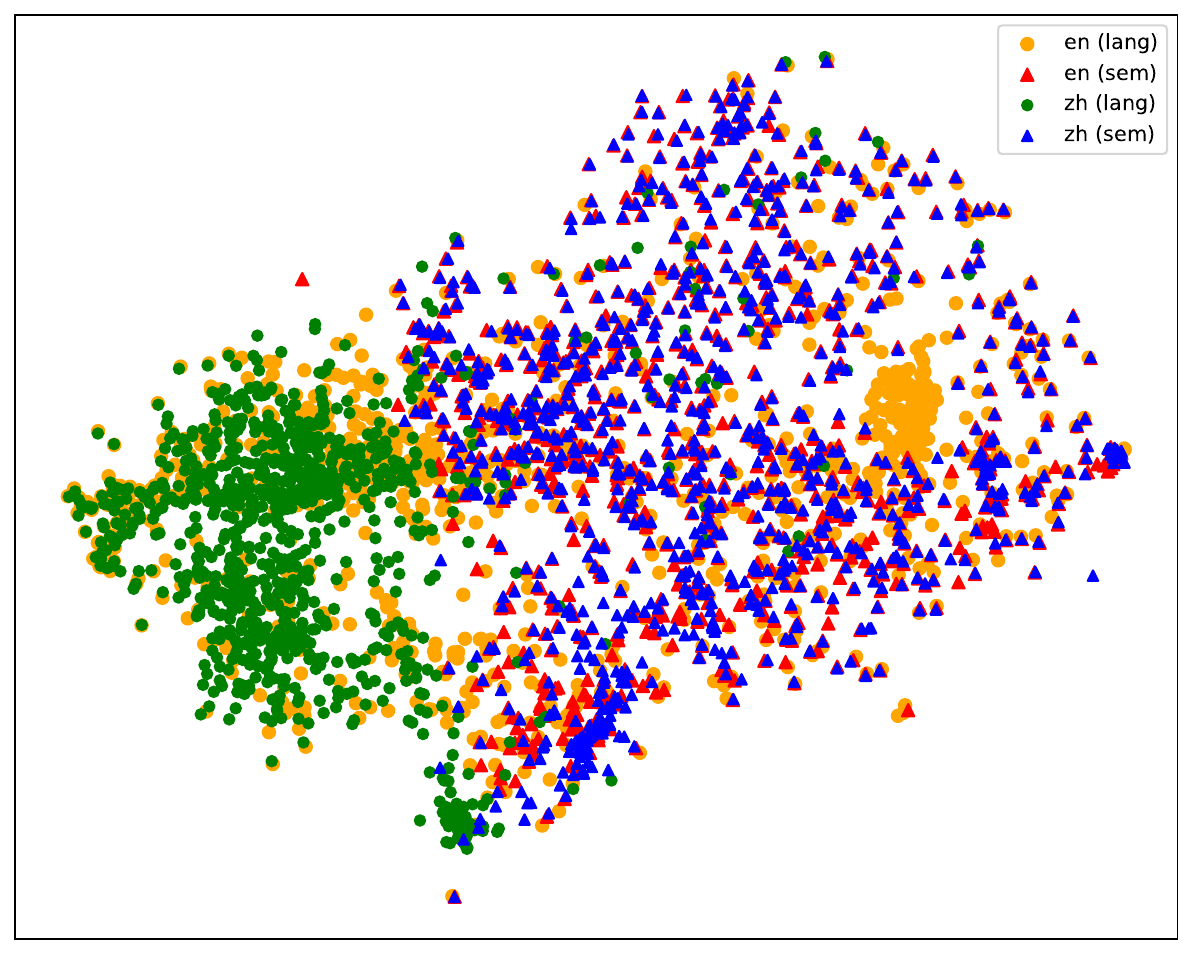}
        \caption{Semantic Leakage}
        \label{fig:embedding_space_a}
    \end{subfigure}
    \hfill % Optional: add some horizontal spacing
    \begin{subfigure}[b]{0.22\textwidth}
        \centering
        \includegraphics[width=\textwidth]{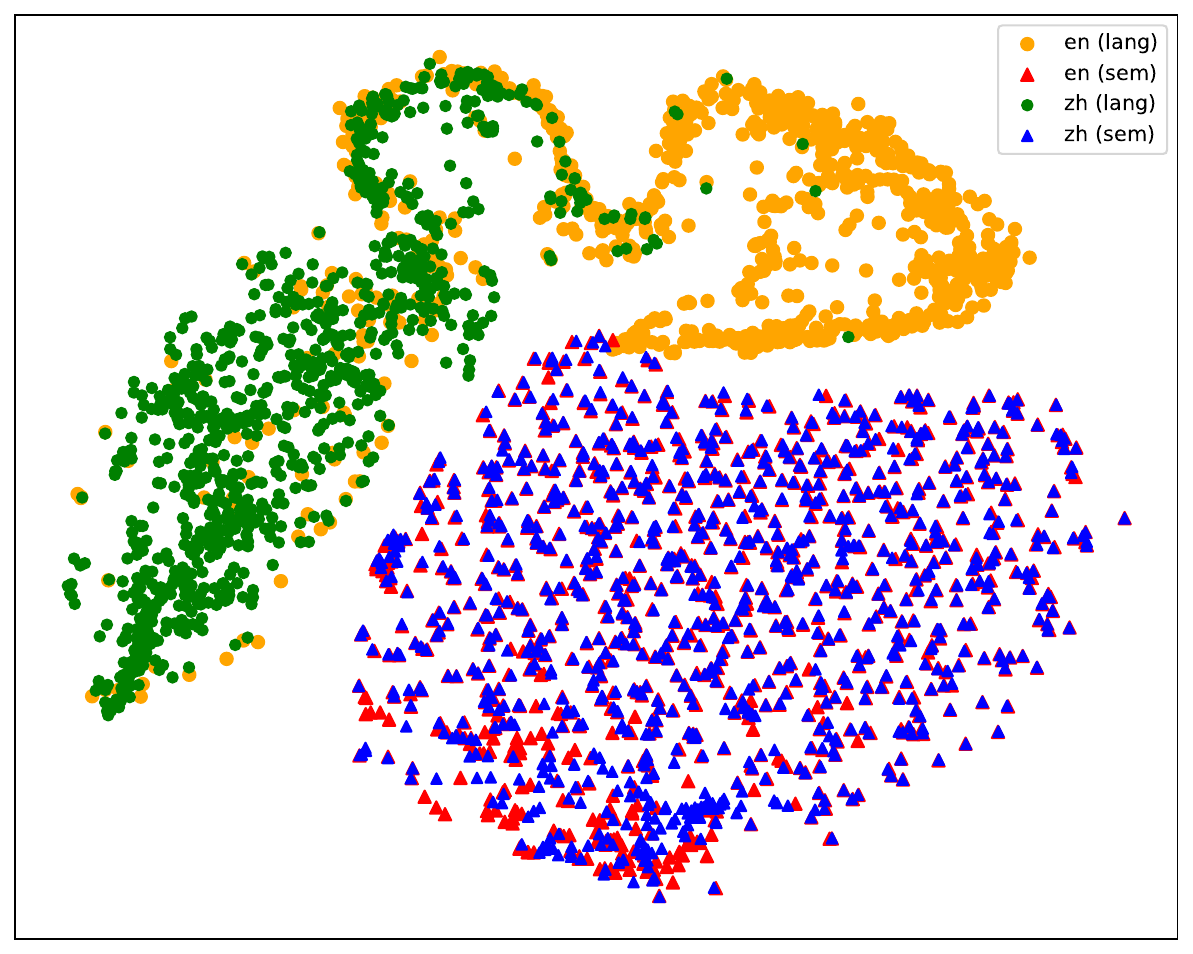}
        \caption{\textsc{ORACLE}}
        \label{fig:embedding_space_b}
    \end{subfigure}
    \caption{Visualization of LaBSE sentence embeddings for 1,000 Chinese-English sentence pairs. Figure 1(a) shows substantial overlap between semantic and language-specific representations. This overlap is effectively mitigated by the proposed \textsc{ORACLE} method, as shown in Figure 1(b).}
    \label{fig:embedding_space}
\end{figure}

Recent approaches to this problem utilize cross-lingual sentence embeddings \cite{schwenk-douze-2017-learning, schwenk-2018-filtering} generated by multilingual pre-trained encoders such as multilingual BERT (\citet{devlin-etal-2019-bert}, mBERT) or XLM-RoBERTa (\citet{conneau-etal-2020-unsupervised}, XLM-R). These embeddings aim to align semantically similar sentences across languages into a unified latent space, facilitating the extraction of pseudo-parallel pairs \cite{wang-etal-2022-english}. However, \citet{tiyajamorn-etal-2021-language} and \citet{kuroda-etal-2022-adversarial} demonstrate that embeddings of parallel sentences from these encoders form clusters by language rather than by semantics. Building on this, they attempt to disentangle language-specific information from sentence embeddings, thereby distilling language-agnostic semantic embeddings.

\begin{figure*}
    \centering
    \includegraphics[width=0.8\textwidth]{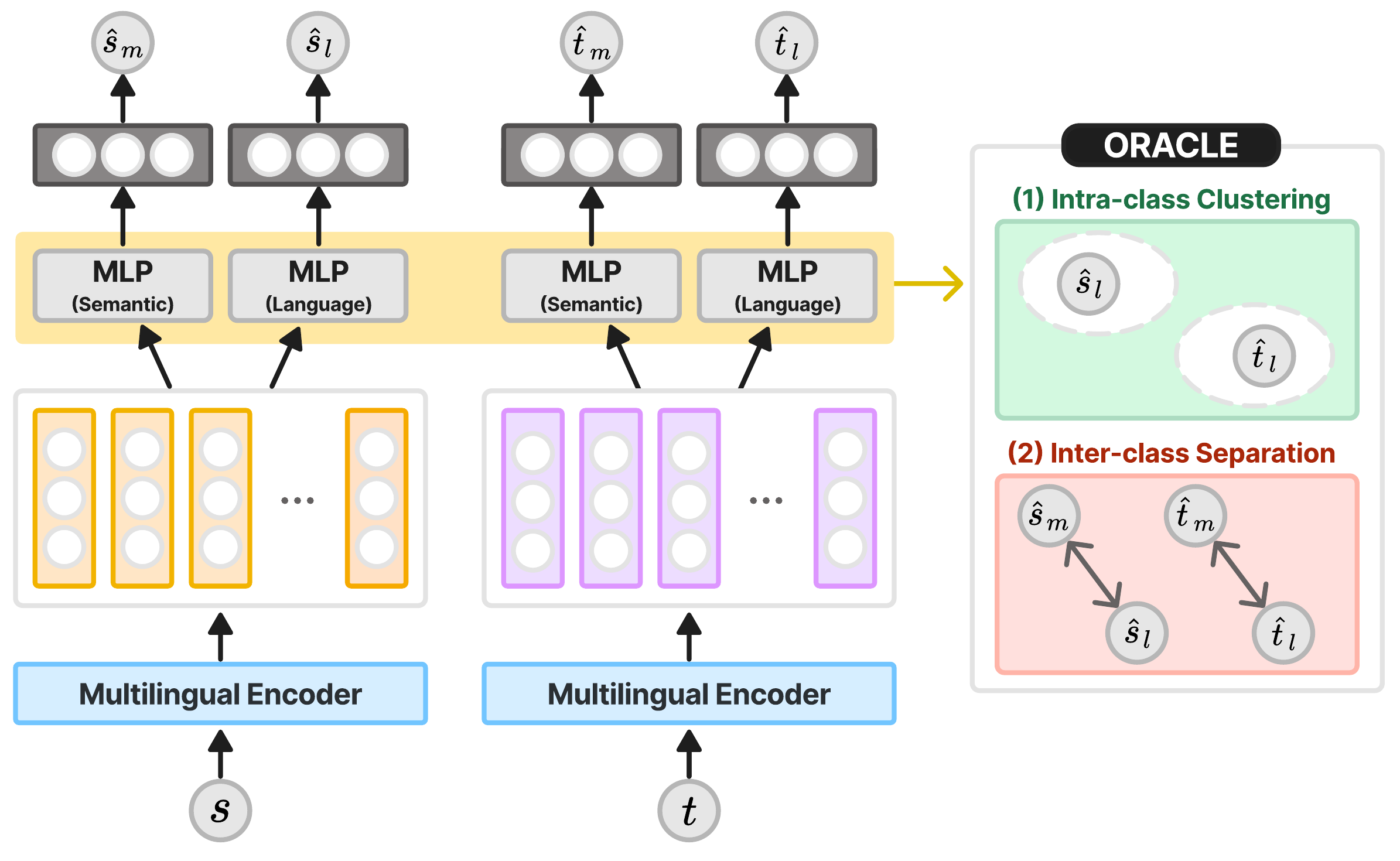}
    \caption{\textsc{ORACLE} objective for training semantic and language MLP networks. \textsc{ORACLE} is composed of two components: \textbf{(1) Intra-class clustering} for bringing related components closer in embedding space, \textbf{(2) Inter-class separation} for ensuring unrelated components to be distant. $s$ and $t$ represent source and target sentence input respectively. $\hat{s}_m$: source semantic representation; $\hat{s}_l$: source language representation; $\hat{t}_m$: target semantic representation; $\hat{t}_l$: target language representation.}
    \label{fig:method}
\end{figure*}

In order to achieve this, two premises need to be considered. Given parallel sentence, 
\begin{enumerate}[label=(\arabic*),topsep=0pt,itemsep=-1ex,partopsep=-1ex,parsep=1ex]
    \item How well are the semantic representations \textbf{aligned}?
    \item How well are the language-specific representations \textbf{separated}?
\end{enumerate}
Prior works have primarily focused on the former, leaving the latter question underexplored. Figure \ref{fig:embedding_space} illustrates sentence embeddings of a parallel corpus pair, indicating that while semantics are well-aligned with previous disentanglement methods, there is still substantial overlap between language-specific and semantic information (Figure \ref{fig:embedding_space_a}). We define this issue as \textbf{semantic leakage}, which undermines the effectiveness of cross-lingual embeddings in accurately mining parallel pairs. By constraining orthogonality between semantic and language representations, we facilitate a clearer separation of language-specific information in the embedding space (Figure \ref{fig:embedding_space_b}).
% to disperse across the embedding space.

% Achieving language-neutral sentence embeddings requires the disentanglement of pure semantics from language-specific features. Popular approaches involve tailored loss functions designed for this separation \cite{tiyajamorn-etal-2021-language, kuroda-etal-2022-adversarial}. However, these methods oftern suffer from semantic leaking, where a significant amount of language-specific information is leaked into the semantic embeddings, reducing their effectiveness in accurately mining pseudo-parallel pairs.

In this work, we introduce \textbf{\textsc{ORACLE}} (ORthogonality Constraint LEarning), a training objective aimed at enforcing orthogonality between semantic and language-specific representations. Our goal is to render these two representations independent to each other, thus ensuring their clear differentiation in the embedding space \cite{mitchell-steedman-2015-orthogonality}. \textsc{ORACLE} consists of two key components: intra-class clustering and inter-class separation. As shown in Figure \ref{fig:method}, intra-class clustering aligns related components more closely, while inter-class separation enforces orthogonality between unrelated components. Our method is designed to be simple and effective, capable of being implemented atop any disentanglement methods.

We explore a range of pre-trained multilingual encoders (LASER \cite{artetxe-schwenk-2019-massively}, InfoXLM \cite{chi-etal-2021-infoxlm}, and LaBSE \cite{feng-etal-2022-language}) to generate initial sentence embeddings. Subsequently, we train each semantic and language multi-layer perceptrons (MLPs) with \textsc{ORACLE} to disentangle the sentence embeddings into semantics and language-specific information. Experimental results on both cross-lingual sentence retrieval tasks \cite{artetxe-schwenk-2019-massively, zweigenbaum-etal-2017-overview} and the Semantic Textual Similarity (STS) task \cite{cer-etal-2017-semeval} demonstrate higher performance on semantic embeddings and lower performance on language embeddings with \textsc{ORACLE}. This suggests that our method not only resolves semantic leakage but also enhances semantic alignment (\S \ref{sec:experiment_results}). Our analysis further reveals that \textsc{ORACLE} leads to robust performance in challenging scenarios such as code-switching (\S \ref{sec:code_switching}).

To summarize, our contributions are threefold: \\
\noindent (1) We make the first attempt to address the issue of \textit{semantic leakage}, wherein a substantial amount of language-specific information is leaked into semantic representations.\\
\noindent (2) We mitigate semantic leakage with \textsc{ORACLE}, a simple and effective training objective that improves disentanglement of semantic and language-specific information. \\
\noindent (3) We show that \textsc{ORACLE} leads to robust mining in code-switched scenarios.

\section{Related work}
\subsection{Cross-lingual Sentence Embeddings}
Earlier works primarily centered on learning sentence-level representations for mining pseudo-parallel pairs. Initial methods utilized neural machine translation (NMT) systems with a shared encoder \cite{schwenk-douze-2017-learning, schwenk-2018-filtering}. This approach inspired supervised approaches which train neural networks with large parallel datasets. For instance, \citet{lee-chen-2017-muse} introduced the multilingual Universal Sentence Encoder (mUSE), a dual-encoder model pre-trained on parallel corpora in 16 languages. Similarly, LASER \cite{artetxe-schwenk-2019-massively} is an encoder-decoder model based on recurrent neural network. More recently, there has been a shift towards using multilingual sentence encoders such as mBERT \cite{devlin-etal-2019-bert}, XLM-R \cite{conneau-etal-2020-unsupervised}, and CRISS \cite{tran2020crosslingual}, which are based on single self-attention networks pre-trained on large monolingual datasets. InfoXLM \cite{chi-etal-2021-infoxlm} extends XLM-R by adding a cross-lingual contrastive pre-training objective to enhance cross-lingual understanding task performance. Subsequently, the Dual Encoder with Anchor Model (DuEAM) \cite{goswami-etal-2021-cross} incorporates a dual-encoder approach and integrates the word mover's distance to better capture semantic similarity between sentences. LaBSE \cite{feng-etal-2022-language} is a state-of-the-art multilingual sentence encoder built upon a dual-encoder framework, pre-trained with both monolingual and bilingual corpora. We leverage several of these multilingual sentence encoders to derive initial cross-lingual sentence embeddings. For our experiments, we specifically focus on three open-source baselines: LASER, InfoXLM, and LaBSE. We investigate the issue of semantic leakage in these encoders and effectively address it by integrating \textsc{ORACLE}.

\subsection{Disentangled Representation Learning}
A high-quality cross-lingual sentence embedding should effectively align semantically similar sentences from different languages in a shared embedding space \cite{wang-etal-2022-english}. However, embeddings obtained from multilingual sentence encoders are often highly biased by language-specific information \cite{tiyajamorn-etal-2021-language}. In this context, previous research has largely focused on learning disentangled representations to separate language-specific elements from semantics \cite{pires-etal-2019-multilingual, libovicky-etal-2020-language, gong2021lawdr, zhao-etal-2021-inducing}. One prevalent method involves training semantic and language networks separately, where the former is responsible for extracting meaning while the latter extracts language-specific information \cite{tiyajamorn-etal-2021-language, kuroda-etal-2022-adversarial, wu-etal-2022-learning}. Specifically, DREAM \cite{tiyajamorn-etal-2021-language} utilize a multi-task training approach with a combination of reconstruction, semantic embedding, and language embedding losses, while MEAT \cite{kuroda-etal-2022-adversarial} introduces novel loss combinations for more direct disentanglement. The distinct loss components of both methods are outlined in Table \ref{tab:dream_meat}.

Although disentangled representation learning has been explored previously, existing methods have primarily focused on aligning semantics. We discover that these approaches suffer from semantic leakage, as evidenced by the high performance of language-specific representations. Our work is the first to address this challenge through \textsc{ORACLE}, which enforces orthogonality between semantic and language representations.

\section{Background}

\begin{table}
\centering
\resizebox{\linewidth}{!}{%
    \begin{tabular}{l c c c c c}
    \toprule
    \textbf{Decomposer} & \textbf{$L_R$} & \textbf{$L_{CR}$} & \textbf{$L_S$} & \textbf{$L_L$} & \textbf{$L_A$}\\ \midrule
    DREAM \cite{tiyajamorn-etal-2021-language} & \cmark &  & \cmark & \cmark &  \\
    MEAT \cite{kuroda-etal-2022-adversarial} & \cmark & \cmark & & \cmark & \cmark \\ \bottomrule
    \end{tabular}
}
\caption{Comparison of loss components in DREAM and MEAT. $L_R$: Reconstruction loss, $L_{CR}$: Cross-Reconstruction loss, $L_S$: Semantic embedding loss, $L_L$: Language embedding loss, $L_A$: Adversarial loss.}
\label{tab:dream_meat}
\end{table}

\subsection{DREAM} 
\label{ref:subsection_dream}
DREAM \cite{tiyajamorn-etal-2021-language} employs two separate multi-layer perceptron (MLP) networks in an autoencoder setup to learn disentangled semantic and language-specific representations. Given a parallel corpus $C = \{(s^1, t^1), ..., (s^n, t^n)\}$, comprising pairs of sentences from a source and target language, each sentence pair $(s^i, t^i)$ is input into a multilingual pre-trained model (PLM). This generates original embeddings for the source $\mathbf{e}^i_s \in \mathbb{R}^d$ and the target sentences $\mathbf{e}^i_t \in \mathbb{R}^d$, where $d$ represents the dimension of the input sentence embeddings. Semantic and language representations are then extracted from these embeddings using a separate semantic MLP network $\mathrm{MLP}_{m}$ (denoted by $m$ to signify ``meaning'') and a language MLP network $\mathrm{MLP}_{l}$.
\begin{equation}
\mathbf{\hat{s}}^i_m = \mathrm{MLP}_m(\mathbf{e}^i_s)
\end{equation}
\begin{equation}
\mathbf{\hat{s}}^i_l = \mathrm{MLP}_l(\mathbf{e}^i_s)
\end{equation}
Here, $\mathbf{\hat{s}}^i_m$, $\mathbf{\hat{s}}^i_l \in \mathbb{R}^d$ represent the semantic and language representations of the source sentence, respectively, and similarly $\mathbf{\hat{t}}^i_m$, $\mathbf{\hat{t}}^i_l \in \mathbb{R}^d$ for the target sentence. We repeat this process across the entire parallel corpus $C$. 

For each language, the extracted semantic and language representations are element-wise summed to reconstruct the original sentence embedding as the final output. DREAM trains the two MLPs in a multi-task fashion, integrating three loss functions:
\begin{equation}
\mathcal{L}_\mathrm{DREAM} = \mathcal{L}_R + \mathcal{L}_S + \mathcal{L}_L
\end{equation}
where $\mathcal{L}_R$ is the reconstruction loss for reconstructing the original sentence embedding using semantic and language representations. $\mathcal{L}_S$ and $\mathcal{L}_L$ are responsible for extracting semantic and language information, respectively. Furthermore, $\mathcal{L}_L$ comprises both the language embedding loss ($\mathcal{L}^m_L$) and the language classification loss ($\mathcal{L}^i_L$), where $\mathcal{L}^m_L$ minimizes the distance within language embeddings and $\mathcal{L}^i_L$ computes the multi-class cross-entropy loss for the language classification task. 

\subsection{MEAT}
MEAT \cite{kuroda-etal-2022-adversarial} builds upon DREAM but incorporates more direct supervision to better disentangle semantic and language representations. MEAT trains the two MLPs with a new combination of four losses:
\begin{equation}
\mathcal{L}_\mathrm{MEAT} = \mathcal{L}_R + \mathcal{L}_{CR} + \mathcal{L}_L + \mathcal{L}_A
\end{equation}
$\mathcal{L}_{CR}$ is the cross-reconstruction loss for reconstructing the original source embedding using semantic from the target and language embedding from the source, and vice versa. $\mathcal{L}_A$ is the adversarial loss designed to reduce language identifiability in semantic representations.

\section{\textsc{ORACLE}}
The two key ingredients of \textsc{ORACLE} are intra-class clustering (\S \ref{subsec:intra}) and inter-class separation (\S \ref{subsec:inter}). We reformulate the losses originally derived in DREAM and MEAT and impose additional constraints to ensure orthogonality between semantic and language embeddings. 
Following the setup introduced in Section \ref{ref:subsection_dream}, \textsc{ORACLE} also uses semantic ($\mathrm{MLP}_m$) and language MLP ($\mathrm{MLP}_l$) to extract semantics ($\mathbf{\hat{s}}^i_m$, $\mathbf{\hat{t}}^i_m$) and language-specific information ($\mathbf{\hat{s}}^i_l$, $\mathbf{\hat{t}}^i_l$) for each language. 

\subsection{Intra-class clustering ($\mathcal{L}_{\mathrm{IC}}$)}
\label{subsec:intra}
$\mathcal{L}_{\mathrm{IC}}$ aims to bring relevant representations closer in the multilingual embedding space. As shown in Figure \ref{fig:embedding_space_a}, we notice that previous methods lack a constraint to enforce language embeddings to be clustered within themselves. This causes the language-specific information to leak into the semantics, making it difficult to capture the underlying semantics of the sentence. We constrain this by imposing pairwise cosine distances of each language embeddings:

% For DREAM, we consider $\mathcal{L}_R$ and $\mathcal{L}_S$ as part of $\mathcal{L}_{\mathrm{IC}}$ while for MEAT, we consider $\mathcal{L}_R$ and $\mathcal{L}_{CR}$. While previous works consider randomly paired sentences,

\begin{equation}
\mathcal{L}_{\mathrm{IC}} = \frac{1}{N} \sum_{i=1}^{N} \big( 2 - \phi(\mathbf{\hat{s}}^i_l, \mathbf{\hat{s}}^j_l) - \phi(\mathbf{\hat{t}}^i_l, \mathbf{\hat{t}}^j_l) \big)
\end{equation}
where $\phi(\cdot)$ denotes pairwise cosine similarity. $\phi(\mathbf{\hat{s}}^i_l, \mathbf{\hat{s}}^j_l)$ and $\phi(\mathbf{\hat{t}}^i_l, \mathbf{\hat{t}}^j_l)$ ($i \neq j$) measures the pairwise cosine similarity of language embeddings in source and target language respectively. We subtract from 2 to transition each of the similarity metric into distance metric. By minimizing $\mathcal{L}_{\mathrm{IC}}$, we aim to cluster language-specific representation for each language.

\subsection{Inter-class separation ($\mathcal{L}_{\mathrm{IS}}$)}
\label{subsec:inter}
Simultaneously, $\mathcal{L}_{\mathrm{IS}}$ enforces irrelevant representations to be clearly separated:

\begin{equation}
\small \mathcal{L}_{\mathrm{IS}} = \frac{1}{N} \sum_{i=1}^{N} \mathrm{max}(0, \phi(\mathbf{\hat{s}}^i_m, \mathbf{\hat{s}}^i_l)) + \mathrm{max}(0, \phi(\mathbf{\hat{t}}^i_m, \mathbf{\hat{t}}^i_l))
\end{equation}
where $\phi(\cdot)$ denotes cosine similarity. We impose a minimum value constraint of 0 to ensure the proper enforcement of orthogonality, indicative of unrelatedness, between the source and target language embeddings. Minimizing $\mathcal{L}_{\mathrm{IS}}$ effectively disentangles semantics from language-specific representations by constraining them to be orthogonal in the embedding space.

Combining with the intra-class clustering objective we get the final loss as:
\begin{equation}
\mathcal{L}_{\mathrm{ORACLE}} = \mathcal{L}_{\mathrm{IC}} + \mathcal{L}_{\mathrm{IS}}.
\end{equation}
We train both MLP networks, $\mathrm{MLP}_m$ and $\mathrm{MLP}_l$, with the combined loss $\mathcal{L}_{\mathrm{ORACLE}}$ in a multi-task learning approach. We integrate $\mathcal{L}_{\mathrm{ORACLE}}$ with the existing loss functions of DREAM or MEAT. This is based on our experiments in Section \ref{sec:oracle_components} where integrating \textsc{ORACLE} with DREAM or MEAT yields better performance than using it as a stand-alone objective.

\section{Experimental setup}

\subsection{Data}
We compile a dataset comprising 12 language pairs sourced from publicly available bilingual corpora\footnote{Our training corpus is obtained from OPUS (\url{https://opus.nlpl.eu/}). Details regarding the training corpus for each language pair are outlined in Appendix \ref{sec:train_corpus}.}. English is chosen as the source language for all pairs. We randomly sample 0.5M sentences for each language pair, which is later split into 0.45M for training and 0.05M for testing. In total, we utilize 6M parallel sentences. We select the language pairs based on diversity in language families, semantic similarity to English, and resource availability. Additional details for each language pair are provided in Table \ref{tab:languages}.

\subsection{Baselines}
Our study encompasses three open-source multilingual sentence encoders to generate initial sentence embeddings:

\begin{itemize}[leftmargin=*, itemsep=7pt, parsep=0pt]
 % \item \textbf{mBERT}: Multilingual masked language model trained on monolingual data from 104 languages \cite{devlin2019bert}.
% 
  \item \textbf{LASER}: Multilingual enc-decoder model trained on 93 languages \cite{artetxe-schwenk-2019-massively}.
 \item \textbf{InfoXLM}: XLM-R \cite{conneau-etal-2020-unsupervised} trained with masked language modeling (MLM), translation language modeling (TLM), and cross-lingual contrastive learning task with monolingual and parallel corpora \cite{chi-etal-2021-infoxlm}.
  \item \textbf{LaBSE}: A dual-encoder framework trained with MLM and TLM on both monolingual and bilingual corpora \cite{feng-etal-2022-language}.

\end{itemize}

Each multilingual sentence encoder is pre-trained with different combinations of languages. Consequently, the list of seen and unseen languages from our training data varies for each encoder, as summarized in Appendix Table \ref{tab:seen_unseen}.

\subsection{Implementation Details}
We train the two MLP layers—a semantic embedding layer and a language embedding layer—to distill semantic and language-specific features while keeping the backbone sentence encoder frozen. The output embedding of the [CLS] token is used for sentence embedding. Further details on training process is detailed in Appendix \ref{appendix:training}.

\begin{table}
\centering
\resizebox{\linewidth}{!}{%
    \begin{tabular}{l c c c c}
    \toprule
    \textbf{Language} & \textbf{Family} & \textbf{ISO Code} & \textbf{Similarity} & \textbf{Resource level}\\ \midrule
    \textbf{English} & Germanic & en & - & high \\
    \textbf{German} & Germanic & de & 0.81 & high \\ 
    \textbf{Portuguese} & Romance & pt & 0.84 & high \\ 
    \textbf{Italian} & Romance & it & 0.85 & high \\ 
    \textbf{Spanish} & Romance & es & 0.86 & high \\ 
    \textbf{French} & Romance & fr & 0.86 & high \\ 
    \textbf{Chinese} & Sino-Tibetan & zh & 0.81 & high \\ 
    \textbf{Arabic} & Semitic & ar & 0.91 & high \\ 
    \textbf{Japanese} & Japonic & ja & 0.69 & high \\ 
    \textbf{Dutch} & Germanic & nl & 0.80 & medium \\ 
    \textbf{Romanian} & Romance & ro & 0.88 & medium \\ 
    \textbf{Guaraní} & Tupi-Guaraní & gn & 0.25 & low \\ 
    \textbf{Aymara} & Andean & ay & 0.18 & low \\ 
    \bottomrule
    \end{tabular}
}
\caption{Summary of 12 languages used for training. Similarity refers to the cosine similarity between 1,000 sample of English and target language sentences measured using LaBSE embeddings.}
\label{tab:languages}
\end{table}

\begin{figure*}
    \centering
    \includegraphics[width=\textwidth]{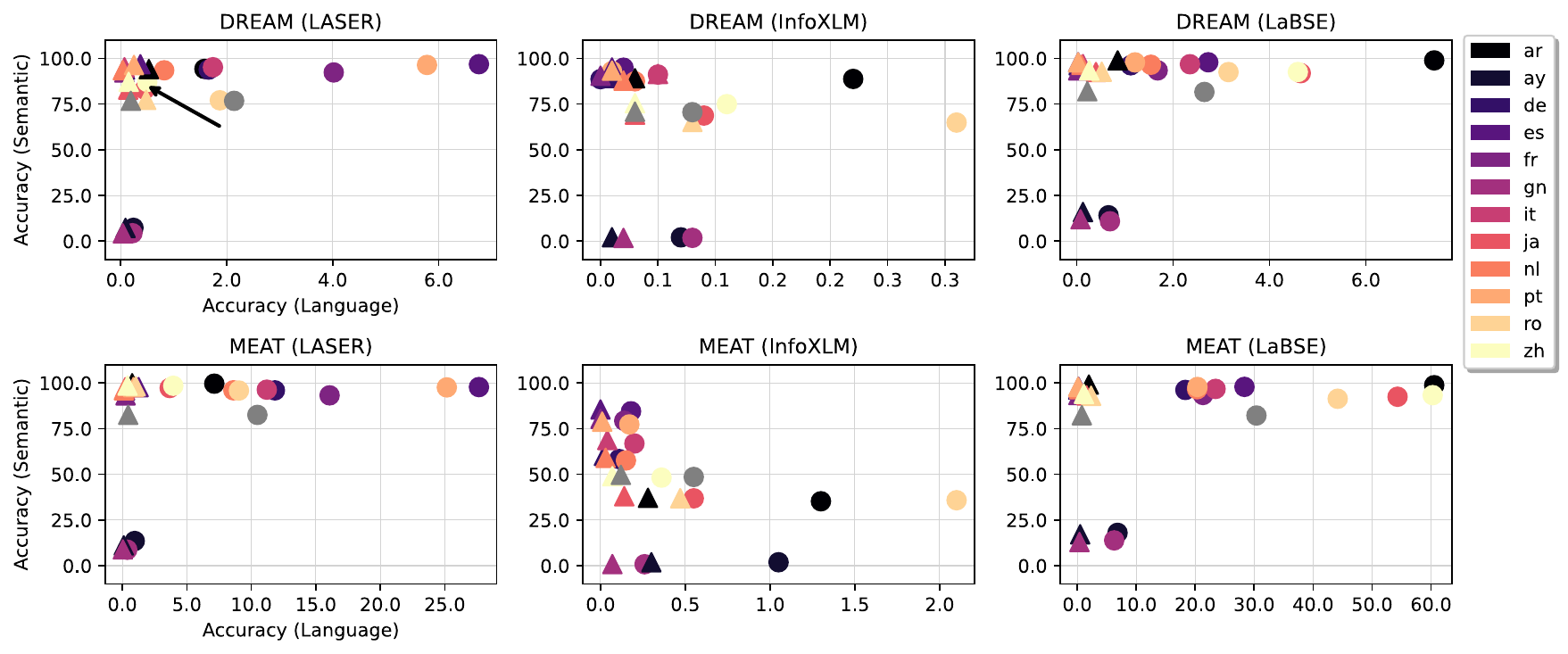}
    \caption{Cross-lingual sentence retrieval performance using our test set, consisting of 0.5M pairs for each language. The optimal representations exhibit high semantic retrieval accuracy and low language embedding retrieval accuracy, aiming for the upper left corner of each plot (indicated by the arrow). \ding{108}: vanilla DREAM or MEAT; \ding{115}: with \textsc{ORACLE} objective. \textcolor{gray}{Grey}: Average accuracy across 12 language pairs. \textit{Top row}: DREAM with each multilingual encoder baselines; \textit{Bottom row}: MEAT with multilingual encoders. Numerical results are in Appendix \ref{sec:cross_lingual_detailed}.}
    \label{fig:cross_retrieval}
\end{figure*}

% , and BUCC \cite{zweigenbaum-etal-2017-overview}.
\subsection{Evaluation task}
\paragraph{Cross-lingual Sentence Retrieval.} 
We evaluate our model on two distinct cross-lingual sentence retrieval tasks: held-out test set and Tatoeba\footnote{\url{https://tatoeba.org/}} \cite{artetxe-schwenk-2019-massively}. Given a list of bilingual sentences, the cross-lingual sentence retrieval task aims to accurately pair sentences that are in a parallel relationship across languages. The dataset consists of up to 1,000 sentences per language along with their English translations. We follow the evaluation setup proposed by \citet{wang-etal-2022-english}, evaluating accuracy in both Tatoeba-14 and Tatoeba-36 settings, each covering 14 languages from LASER and 36 languages from the XTREME benchmark \cite{hu2020xtreme}. We measure retrieval accuracy using both semantic and language-specific representations. Lower language embedding retrieval results suggest reduced semantic leakage in these representations, while higher semantic retrieval accuracy indicates improved semantic alignment in bilingual sentence pairs. 

% BUCC is a bitext mining task where the model is tasked to rank all possible sentence pairs and predict sentences with scores higher than the threshold. We report F1-scores for semantic and language-specific representations across 4 language pairs.

\paragraph{Semantic Textual Similarity.} 
We also report performance on the SemEval-2017 Semantic Textual Similarity (STS) task \cite{cer-etal-2017-semeval}. This task involves 7 cross-lingual and 3 monolingual sentence pairs. We aim to achieve high Spearman's rank correlation coefficients ($\rho$) with semantic representations, indicating better semantic alignment, while expecting lower coefficients with language representations, indicating effective separation.

\section{Results}
\label{sec:experiment_results}

\subsection{Cross-lingual Sentence Retrieval}
\paragraph{Held-out Test Set.}
Figure \ref{fig:cross_retrieval} illustrates the performance of cross-lingual sentence retrieval on our held-out test set, consisting of 0.5M parallel sentences per language pair. We assess retrieval accuracy using semantic and language-specific representations of these parallel sentences. The optimal representation entails high semantic accuracy and low language embedding accuracy. Notably, applying \textsc{ORACLE} shifts performance towards the upper left quadrant, indicative of higher semantic accuracy and reduced language embedding accuracy across all encoder baselines. We report detailed numerical results in Appendix Table \ref{tab:test_set}.

%Although the differences in average semantic performance between vanilla DREAM or MEAT and \textsc{ORACLE} may be small, they are statistically significant ($p \leq 0.1$) for all cases. 
% The average gain across language pairs is +

\paragraph{Tateoba.}
We draw similar conclusions from another cross-lingual retrieval task, Tatoeba, as shown in Table \ref{tab:tatoeba}. Utilizing disentangled representations with \textsc{ORACLE} generally yields superior performance compared to representations learned by existing methods such as DREAM and MEAT. One exception is DREAM with LaBSE sentence embeddings, for which the accuracy drops by 0.06 points after integrating \textsc{ORACLE}. 

Furthermore, we observe that models exhibit stronger performance from English (\textsc{En-XX}) than into English (\textsc{XX-En}) directions. Specifically, for Tatoeba-14, the semantic accuracy difference between the two settings of the vanilla model is smallest for LaBSE at 0.14 points, 0.69 points for LASER, and 15.6 points for InfoXLM on average. We notice a similar trend with the application of \textsc{ORACLE}, with the smallest difference for LaBSE at 0.08 points, 0.22 points for LASER, and 15.78 points for InfoXLM on average. We attribute this to \textsc{En-XX} setting being similar to our training corpus. We present comprehensive results on Tatoeba in Appendix \ref{sec:tatoeba_detailed}.

\definecolor{salmon}{rgb}{0.9609, 0.5664, 0.3086}
\definecolor{light_blue}{rgb}{0.3516, 0.7031, 0.7734}
\definecolor{green}{rgb}{0.3711, 0.7734, 0.3516}

\begin{figure}[!htp]
    \centering
    \includegraphics[width=\linewidth]{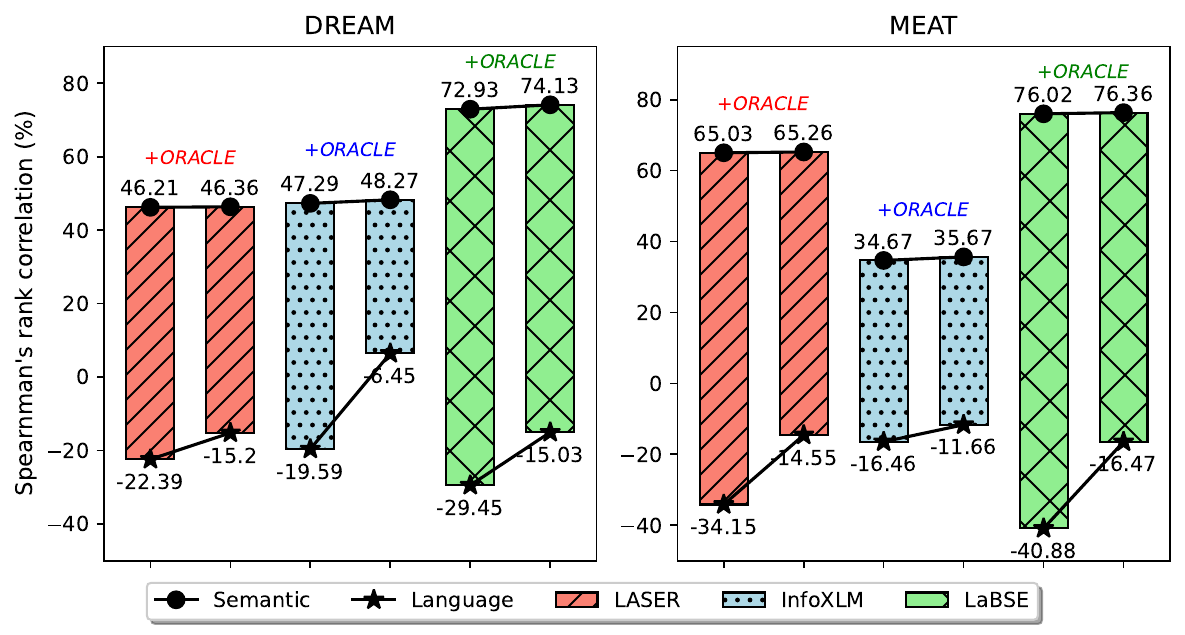}
    \caption{Spearman's rank correlation (\%) from the STS task for each multilingual encoder baseline. Length of the bars reflects the performance gap between semantic (\ding{108}) and language-specific (\ding{72}) representations. Each set of three bars displays results for \textcolor{salmon}{LASER}, \textcolor{light_blue}{InfoXLM}, and \textcolor{green}{LaBSE}. Within each color set, the first bar represents the vanilla approach, and the second bar denotes the integration of \textsc{ORACLE} objective.}
    \label{fig:sts}
\end{figure}

\paragraph{Seen vs. Unseen.}
Each multilingual encoder unsurprisingly show lower performance for their unseen target languages, as indicated in Table \ref{tab:seen_unseen}. One exception is the performance of LASER embeddings on Aymara (ay), which shows low performance despite being a seen language. Additionally, we note that \textsc{ORACLE} has a greater impact on the semantic embedding accuracy of unseen languages compared to seen languages. When training with \textsc{ORACLE}, the average semantic accuracy of the seen languages increases from 83.32→83.33 for LASER, 84.29→84.63 for InfoXLM, and 95.43→95.61 for LaBSE. The gap is more significant for unseen languages, increasing from 8.73→8.91 for LASER, 1.93→2.43 for InfoXLM, and 12.51→13.96 for LaBSE. This trend suggests that \textsc{ORACLE} helps bridge the performance gap between seen and unseen languages.

\begin{table}
\centering
\resizebox{\linewidth}{!}{%
    \begin{tabular}{l l c c c c c c c c c c c c c c c c c}
    \toprule
    \multirow{2}{*}{\textbf{Encoder}} & \multirow{2}{*}{\textbf{Objective}} & \multicolumn{2}{c}{\textbf{Tatoeba-14}} & \multicolumn{2}{c}{\textbf{Tatoeba-36}} \\
    \cmidrule(lr){3-4} \cmidrule(lr){5-6}
    & & (\textsc{En-XX}) & (\textsc{XX-En}) & (\textsc{En-XX}) & (\textsc{XX-En}) \\ \midrule\midrule

     \multicolumn{6}{c}{\textit{Semantic Embedding} (↑)} \\ \midrule

     % \multirow{4}{*}{mBERT} & DREAM & 20.21 & 24.31 & 18.31 & 22.04 \\
     %   & +\textsc{ORACLE} & \textbf{20.23} & \textbf{24.35} & 18.14 & \textbf{22.07} \\
     %   & MEAT & 13.38 & 14.84 & 11.78 & 11.74 \\
     %  &  +\textsc{ORACLE} & \textbf{14.18} & \textbf{15.35} & \textbf{12.52} & \textbf{12.23} \\ \midrule      
      
      \multirow{4}{*}{LASER} & DREAM & 68.68 & 69.53 & 59.94 & 62.01 \\
       & +\textsc{ORACLE} & \textbf{68.82} & \textbf{69.66} & \textbf{60.14} & \textbf{62.11} \\
       & MEAT & 88.48 & 89.00 & 80.56 & 79.26 \\
      & +\textsc{ORACLE} & 88.30 & 87.70 & \textbf{81.06} & \textbf{79.27} \\ \midrule
 
        \multirow{4}{*}{InfoXLM} & DREAM & 42.20 & 51.40 & 39.51 & 47.10 \\
       & +\textsc{ORACLE} & \textbf{42.35} & \textbf{51.87} & \textbf{39.73} & \textbf{47.71} \\
      & MEAT & 31.50 & 53.49 & 28.21 & 44.53 \\
      &  +\textsc{ORACLE} & \textbf{32.79} & \textbf{54.83} & \textbf{29.53} & \textbf{45.63} \\ \midrule

       \multirow{4}{*}{LaBSE} & DREAM & 95.57 & 95.76 & 95.27 & 95.09 \\
       & +\textsc{ORACLE} & \textbf{95.69} & 95.75 & 95.26 & 95.03 \\
      & MEAT & 95.67 & 95.76 & 95.33 & 95.06 \\
      &  +\textsc{ORACLE} & \textbf{96.06} & \textbf{96.16} & \textbf{95.58} & \textbf{95.48} \\ \midrule\midrule

      \multicolumn{6}{c}{\textit{Language Embedding} (↓)} \\ \midrule

      % \multirow{4}{*}{mBERT} & DREAM & 0.19 & 0.17 & 0.20 & 0.16 \\
      %  & +\textsc{ORACLE} & \textbf{0.12} & \textbf{0.12} & \textbf{0.15} & \textbf{0.14} \\
      %  & MEAT & 0.17 & 0.39 & 0.21 & 0.33 \\
      % &  +\textsc{ORACLE} & \textbf{0.13} & \textbf{0.16} & \textbf{0.17} & \textbf{0.18} \\ \midrule

      \multirow{4}{*}{LASER} & DREAM & 1.58 & 1.35 & 1.44 & 1.21 \\
       & +\textsc{ORACLE} & \textbf{0.17} & \textbf{0.27} & \textbf{0.20} & \textbf{0.26} \\
       & MEAT & 12.52 & 10.93 & 10.12 & 7.86 \\
      &  +\textsc{ORACLE} & \textbf{0.34} & \textbf{0.36} & \textbf{0.37} & \textbf{0.41} \\ \midrule

       \multirow{4}{*}{InfoXLM} & DREAM & 0.31 & 0.27 & 0.35 & 0.37 \\
       & +\textsc{ORACLE} & \textbf{0.12} & \textbf{0.12} & \textbf{0.14} & \textbf{0.14} \\
       & MEAT & 0.33 & 1.92 & 0.36 & 2.32 \\
      & +\textsc{ORACLE} & \textbf{0.14} & \textbf{0.18} & \textbf{0.17} & \textbf{0.20} \\ \midrule

      \multirow{4}{*}{LaBSE} & DREAM & 18.39 & 18.09 & 19.33 & 19.58 \\
       & +\textsc{ORACLE} & \textbf{1.26} & \textbf{1.36} & \textbf{1.50} & \textbf{1.70} \\
       & MEAT & 87.35 & 36.66 & 86.51 & 40.61 \\
      & +\textsc{ORACLE} & \textbf{8.48} & \textbf{7.00} & \textbf{9.92} & \textbf{8.41} \\

    \bottomrule
    \end{tabular}
}
\caption{Cross-lingual retrieval accuracy with Tatoeba dataset. We report the accuracy in both directions (from English and into English). \textbf{Bold} denotes better performance than the vanilla approach. All improvements are statistically significant with $p$-value $\leq 0.001$.}
\label{tab:tatoeba}
\end{table}

\subsection{Semantic Textual Similarity}
In Figure \ref{fig:sts}, we present the average Spearman's rank correlation coefficient across 10 STS tasks. The lengths of the bars indicate the performance gap between semantic and language-specific representations. With \textsc{ORACLE}, we observe a stronger positive correlation with STS scores for semantics and a stronger negative correlation for language representations. The extent of improvement in semantic results differs depending on both the encoder and the objective loss function. For DREAM, the highest gain is observed for LaBSE as +1.2 and the lowest for LASER as +0.15. Conversely, for MEAT, the highest gain is observed for InfoXLM as +1.0 and the lowest for LASER as +0.23.

\definecolor{dark green}{rgb}{0.273, 0.508, 0.082}

\begin{figure*}[]
\centering

\begin{subfigure}{.22\textwidth}
  \centering
  \includegraphics[width=\linewidth]{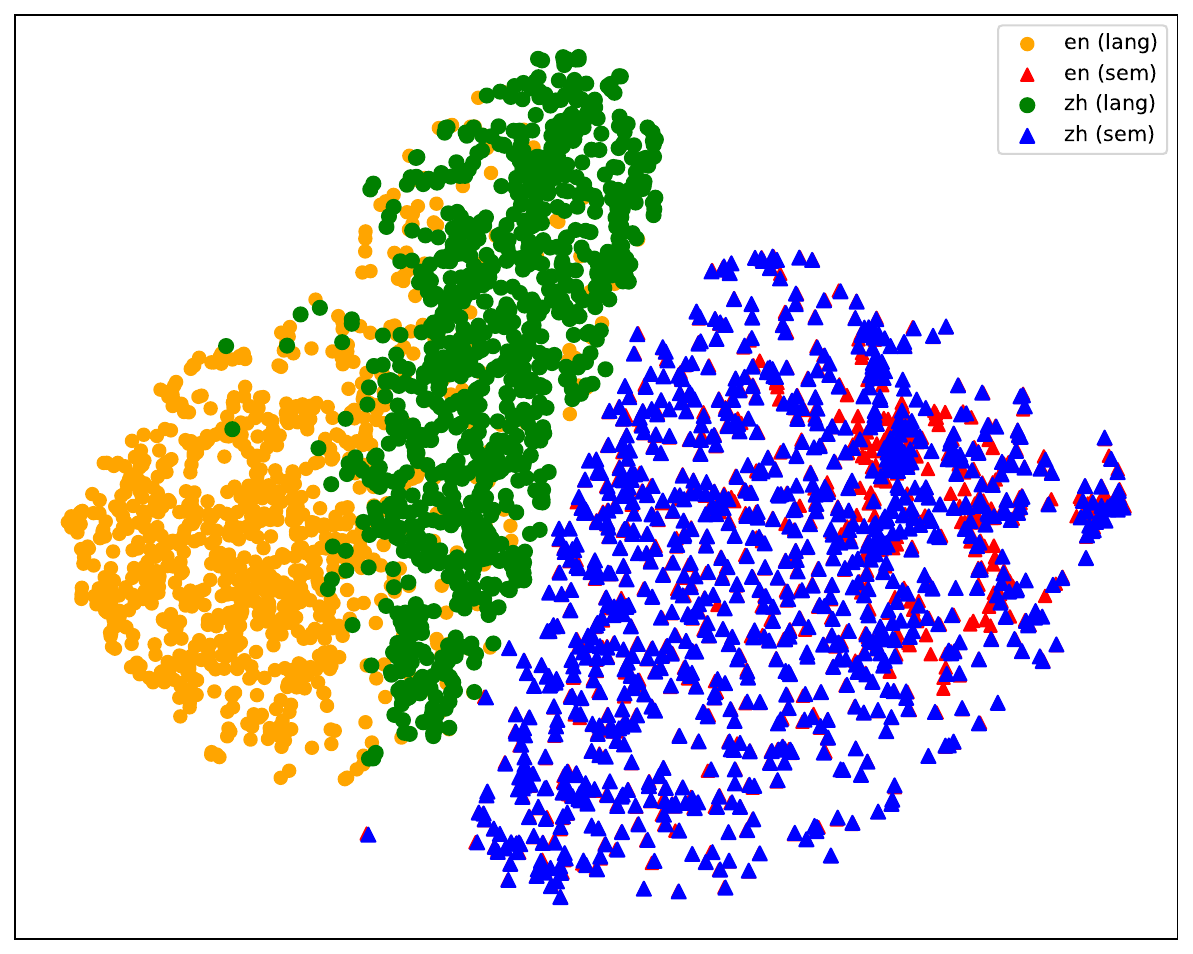}
  \caption{DREAM}
  \label{fig:enzh_dream}
\end{subfigure}%
\begin{subfigure}{.22\textwidth}
  \centering
  \includegraphics[width=\linewidth]{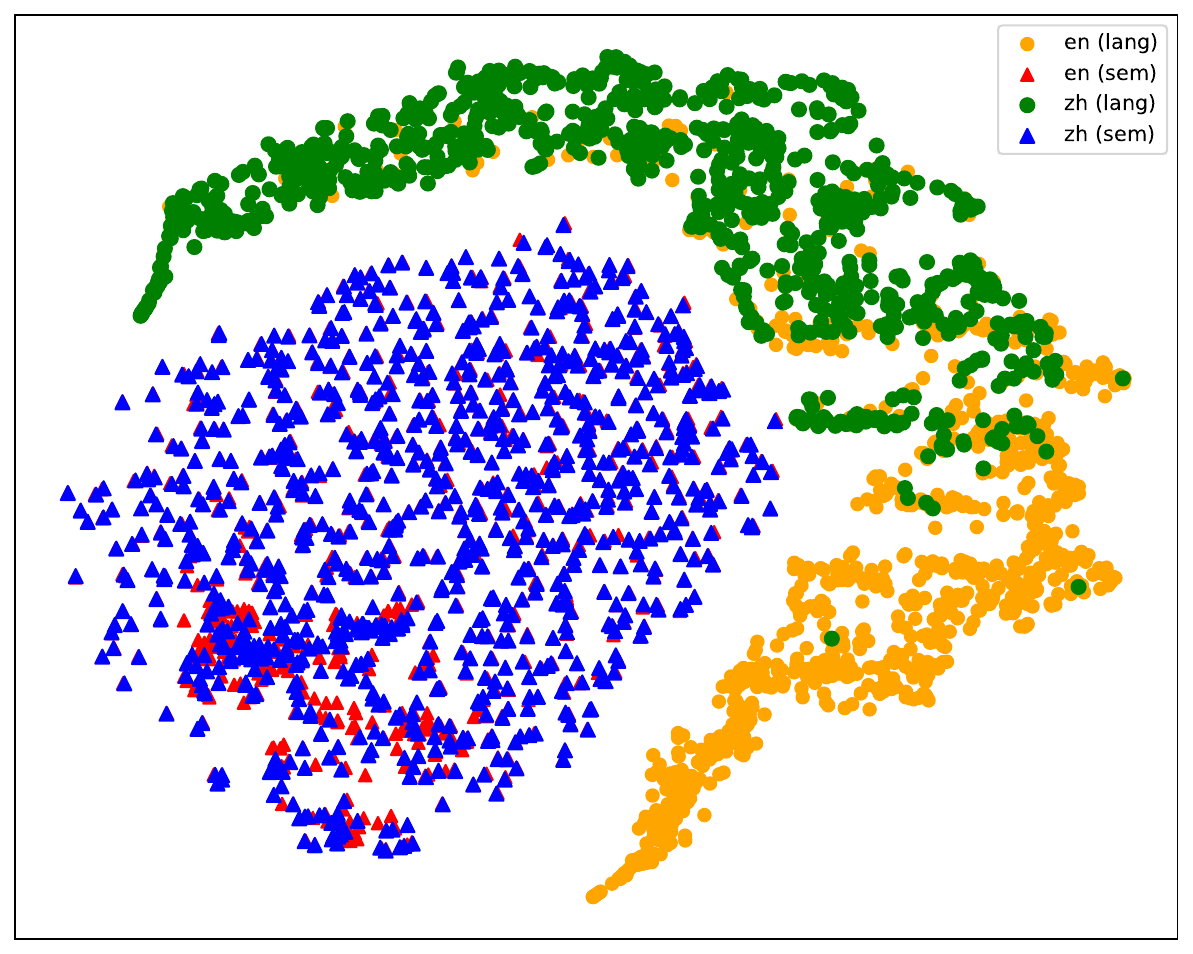}
  \caption{DREAM + \textsc{ORACLE}}
  \label{fig:enzh_dream_oracle}
\end{subfigure}
\begin{subfigure}{.22\textwidth}
  \centering
  \includegraphics[width=\linewidth]{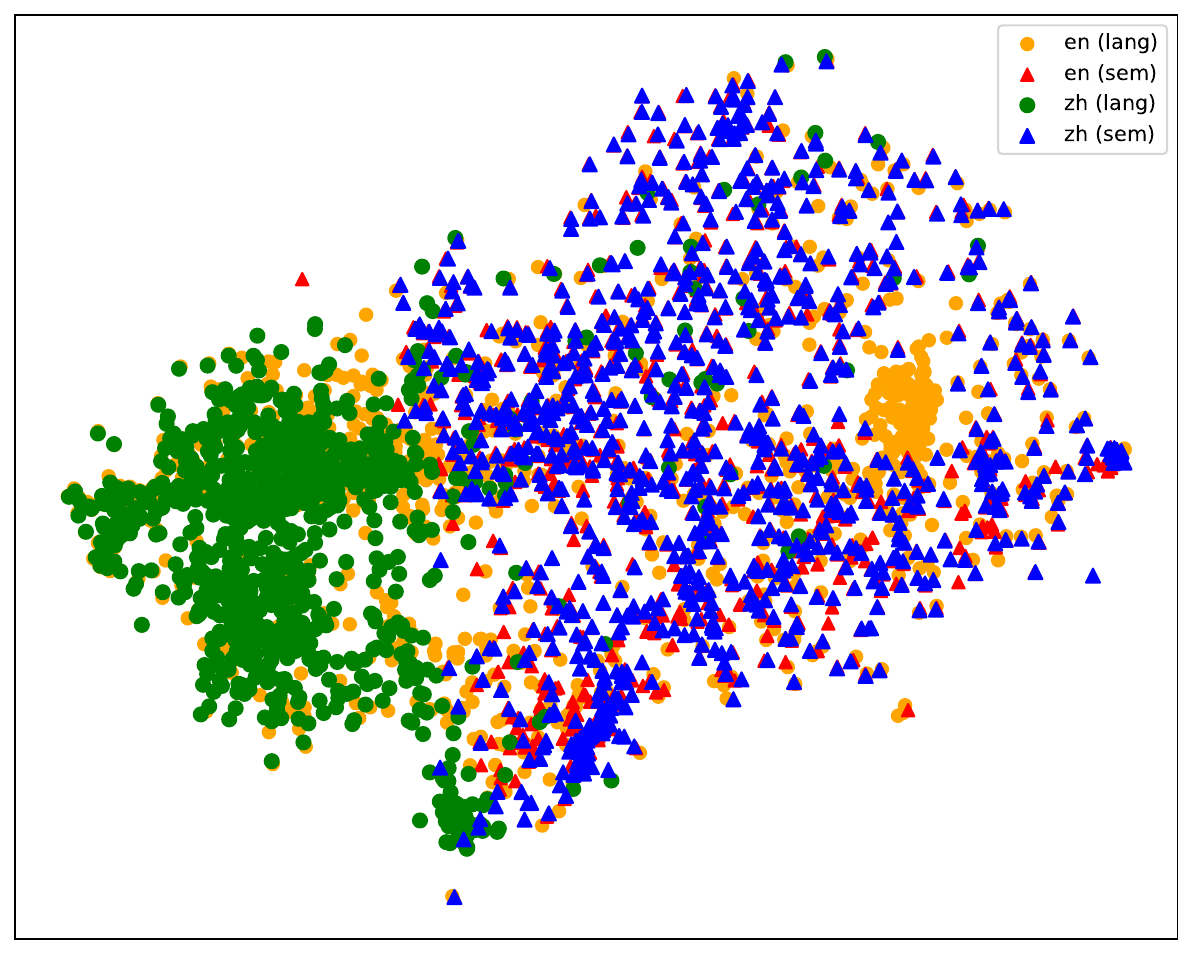}
  \caption{MEAT}
  \label{fig:enzh_meat}
\end{subfigure}%
\begin{subfigure}{.22\textwidth}
  \centering
  \includegraphics[width=\linewidth]{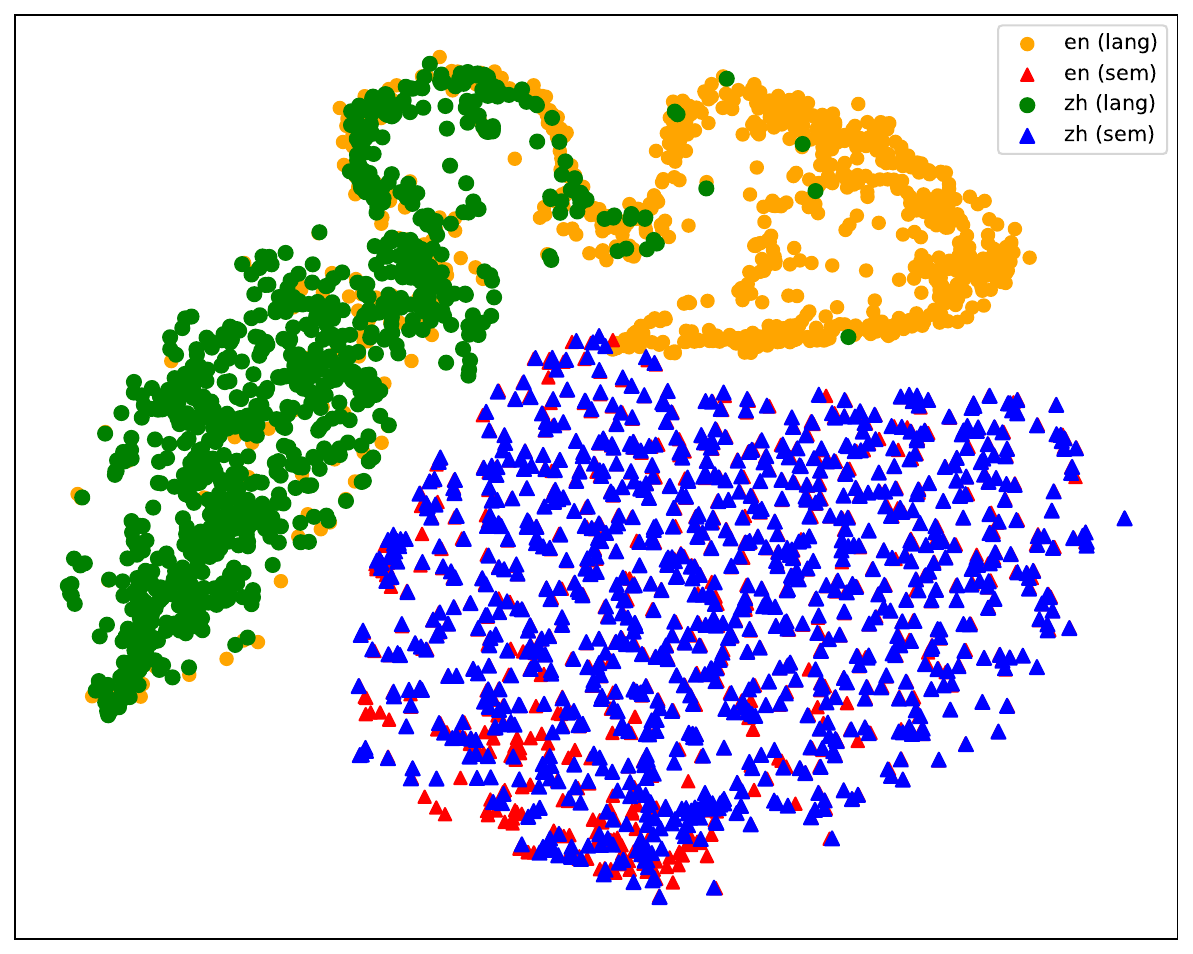}
  \caption{MEAT + \textsc{ORACLE}}
  \label{fig:enzh_meat_oracle}
\end{subfigure}

\caption{Visualization of English-Chinese sentence embeddings from our held-out test set. \textcolor{orange}{Orange} and \textcolor{dark green}{green} denote language embeddings of English and Chinese respectively. \textcolor{red}{Red} and \textcolor{blue}{blue} represent their semantic counterparts. With \textsc{ORACLE}, we can preserve the semantic alignment and clearly divide the language-specific representations.}
\label{fig:visualization_enar}
\end{figure*}

\paragraph{Monolingual vs Cross-lingual.}
We categorize the STS results into two groups of language pairs: monolingual and cross-lingual. For both DREAM and MEAT, regardless of integrating \textsc{ORACLE}, the semantic embedding performance of monolingual language pairs is superior to that of cross-lingual language pairs. However, while the performance gap between monolingual and cross-lingual language pairs is larger for vanilla DREAM or MEAT, \textsc{ORACLE} can mitigate this gap. When applying \textsc{ORACLE}, the performance gap decreases by approximately 0.73 points for LASER, 1.47 points for InfoXLM, and 0.50 points for LaBSE. We report detailed results for each monolingual and cross-lingual language pairs in Appendix \ref{sec:sts_detailed}.

\section{Detailed Analysis}

\subsection{Code-switching}
\label{sec:code_switching}
We manually create a code-switched dataset using bilingual dictionaries from MUSE \cite{conneau2018word}. For each language pair, we randomly replace words in the source sentence with corresponding translations in the target language. Further implementation details are provided in Appendix \ref{sec:code_switching}. As illustrated in Appendix Table \ref{tab:code-switch-results}, our results confirm that integrating \textsc{ORACLE} enhances both semantic and language embedding accuracy, even in practical and challenging scenarios likely encountered during parallel mining, such as code-switching.

% Furthermore, error analysis of the retrieved examples support our observation that \textsc{ORACLE} shows robust mining compared to previous approaches. Specifically, we observe that vanilla DREAM or MEAT are often misled by false cognates \cite{Lalor_Kirsner_2001} such as ``\textit{violation}'' and ``\textit{violence}'' in the source text while using \textsc{ORACLE} can mitigate this issue. We report comprehensive findings in Appendix \ref{sec:cs_error_analysis}.

\subsection{Visualization}
\label{sec:visualization}
In Figure \ref{fig:visualization_enar}, we visualize the LaBSE sentence embedding space using 1,000 English-Chinese sentence pairs from our held-out test set. While previous methods ((a) and (c)) effectively align semantic representations, there is still substantial overlap in the language-specific representations. By applying ORACLE ((b) and (d)), we aim to mitigate the semantic leakage issue, distancing the language representations in parallel sentences while maintaining semantic alignment. We show that this trend is consistent across all language pairs through the visualizations in Appendix \ref{sec:visualizations}.

\subsection{\textsc{ORACLE} Components}
\label{sec:oracle_components}
\textsc{ORACLE} is a multi-task learning objective consisting of two components: intra-class clustering and inter-class separation. Our analysis in Table \ref{tab:oracle_comp} reveals the distinct impact of each component. Interestingly, using only the inter-class clustering loss demonstrates competitive performance, highlighting its critical role in the effectiveness of \textsc{ORACLE}. However, employing either intra-class clustering or inter-class separation alone presents trade-offs. Combining both components yields the most balanced performance, with highest semantic and lowest language embedding retrieval accuracy.

\begin{figure}[!htb]
    \centering
    \includegraphics[width=\linewidth]{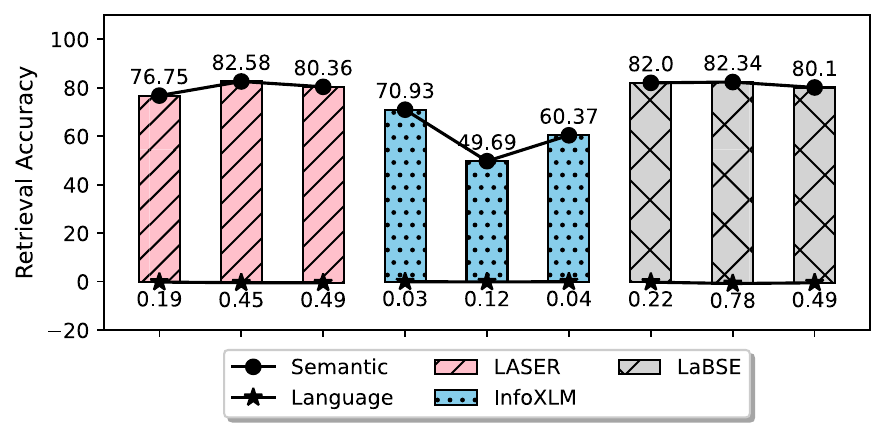}
    \caption{Performance gap between using \textsc{ORACLE} with DREAM (\textit{left}), MEAT (\textit{middle}) or as a stand-alone objective (\textit{right}).}
    \label{fig:oracle_alone}
\end{figure}

Furthermore, we discuss the potential of \textsc{ORACLE} as a stand-alone objective. In Figure \ref{fig:oracle_alone}, we illustrate the performance gap when \textsc{ORACLE} is used alone versus alongside DREAM or MEAT losses. We observe that \textsc{ORACLE} alone effectively mitigates semantic leakage with low language retrieval accuracy. However, this is offset by a decrease in semantic alignment compared to its use with DREAM. Therefore, we opt to integrate \textsc{ORACLE} with previous approaches, making it easily adaptable to various frameworks.

\begin{table}[!htp]
\centering
\resizebox{\linewidth}{!}{%
    \begin{tabular}{l c c c}
    \toprule
    \textbf{Objective} & \textbf{Tatoeba-14} & \textbf{Tatoeba-36} & \textbf{STS} \\ \midrule\midrule
    
     \multicolumn{4}{c}{\textit{Semantic Embedding} (↑)} \\ \midrule

       \textsc{ORACLE} & \textbf{96.11} & 95.53 & \textbf{74.21} \\
       - \textit{$\mathcal{L}_{\mathrm{IC}}$} & 95.89 & 95.38 & 74.13 \\
       - \textit{$\mathcal{L}_{\mathrm{IS}}$} & 96.11 & \textbf{95.54} & 72.81 \\ \midrule\midrule

      \multicolumn{4}{c}{\textit{Language Embedding} (↓)} \\ \midrule

       \textsc{ORACLE} & \textbf{7.74} & \textbf{9.17} & \textbf{16.47} \\
       - \textit{$\mathcal{L}_{\mathrm{IC}}$} & 37.78 & 39.15 & 30.14 \\
       - \textit{$\mathcal{L}_{\mathrm{IS}}$} & 8.07 & 9.59 & 18.20 \\

    \bottomrule
    \end{tabular}
}
\caption{Performance change when removing each component of \textsc{ORACLE} from LaBSE sentence embeddings. $\mathcal{L}_{\mathrm{IC}}$: Intra-class clustering; $\mathcal{L}_{\mathrm{IS}}$: Inter-class separation. \textbf{Bold} denotes best results for each semantic and language embedding.}
\label{tab:oracle_comp}
\end{table}

% \subsection{Bilingual vs. Multilingual}
% We train disentangled representations for 12 language pairs altogether in a multilingual, multi-task manner (the multilingual version). Additionally, we train a bilingual version for 3 language pairs, en-de, en-es, and en-fr, each with equal dataset sizes as the multilingual version. In Appendix Table \ref{tab:bimulti}, we examine that the bilingual approach better disentangle semantic and language-specific representations with lower language retrieval accuracy. In contrast, the multilingual version needs to share the embedding space with many languages \cite{huang-etal-2021-improving-zero}, thus showing lower accuracy in language embedding separation. However, the bilingual version does not significantly improve semantic alignment for the specifically trained language pair. Further, we also observe an over-fitting behavior of MEAT with extremely high language retrieval accuracy, indicating lower robustness of the bilingual training method compared to the multilingual version.

\section{Conclusion}
We explore the issue of semantic leakage, which we define as when language-specific information is leaked into the semantic representations, across various multilingual encoders and objective functions. Addressing this issue is crucial for achieving disentangled semantic and language representations, which is a cornerstone for effective parallel mining. We introduce \textsc{ORACLE}, a simple and effective training objective designed to enforce orthogonality between semantic and language embeddings. Through comprehensive evaluations, we demonstrate that integrating \textsc{ORACLE} not only improves semantic alignment but also ensures clear separation of language representations, as evidenced by embedding space visualization. Further, we conduct detailed analysis to understand the roles of the two key components of \textsc{ORACLE}: intra-class clustering and inter-class separation. While our study primarily focuses on integrating \textsc{ORACLE} with DREAM and MEAT, our method is easily adaptable to various frameworks, offering promising avenues for future work.

% Additionally, we confirm the efficacy of \textsc{ORACLE} in real-world challenging scenarios using code-switched dataset.

\section{Limitations}
Our work highlights the effectiveness of \textsc{ORACLE} in addressing semantic leakage and improving semantic alignment. While \textsc{ORACLE} demonstrates competitive performance as a stand-alone objective, its integration with DREAM or MEAT losses yields even better results. This limits the usage of \textsc{ORACLE} to be used alongside other methods. This opens many questions for future work to further explore the optimal combination of existing approaches and \textsc{ORACLE}.

Moreover, our study assesses the disentanglement of semantic and language representations in embeddings, focusing on two key aspects: the alignment of semantics in bilingual sentence pairs and the separation of language-specific information. While \textsc{ORACLE} effectively addresses the separation of language-specific information, we notice a trade-off in semantic alignment for certain language pairs. Future works can delve into methods that more efficiently mitigate semantic leakage without compromising semantic representation quality.

Lastly, our experiments are limited to 12 selected language pairs for training. To expand the scope of our study, future work could involve a wider array of language pairs and a broader range of multilingual encoder baselines.

\bibliography{custom}

\appendix

\section{Implementation Details}
\label{sec:training_details}
\subsection{Training Corpus}
\label{sec:train_corpus}
In this section, we discuss the implementation details of our \textsc{ORACLE} objective. We describe the specific training corpus utilized for each language pair in Table \ref{tab:training_corpus}.

\subsection{Seen vs. Unseen Languages}
In Table \ref{tab:seen_unseen}, we present the list of seen and unseen languages for each multilingual sentence encoder baseline, listed in alphabetical order. Across all encoders, Guaraní (gn) is categorized as an unseen language, while Aymara (ay) is classified as an unseen language for InfoXLM and LaBSE.

\subsection{Training Details}
\label{appendix:training}
Size of each MLP layer is embedding size of the encoder (1024 for LASER and 768 for XLM-R and LaBSE) by the number of language pairs (12). For training, we use Adam optimizer with an initial learning rate as 1e-5 and a batch size of 512. We train the model for 10,000 iterations, evaluating the model’s performance on the validation set at the end of each iteration. We implement early stopping to halt training when there is no improvement over 10 consecutive iterations. We find that DREAM converges in approximately 250 iterations and MEAT in 20 iterations.

\section{Detailed Results}
\label{sec:detailed_results}

\subsection{Held-out Test set}
\label{sec:cross_lingual_detailed}
In Table \ref{tab:test_set}, we present detailed results for the cross-lingual sentence retrieval task using our held-out test set. The top section shows the performance of the initial sentence embeddings from LASER, InfoXLM, and LaBSE. In the middle section, we detail the accuracy of extracted semantic embeddings, while the bottom rows represent the language embedding accuracy. \textsc{ORACLE}, particularly for LaBSE, notably reduces language embedding accuracy, indicating a mitigation of semantic leakage compared to the vanilla DREAM or MEAT frameworks. Additionally, we observe an improvement in the semantic retrieval accuracy across all encoder baselines on average.

%This comes with a slight decrease in semantic accuracy. However, combining ORACLE with MEAT improves semantic retrieval accuracy, achieving the highest accuracy on average.

% \input{tables/oracle_meat}

\subsection{Tatoeba}
\label{sec:tatoeba_detailed}
In our analysis of the Tatoeba dataset detailed in Table \ref{tab:tatoeba_detailed}, we exclude two language pairs (en-ay and en-gn) as Tatoeba does not support them. We show that a similar trend is observed: training MLP networks with \textsc{ORACLE} not only improves semantic alignment but also effectively addresses the semantic leakage issue in the vanilla DREAM or MEAT. Also, we observe that training LaBSE sentence embeddings with \textsc{ORACLE} yields state-of-the-art semantic retrieval accuracy compared to previous methods.

% \subsection{BUCC}
% BUCC is a bitext mining task where the model ranks all possible sentence pairs and predicts sentences with higher score than the fixed threshold. We report F1-scores for semantic and language-specific representations across language pairs: de-en, fr-en, ru-en, and zh-en. Our findings align with trends observed in the held-out test set (Table \ref{tab:test_set}) and Tatoeba (Table \ref{tab:tatoeba_detailed}). Using \textsc{ORACLE} to learn disentangled representations raises the score for semantic and lowers the score for language embedding.

% \input{tables/bucc}

\subsection{Semantic Textual Similarity}
\label{sec:sts_detailed}
We present detailed numerical results for the monolingual and cross-lingual STS benchmark in Table \ref{tab:sts}. The results support our observation from the cross-lingual retrieval tasks that \textsc{ORACLE} helps address both semantic alignment and the semantic leakage issue.

% \section{Detailed Ablations}
% \label{sec:detailed_ablation}
% \subsection{\textsc{ORACLE} Components}
% \label{sec:oracle_meat}
% We examine the impact of each \textsc{ORACLE} component with MEAT in Table \ref{tab:oracle_meat}. Note that we report the numerical values for DREAM in Section \ref{sec:oracle_components}. We can observe similar trends for MEAT; intra-class clustering and inter-class separation collaboratively enhance semantic alignment and separation of language-specific information. Notably for MEAT, removing inter-class separation significantly diminishes performance, highlighting its importance in the framework.

\begin{table}
\centering
\resizebox{\linewidth}{!}{%
    \begin{tabular}{l l}
    \toprule
    \textbf{Training corpus} & \textbf{Language pair}\\ \midrule
    \textbf{Europarl} & en-de, en-es, en-fr, en-it, en-nl, en-pt \\
    \textbf{Wikimatrix} & en-ar, en-ja, en-ro, en-zh \\ 
    \textbf{Tatoeba} & en-gn \\ 
    \textbf{NLLB} & en-ay \\ 
    \bottomrule
    \end{tabular}
}
\caption{Summary of training corpus for each language pair.}
\label{tab:training_corpus}
\end{table}
\begin{table}
\centering
\resizebox{\linewidth}{!}{%
    \begin{tabular}{l l l}
    \toprule
    \textbf{Encoder} & \textbf{Seen} & \textbf{Unseen}\\ \midrule
    % \textbf{mBERT} & ar, de, en, es, fr, it, ja, nl, pt, ro, zh & ay, gn \\
    \textbf{LASER} & ar, ay, de, en, es, fr, it, ja, nl, pt, ro, zh & gn \\
    \textbf{InfoXLM} & ar, de, en, es, fr, it, ja, nl, pt, ro, zh & ay, gn \\
    \textbf{LaBSE} & ar, de, en, es, fr, it, ja, nl, pt, ro, zh & ay, gn \\
    % \textbf{} & ar, de, en, es, fr, it, ja, nl, pt, ro, zh & ay, gn \\ 
    \bottomrule
    \end{tabular}
}
\caption{Seen and unseen languages for each pre-trained multilingual encoder. Note that seen refers to languages used during pre-training.}
\label{tab:seen_unseen}
\end{table}

\section{Code-switching}
\subsection{Dataset Construction}
For our code-switching evaluation, we utilize bilingual dictionaries sourced from MUSE \cite{conneau2018word}. MUSE provides dictionaries in both the to English (\textsc{XX-En}) and from English (\textsc{En-XX}) directions. Specifically, we focus on dictionaries with the \textsc{XX-En} direction. These dictionaries comprise root words in the source language paired with their corresponding translations in the target language. As noted by \citet{conneau2018word}, the translations are generated using an internal translation tool, which accounts for word polysemy, resulting in some root words having multiple translations.

For each language pair listed in Table \ref{tab:tatoeba}, we randomly substitute words in the source sentences with their corresponding translations in the target language, utilizing the dictionaries from MUSE. We ensure that the selected sentences of our code-switching evaluation contain at least one code-switched word. The resulting dataset comprises 1,000 sentences per language pair. We show examples of the manually created code-switched dataset in Table \ref{tab:code-switch}.

\begin{table*}
\centering
\resizebox{\linewidth}{!}{%
    \begin{tabular}{l l}
    \toprule
    \textbf{Language pair} & \textbf{Code-switched Example}\\ \midrule
    
    \multirow{2}{*}\textbf{De-En} & Source: Wie \textit{long} \textit{should} Tom \textit{and} \textit{I} hierbleiben? \\
    & Target: How long are Tom and I supposed to stay here? \\ \midrule

    \multirow{2}{*}\textbf{Fr-En} & Source: Je \textit{am here} jusqu'à \textit{three} heures. \\
    & Target: I will stay here till three o'clock. \\ \midrule

    \multirow{2}{*}\textbf{It-En} & Source: Fadil sparò al \textit{dog} di Dania. \\
    & Target: Fadil shot Dania's dog. \\ \midrule

    \multirow{2}{*}\textbf{Ro-En} & Source: E \textit{traditional} să gates \textit{black} la înmormântare. \\
    & Target: It is traditional to wear black to a funeral. \\    
    \bottomrule
    \end{tabular}
}
\caption{Examples of code-switched dataset manually created using bilingual dictionaries from MUSE \cite{conneau2018word}. \textit{Italic} represent words that are code-switched in the source sentence.}
\label{tab:code-switch}
\end{table*}

\subsection{Results}
In Table \ref{tab:code-switch-results}, we present the retrieval accuracy achieved on our code-switched dataset. Similar to the trends observed in other tasks, integrating \textsc{ORACLE} consistently improves both semantic and language embedding accuracy across all multilingual encoder baselines.

% \subsection{Error Analysis}
% \label{sec:cs_error_analysis}
% We conduct further error analysis to examine cases where vanilla DREAM or MEAT fails and \textsc{ORACLE} succeeds. Through manual inspection, we observe that vanilla DREAM or MEAT are often misled by false cognates. False cognates indicate words with similar form but unrelated meanings, which can be within the same language or from different languages \cite{Lalor_Kirsner_2001}. We present several examples of error cases in Table \ref{tab:error_analysis}.

\section{Visualizations}
\label{sec:visualizations}
\begin{figure*}[]
\centering

\begin{subfigure}{.24\textwidth}
  \centering
  \includegraphics[width=\linewidth]{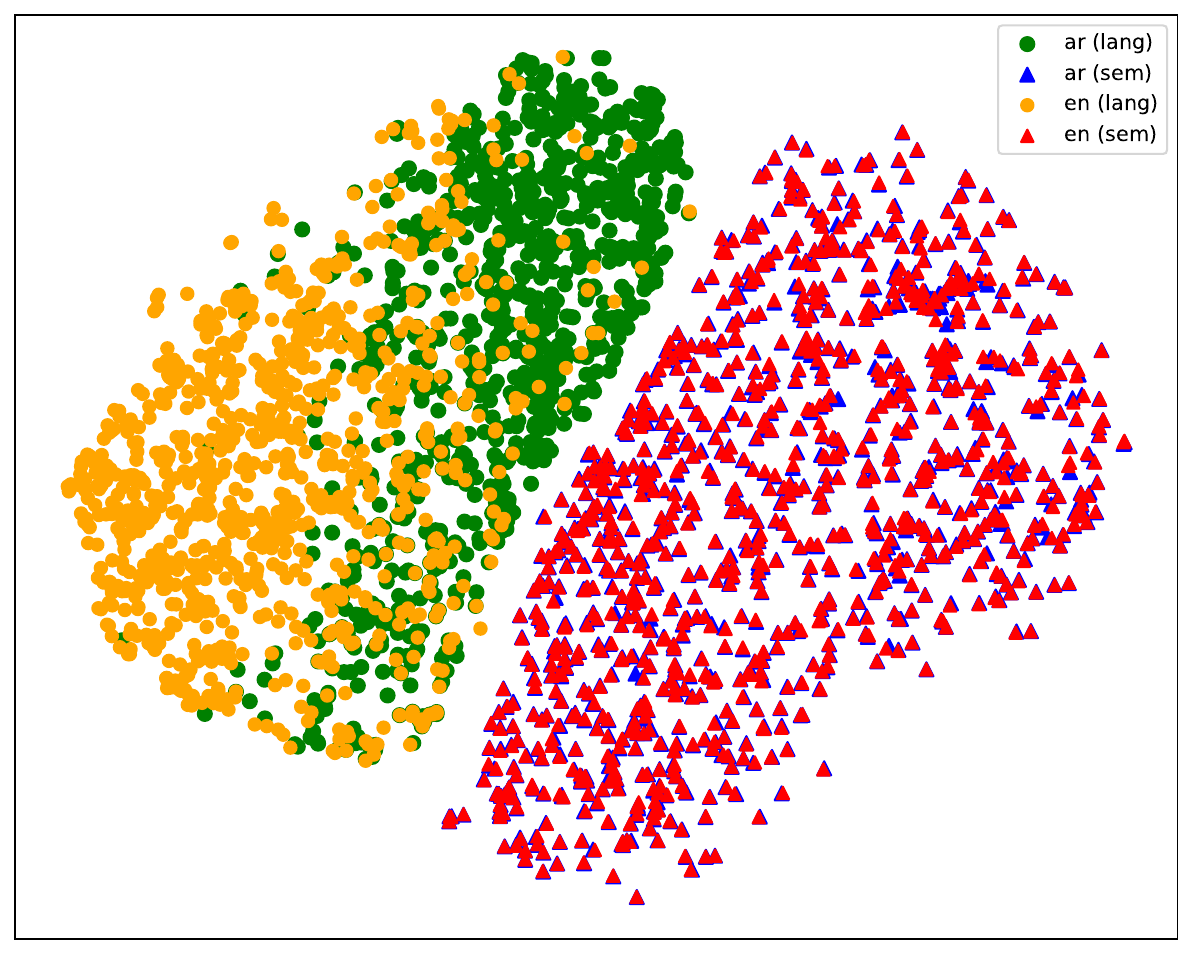}
  \caption{DREAM}
\end{subfigure}%
\begin{subfigure}{.24\textwidth}
  \centering
  \includegraphics[width=\linewidth]{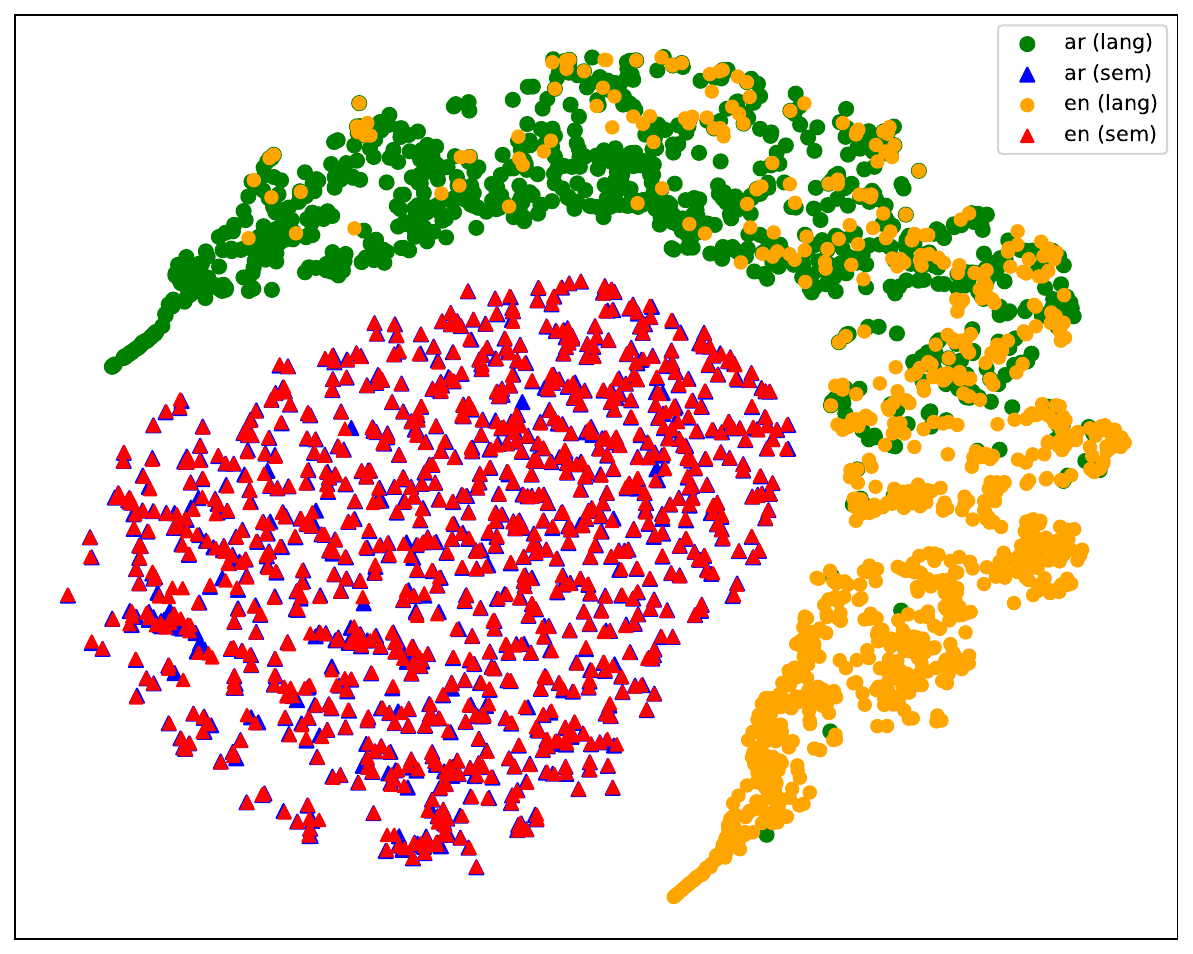}
  \caption{DREAM + \textsc{ORACLE}}
\end{subfigure}
\begin{subfigure}{.24\textwidth}
  \centering
  \includegraphics[width=\linewidth]{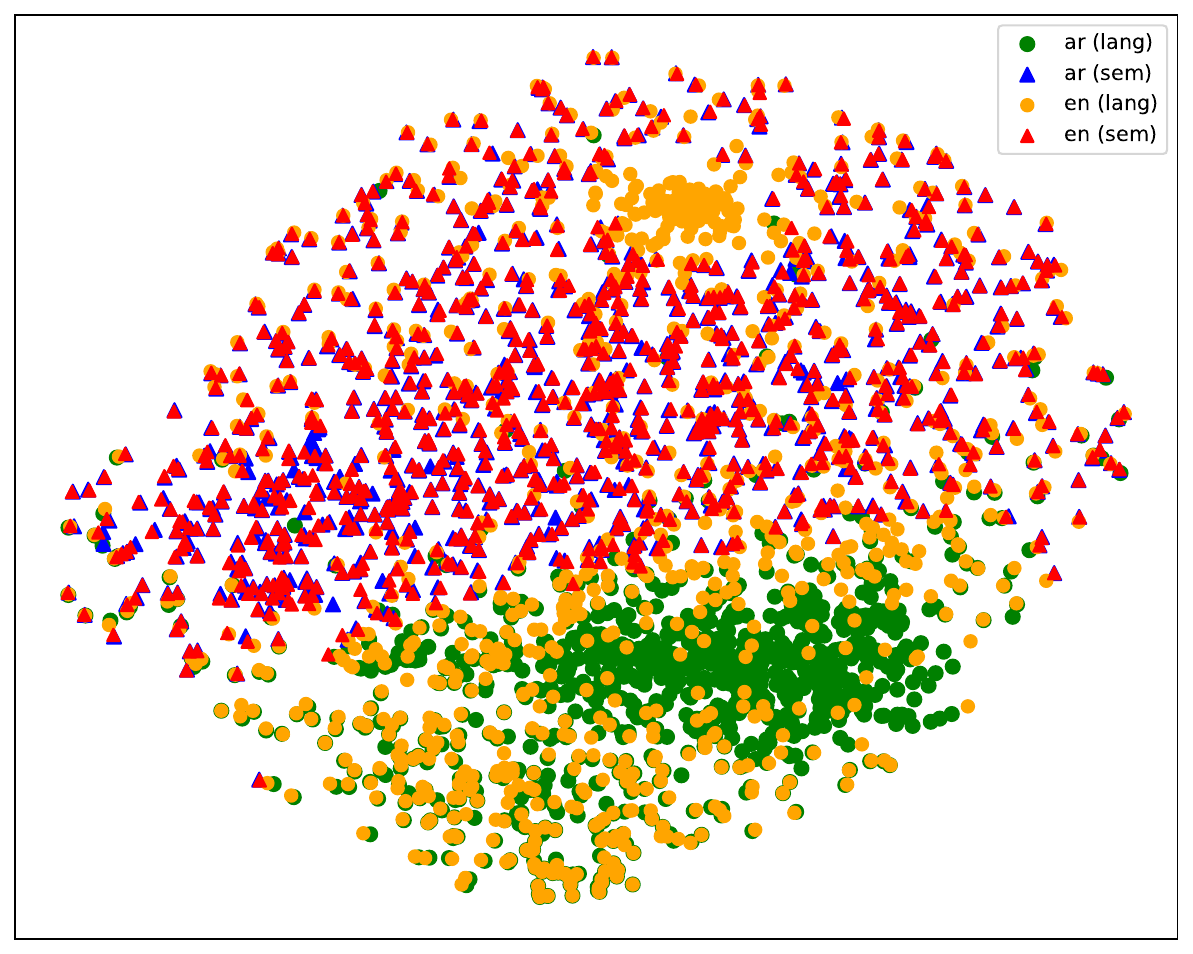}
  \caption{MEAT}
\end{subfigure}%
\begin{subfigure}{.24\textwidth}
  \centering
  \includegraphics[width=\linewidth]{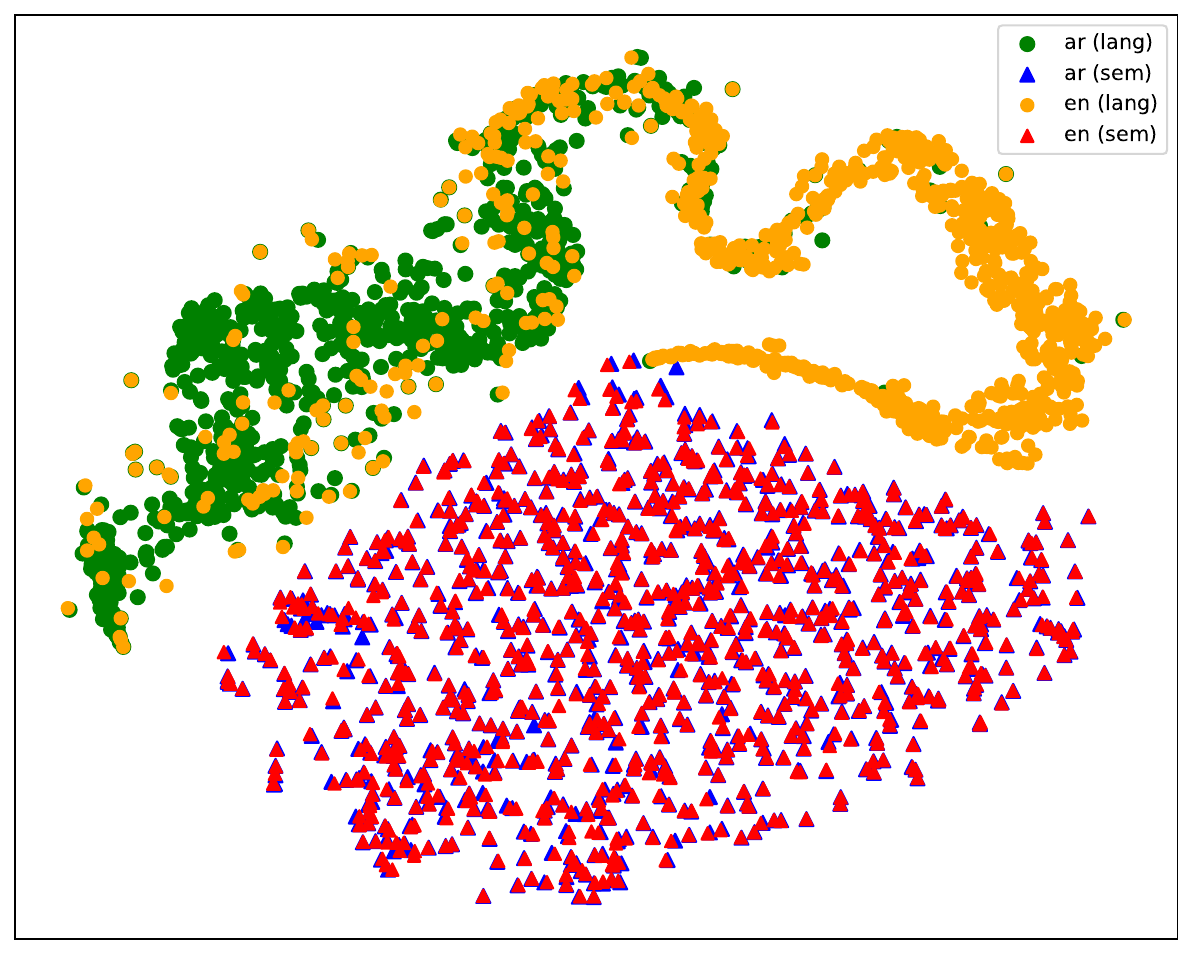}
  \caption{MEAT + \textsc{ORACLE}}
\end{subfigure}

\caption{LaBSE sentence embeddings for English-Arabic sentence pair.}
\label{fig:enar}
\end{figure*}

\begin{figure*}[]
\centering

\begin{subfigure}{.24\textwidth}
  \centering
  \includegraphics[width=\linewidth]{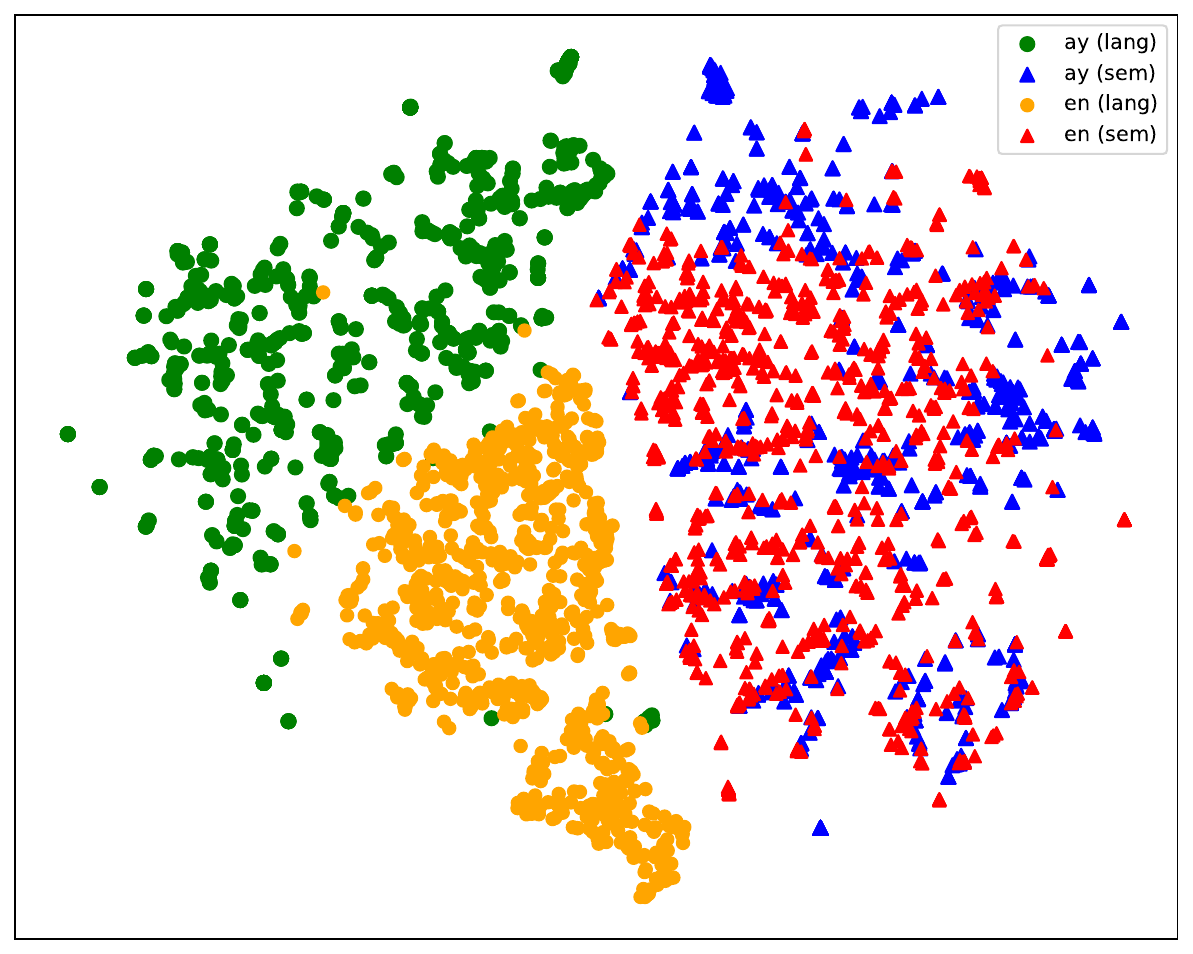}
  \caption{DREAM}
\end{subfigure}%
\begin{subfigure}{.24\textwidth}
  \centering
  \includegraphics[width=\linewidth]{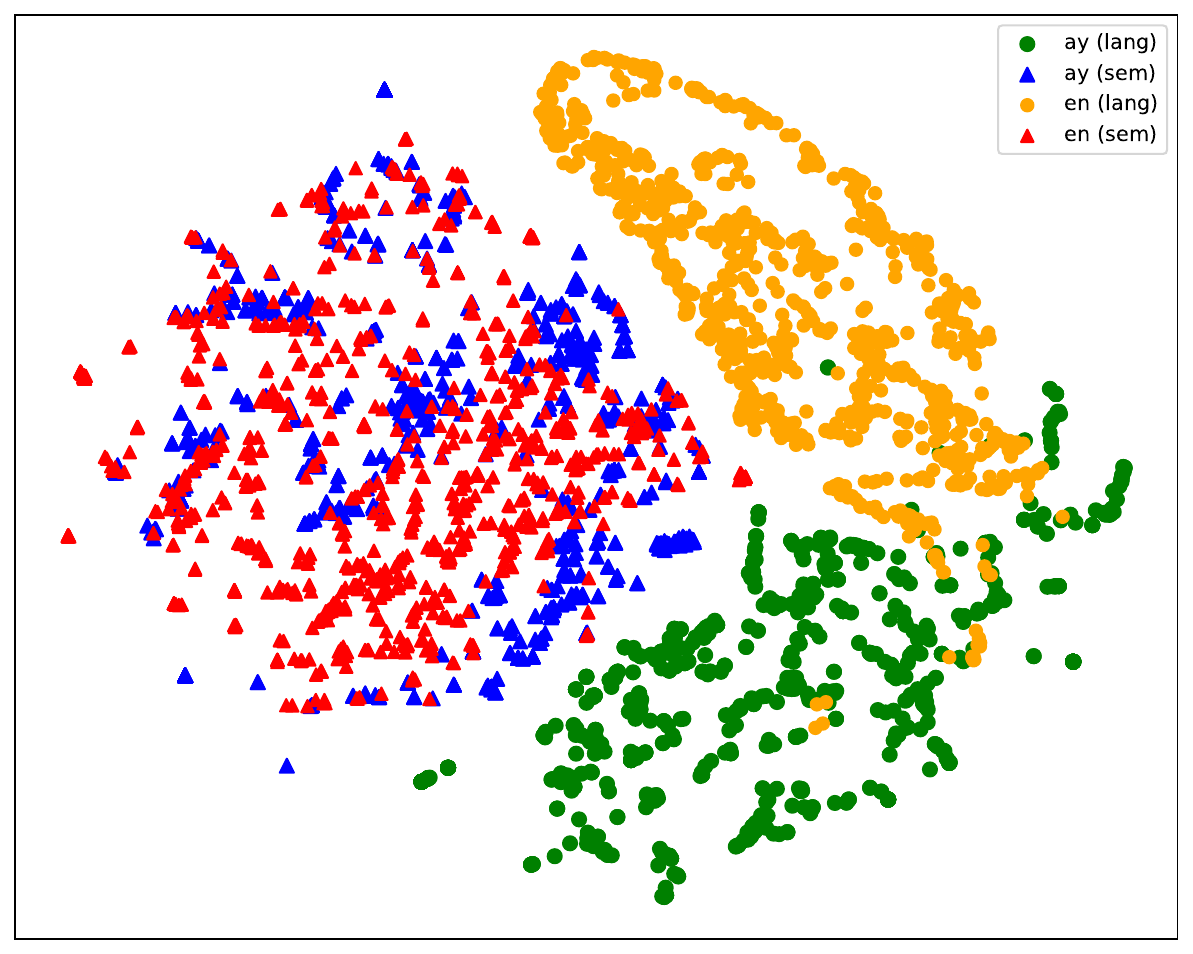}
  \caption{DREAM + \textsc{ORACLE}}
\end{subfigure}
\begin{subfigure}{.24\textwidth}
  \centering
  \includegraphics[width=\linewidth]{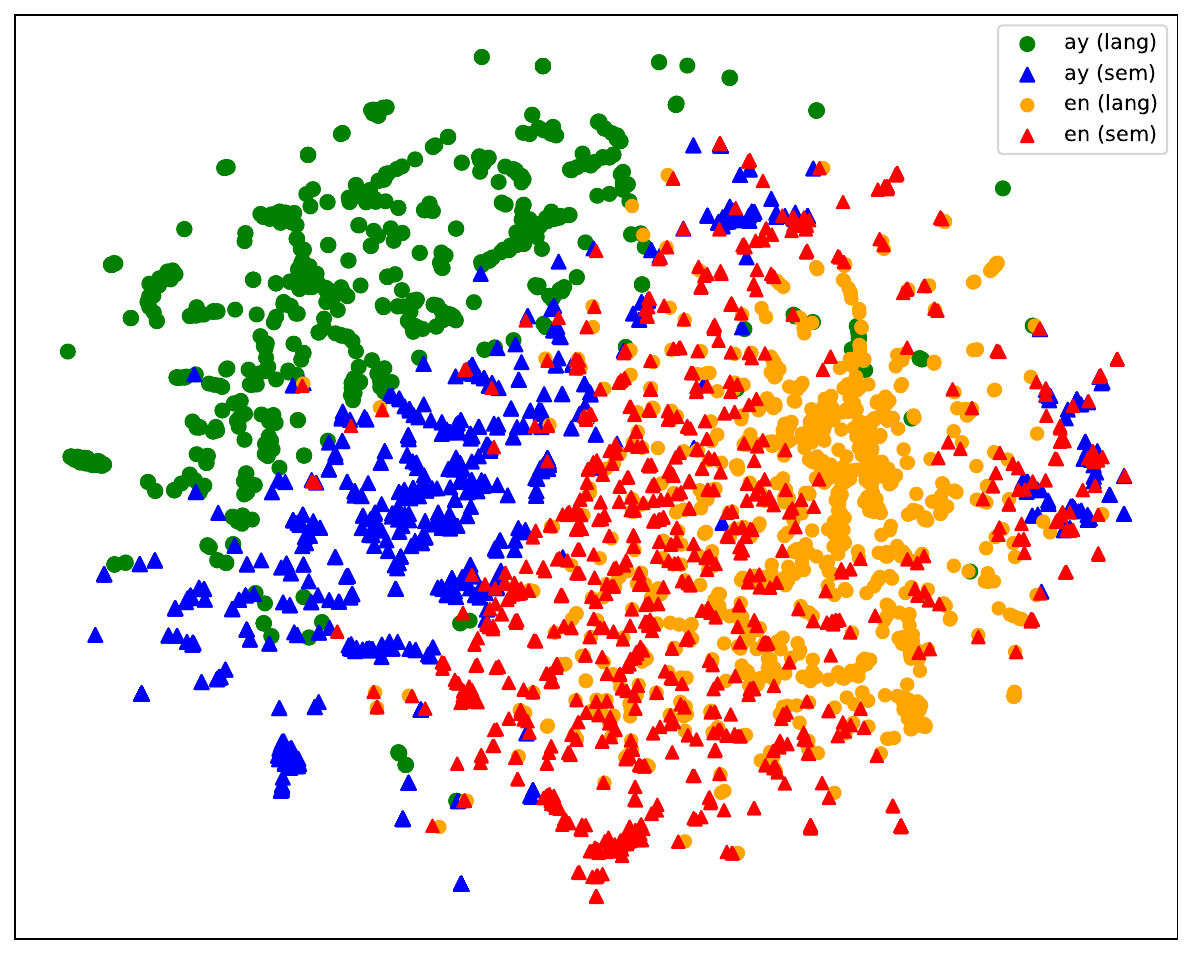}
  \caption{MEAT}
\end{subfigure}%
\begin{subfigure}{.24\textwidth}
  \centering
  \includegraphics[width=\linewidth]{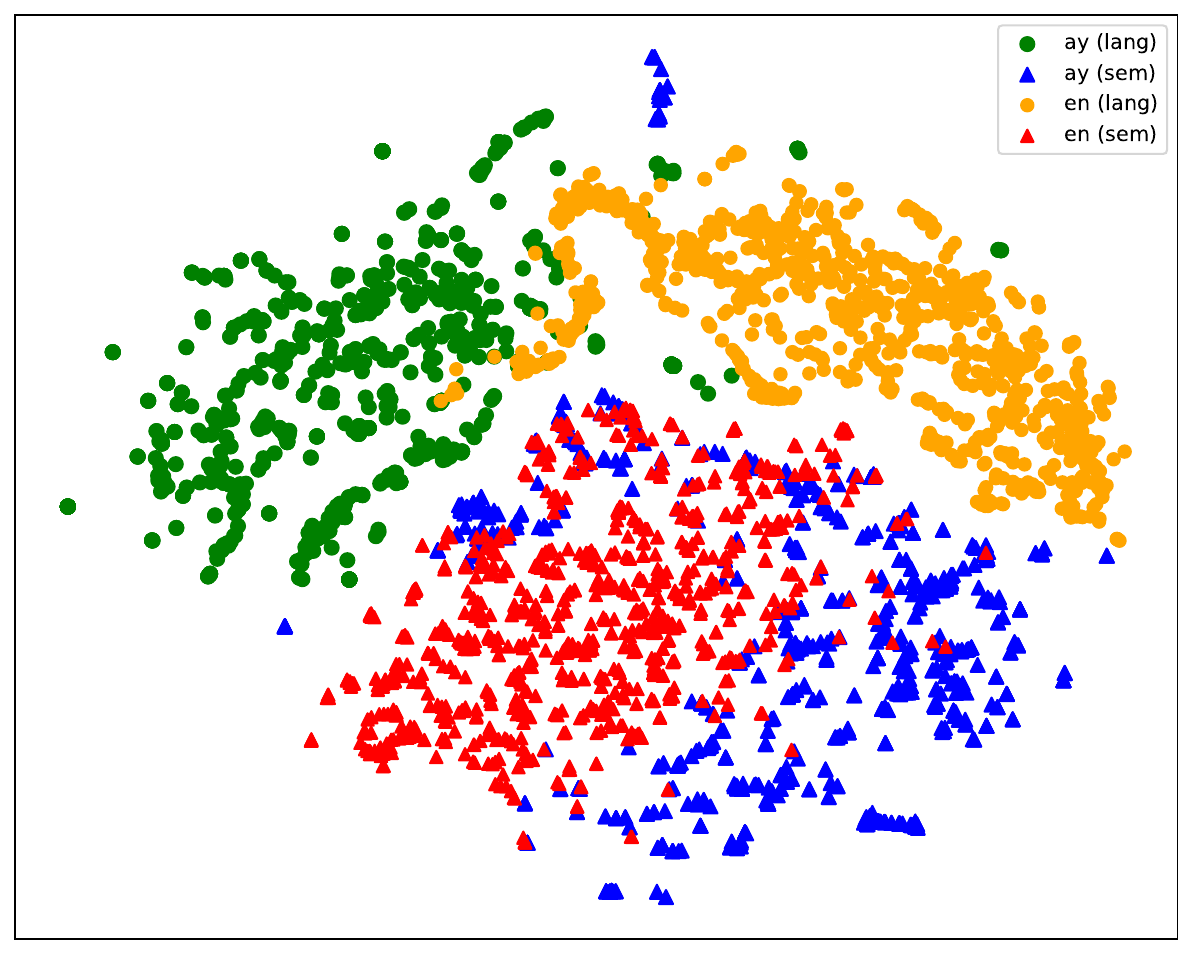}
  \caption{MEAT + \textsc{ORACLE}}
\end{subfigure}

\caption{LaBSE sentence embeddings for English-Aymara sentence pair.}
\label{fig:enay}
\end{figure*}

\begin{figure*}[]
\centering

\begin{subfigure}{.24\textwidth}
  \centering
  \includegraphics[width=\linewidth]{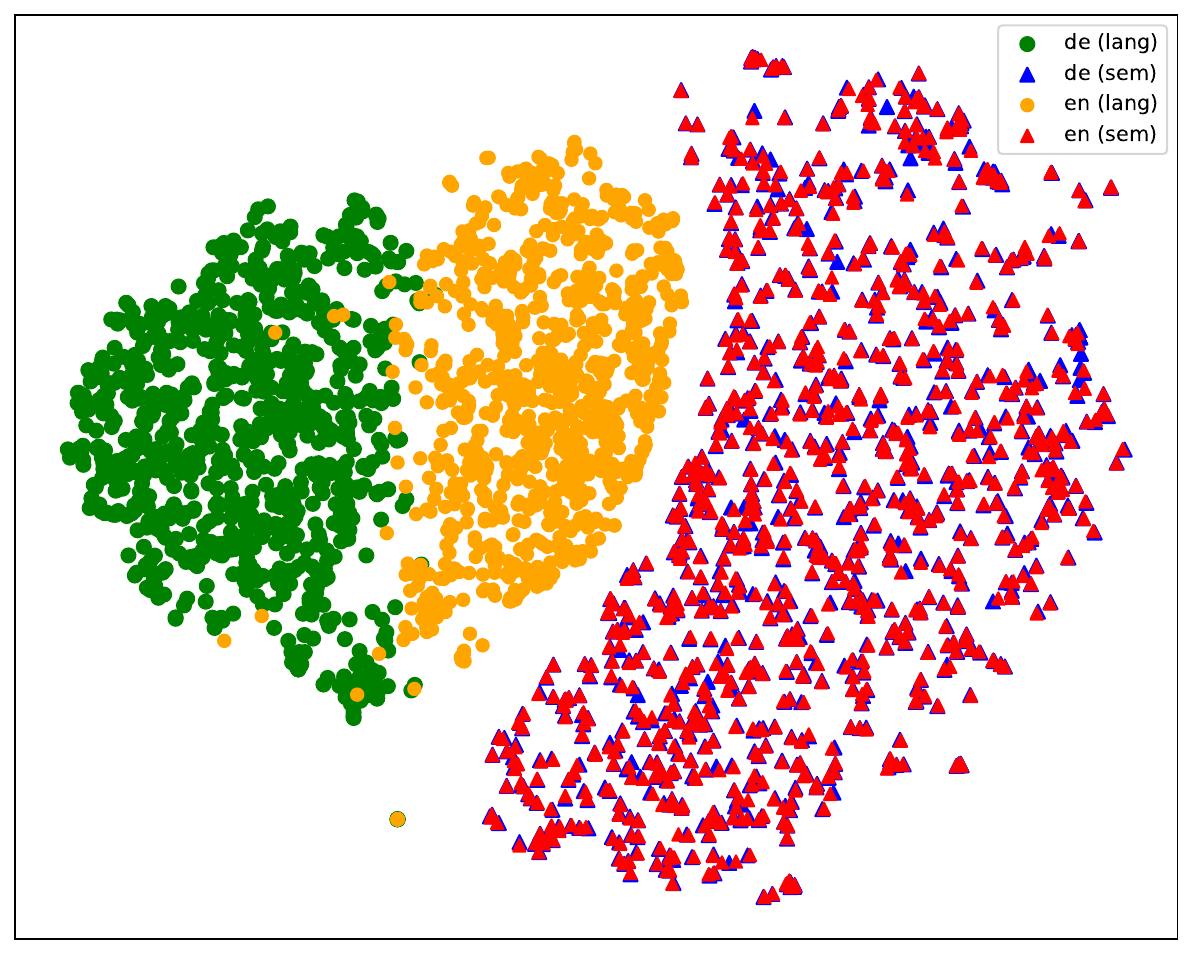}
  \caption{DREAM}
\end{subfigure}%
\begin{subfigure}{.24\textwidth}
  \centering
  \includegraphics[width=\linewidth]{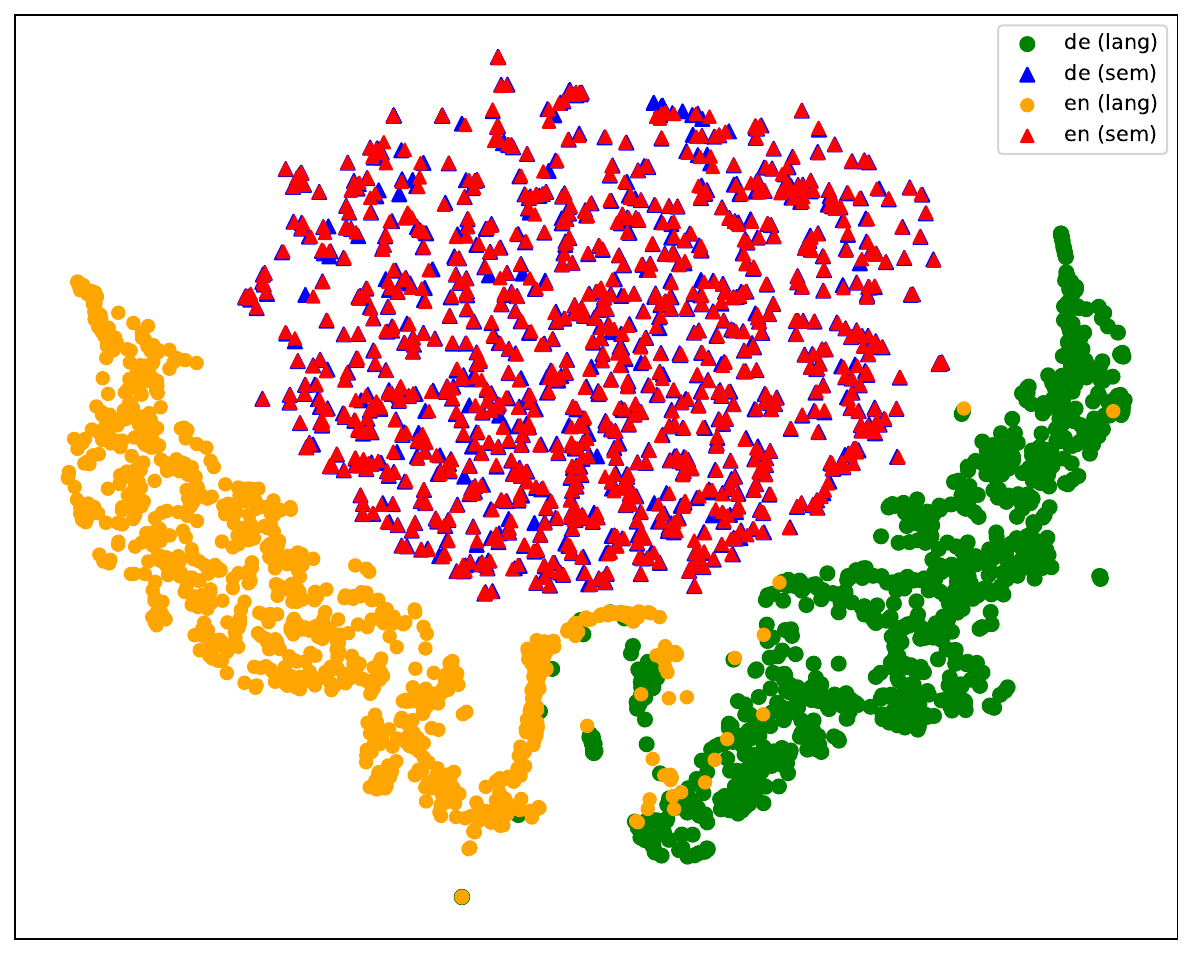}
  \caption{DREAM + \textsc{ORACLE}}
\end{subfigure}
\begin{subfigure}{.24\textwidth}
  \centering
  \includegraphics[width=\linewidth]{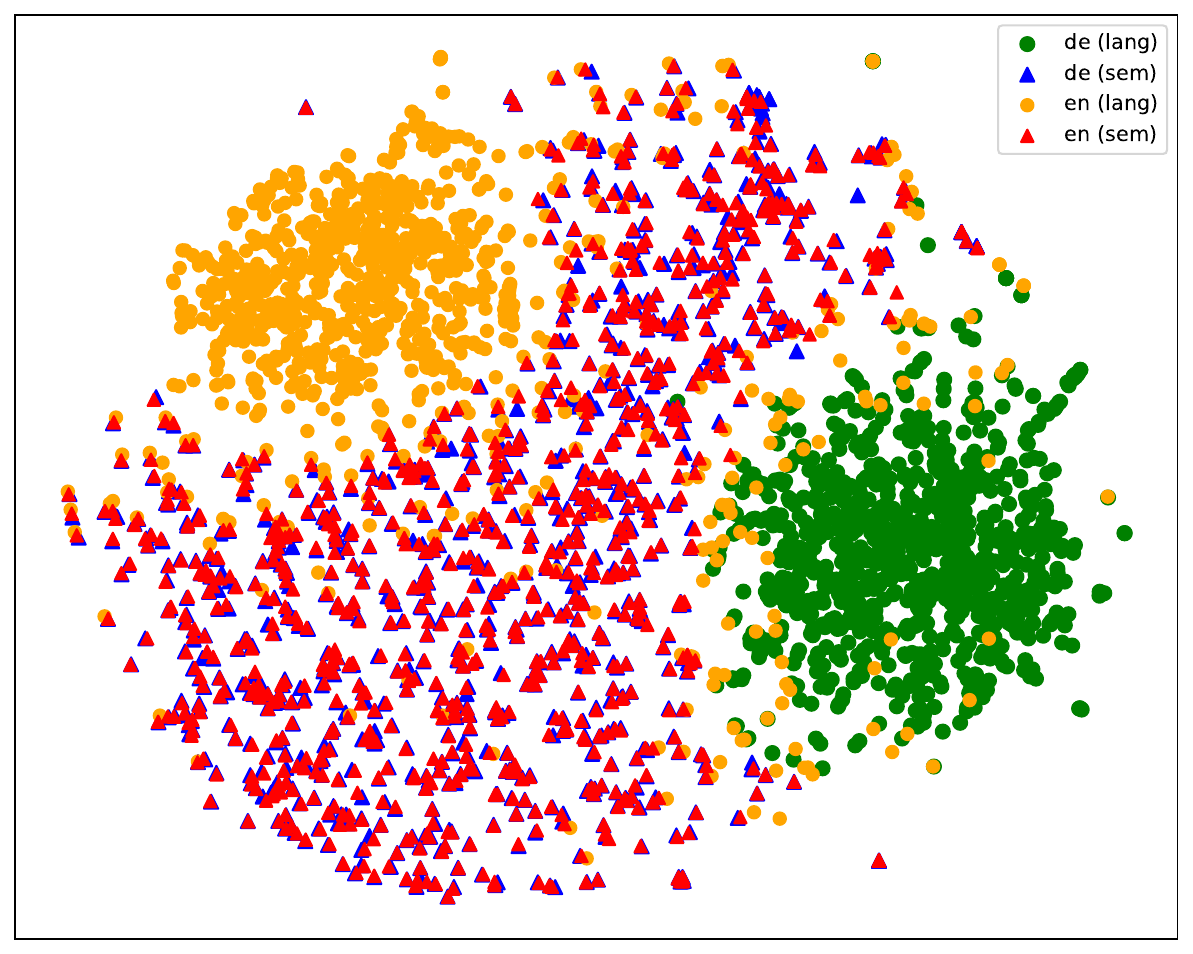}
  \caption{MEAT}
\end{subfigure}%
\begin{subfigure}{.24\textwidth}
  \centering
  \includegraphics[width=\linewidth]{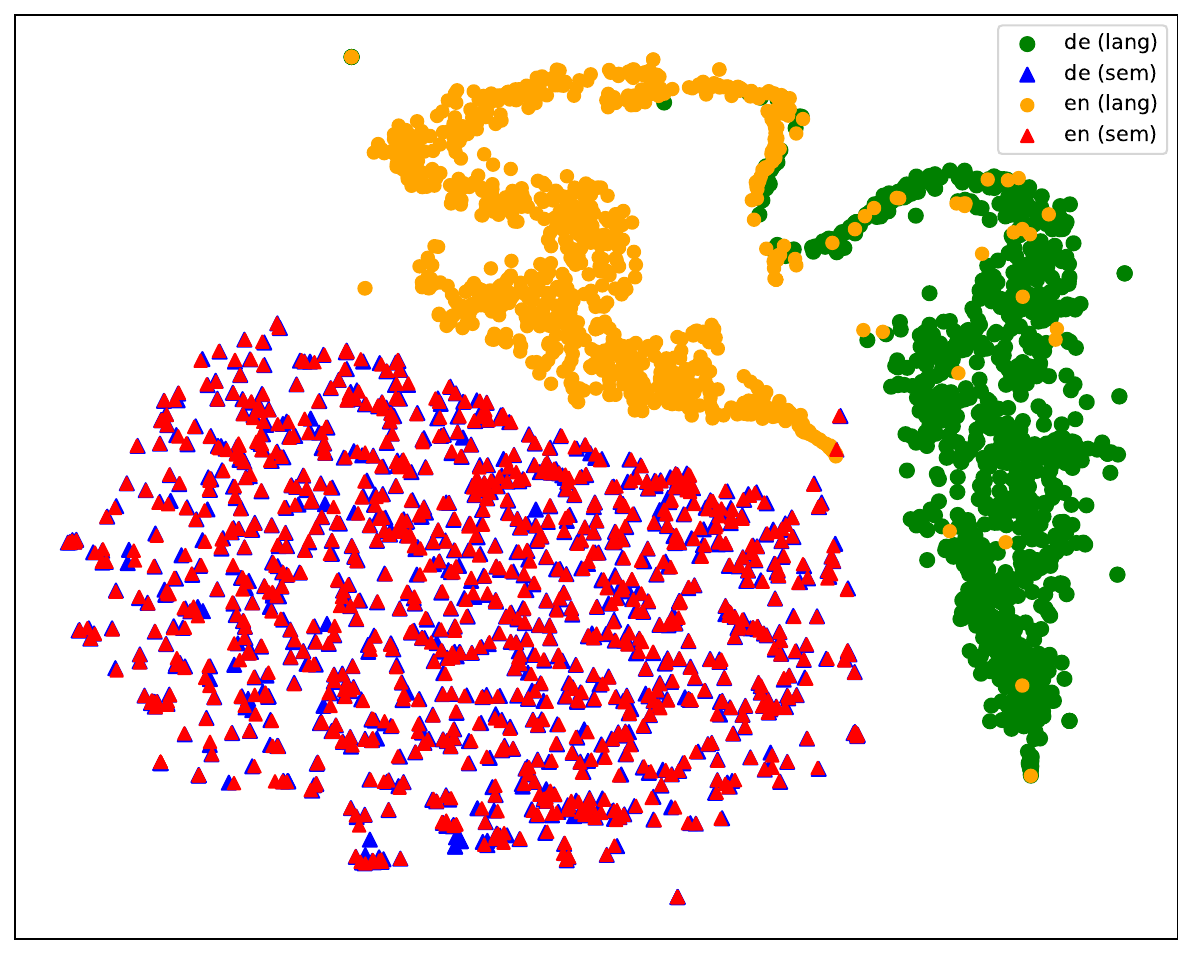}
  \caption{MEAT + \textsc{ORACLE}}
\end{subfigure}

\caption{LaBSE sentence embeddings for English-German sentence pair.}
\label{fig:ende}
\end{figure*}

\begin{figure*}[]
\centering

\begin{subfigure}{.24\textwidth}
  \centering
  \includegraphics[width=\linewidth]{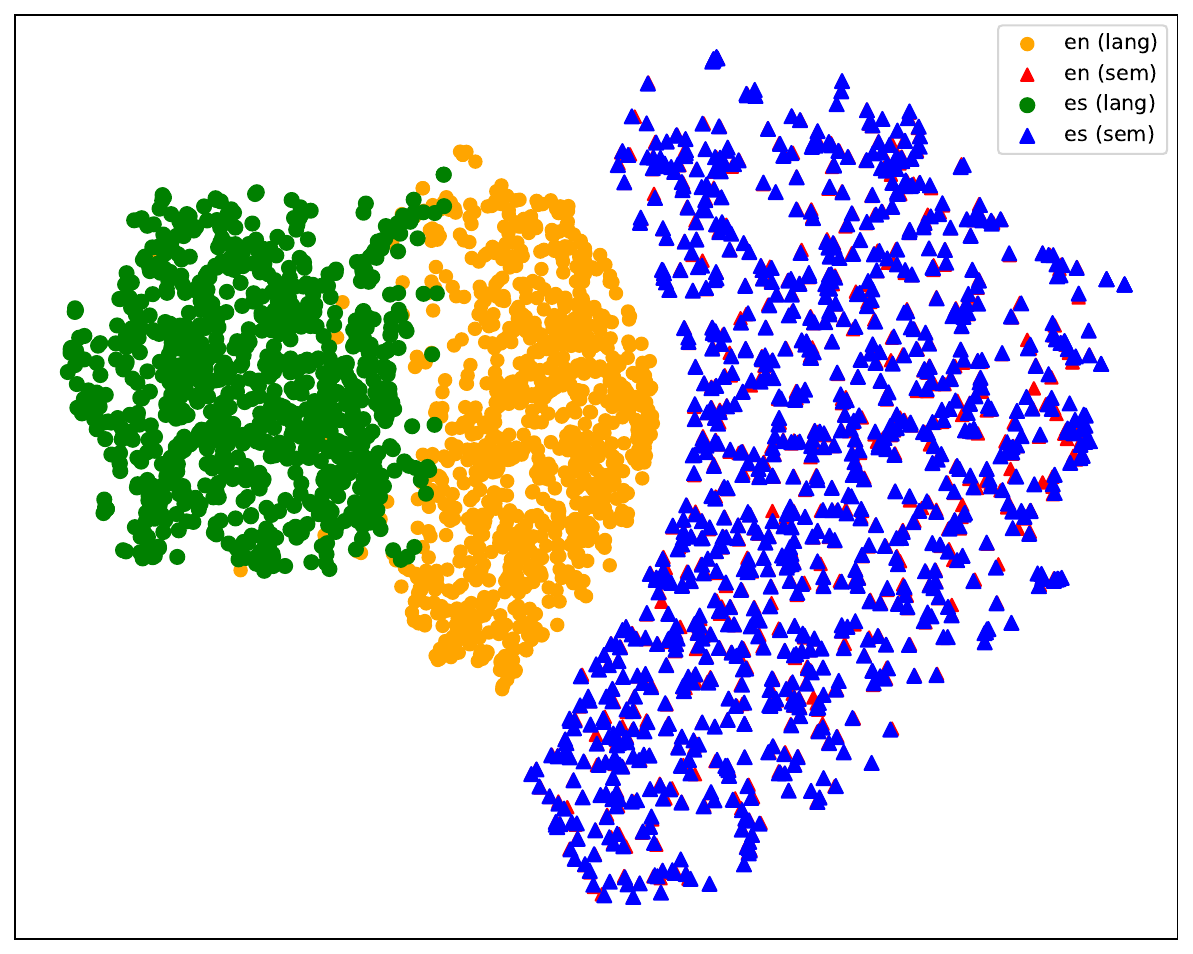}
  \caption{DREAM}
\end{subfigure}%
\begin{subfigure}{.24\textwidth}
  \centering
  \includegraphics[width=\linewidth]{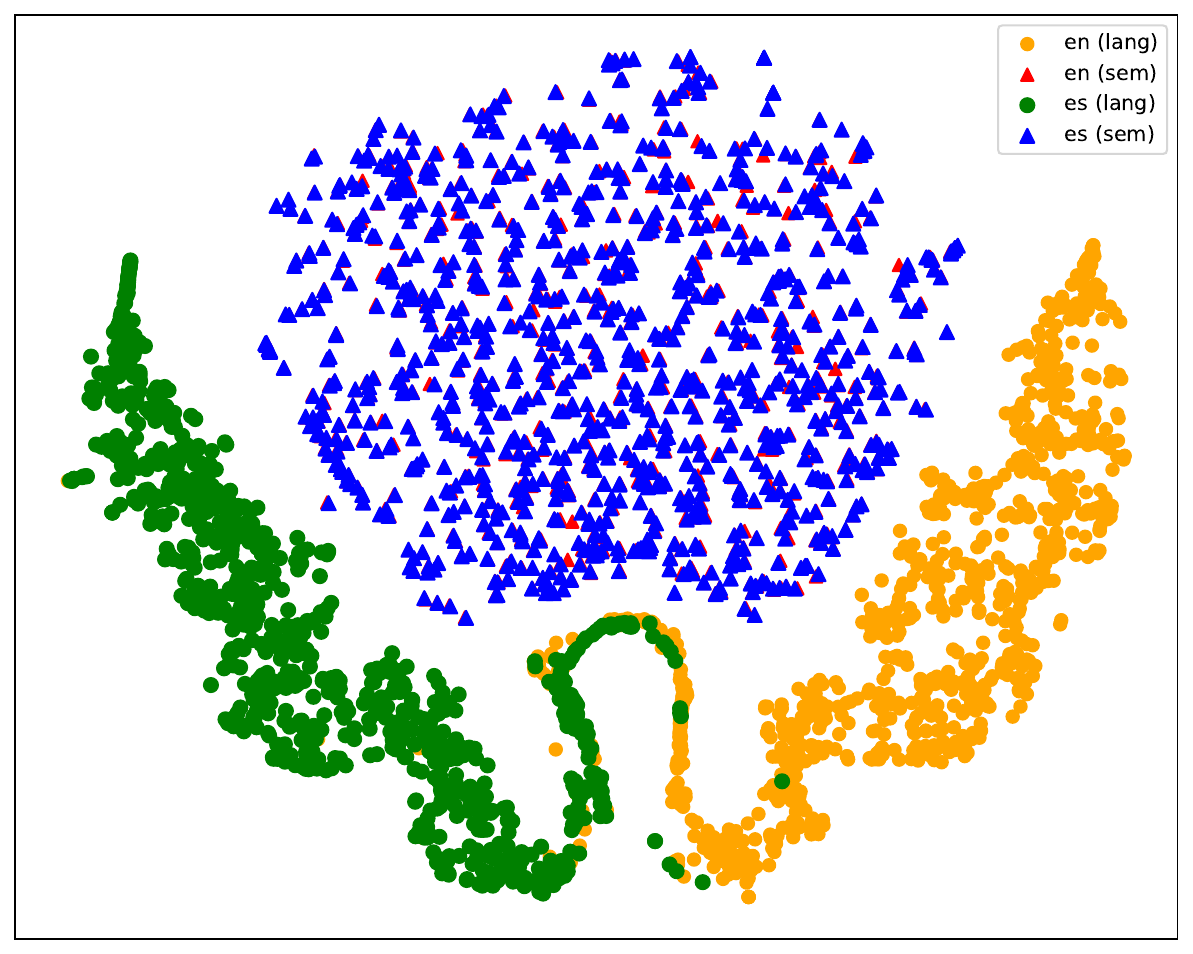}
  \caption{DREAM + \textsc{ORACLE}}
\end{subfigure}
\begin{subfigure}{.24\textwidth}
  \centering
  \includegraphics[width=\linewidth]{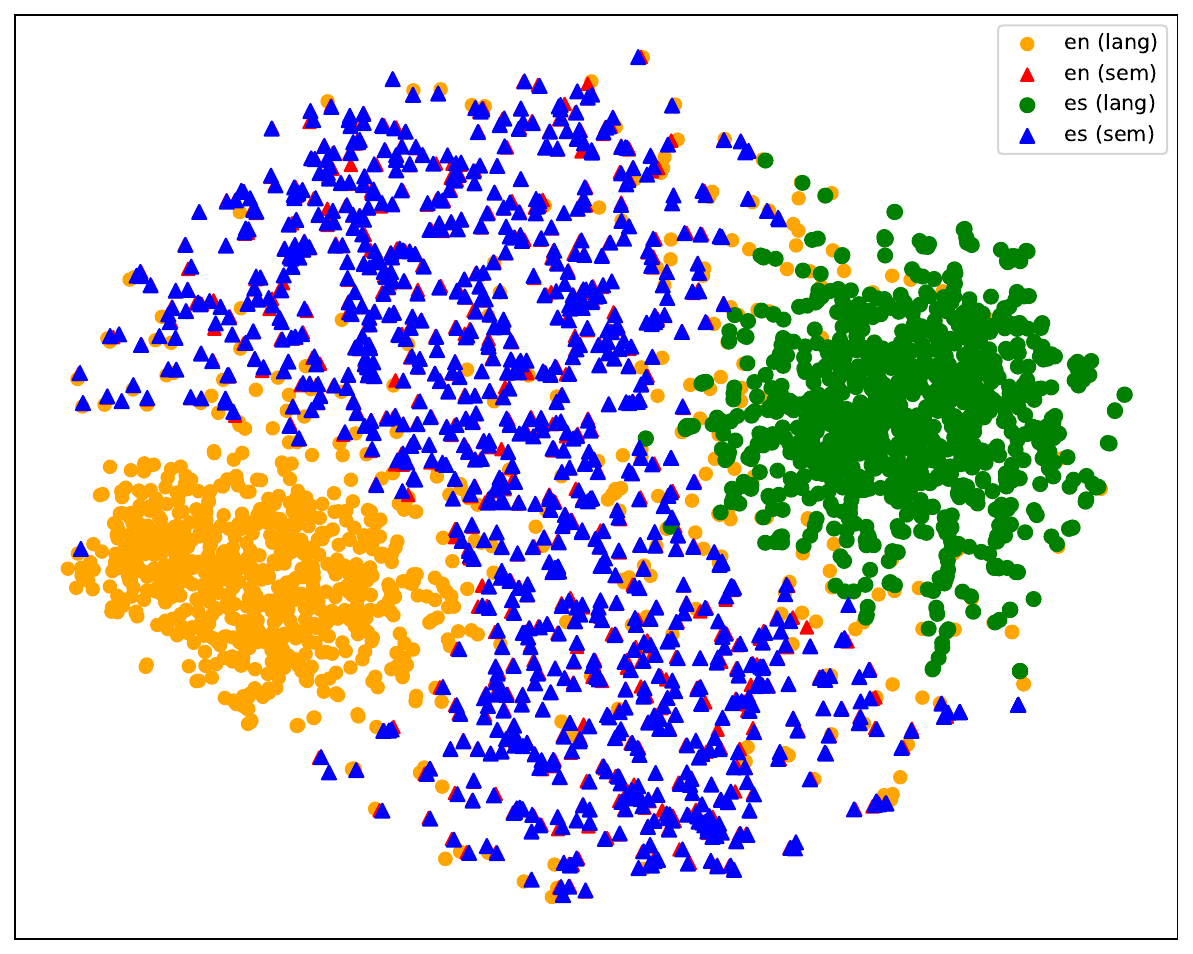}
  \caption{MEAT}
\end{subfigure}%
\begin{subfigure}{.24\textwidth}
  \centering
  \includegraphics[width=\linewidth]{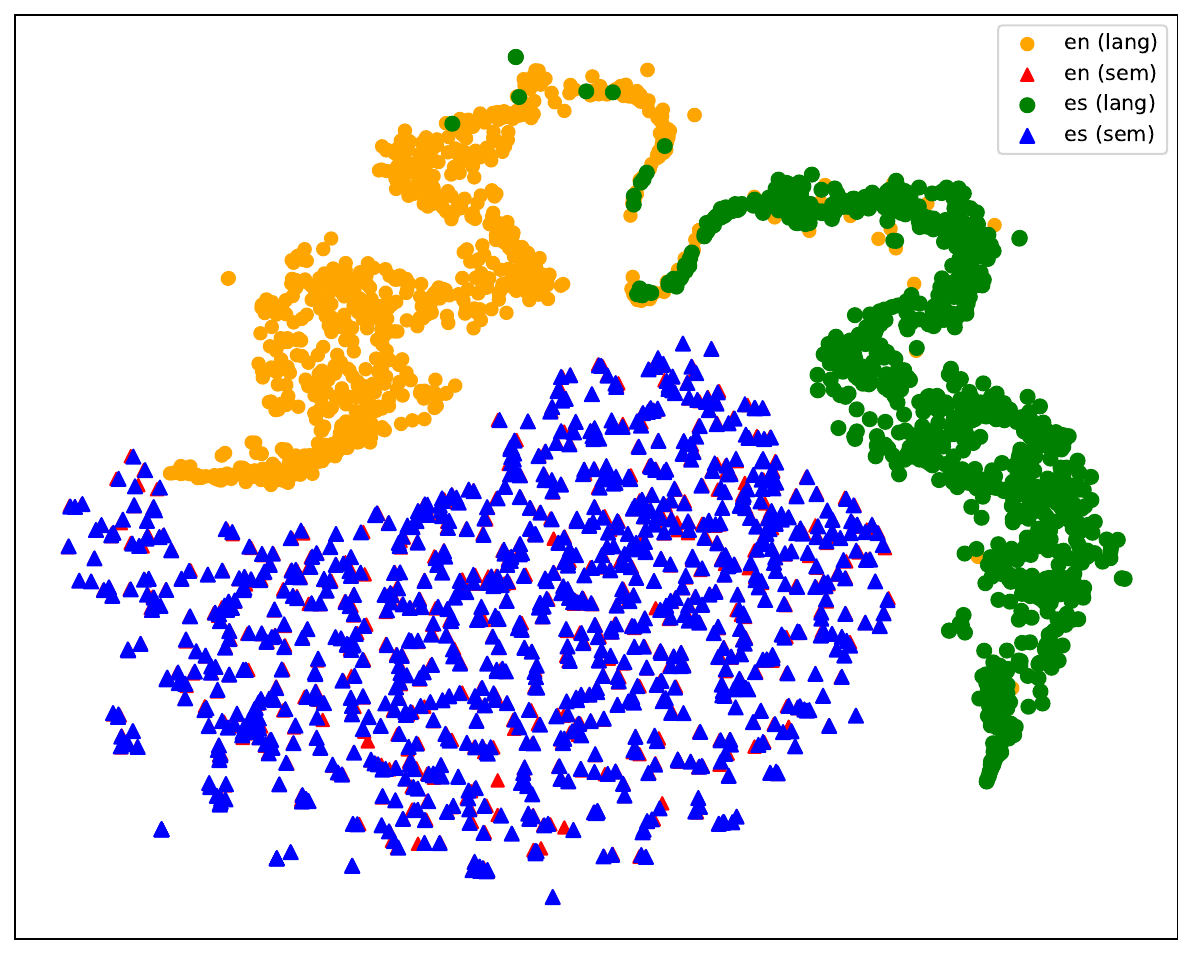}
  \caption{MEAT + \textsc{ORACLE}}
\end{subfigure}

\caption{LaBSE sentence embeddings for English-Spanish sentence pair.}
\label{fig:enes}
\end{figure*}

\begin{figure*}[]
\centering

\begin{subfigure}{.24\textwidth}
  \centering
  \includegraphics[width=\linewidth]{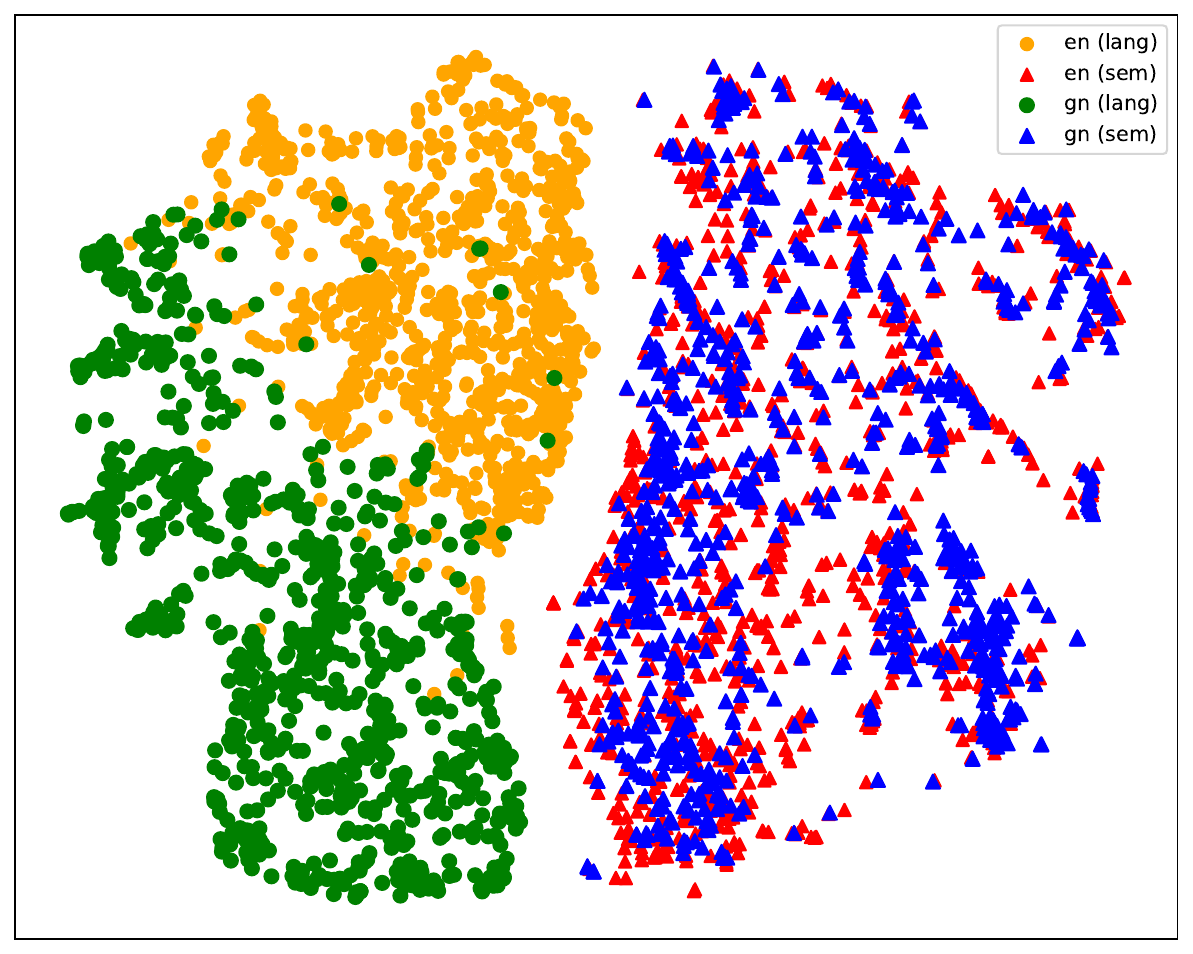}
  \caption{DREAM}
\end{subfigure}%
\begin{subfigure}{.24\textwidth}
  \centering
  \includegraphics[width=\linewidth]{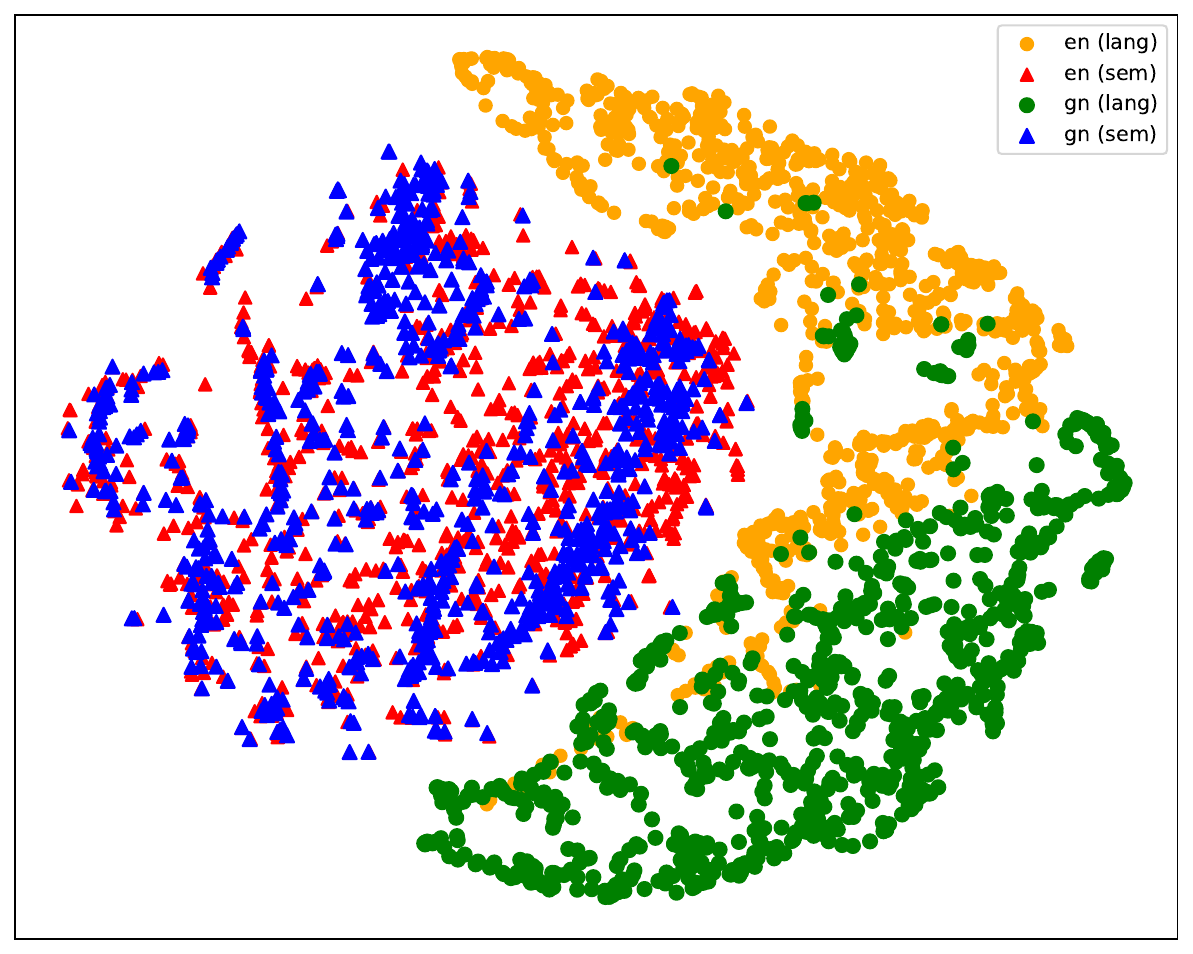}
  \caption{DREAM + \textsc{ORACLE}}
\end{subfigure}
\begin{subfigure}{.24\textwidth}
  \centering
  \includegraphics[width=\linewidth]{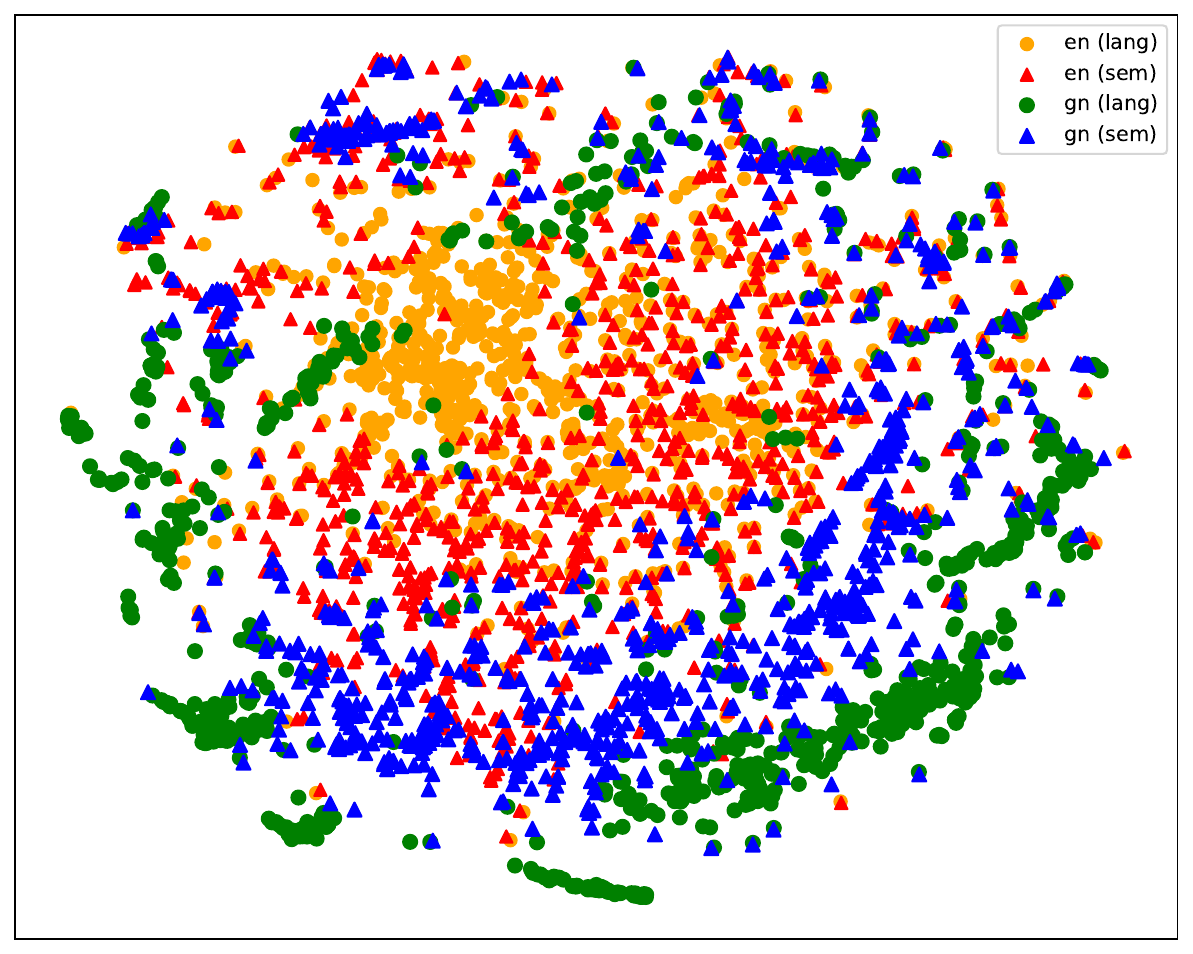}
  \caption{MEAT}
\end{subfigure}%
\begin{subfigure}{.24\textwidth}
  \centering
  \includegraphics[width=\linewidth]{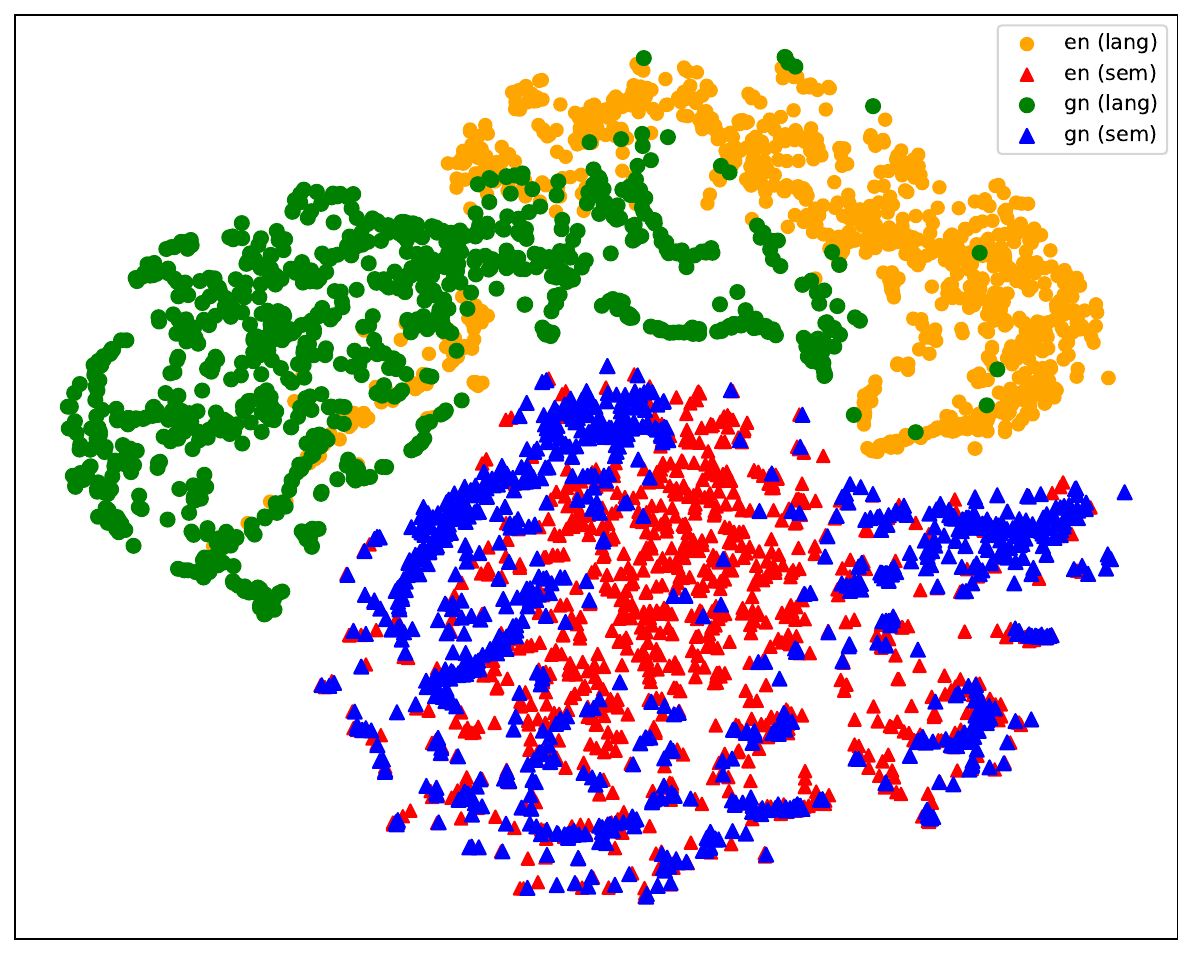}
  \caption{MEAT + \textsc{ORACLE}}
\end{subfigure}

\caption{LaBSE sentence embeddings for English-Guaraní sentence pair.}
\label{fig:engn}
\end{figure*}

\begin{figure*}[]
\centering

\begin{subfigure}{.24\textwidth}
  \centering
  \includegraphics[width=\linewidth]{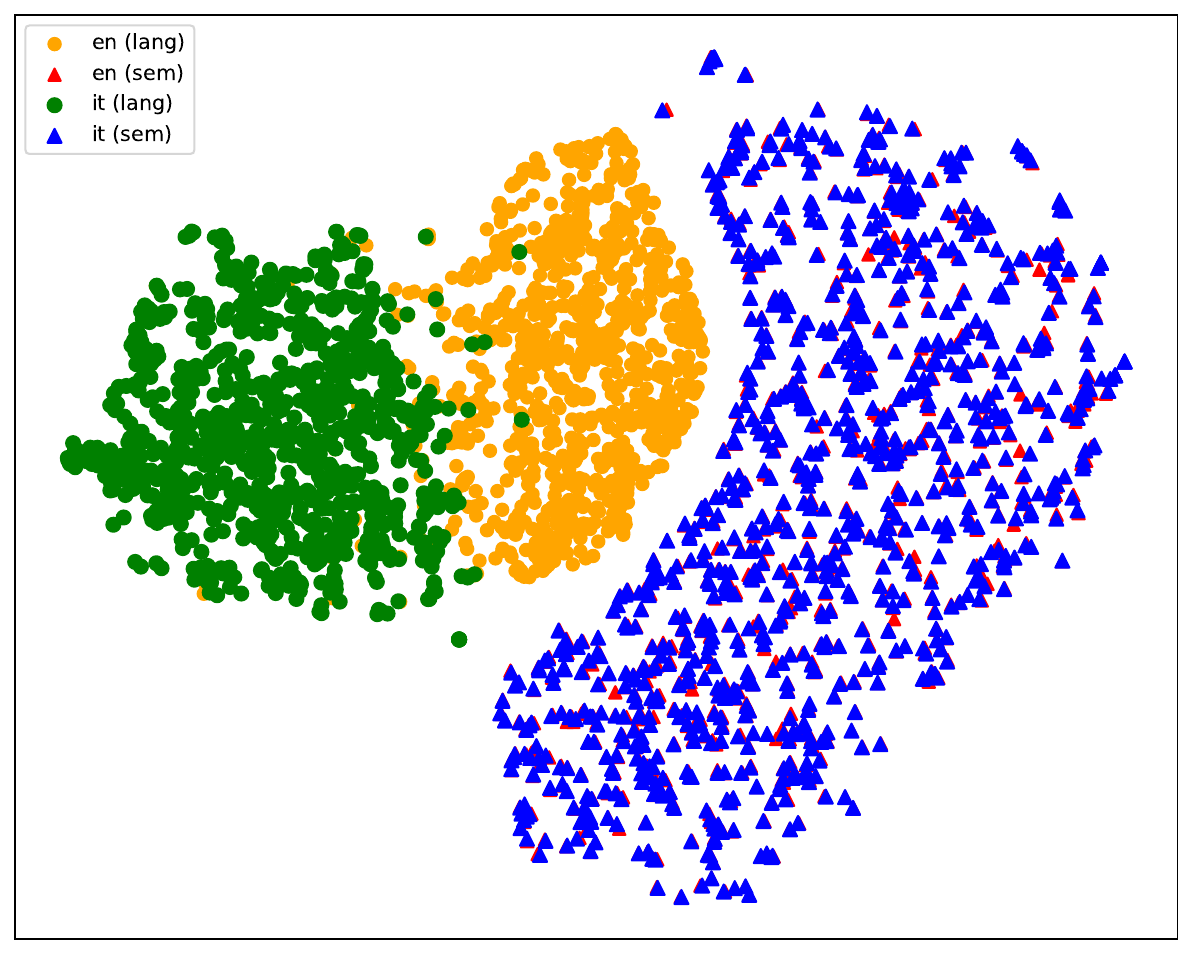}
  \caption{DREAM}
\end{subfigure}%
\begin{subfigure}{.24\textwidth}
  \centering
  \includegraphics[width=\linewidth]{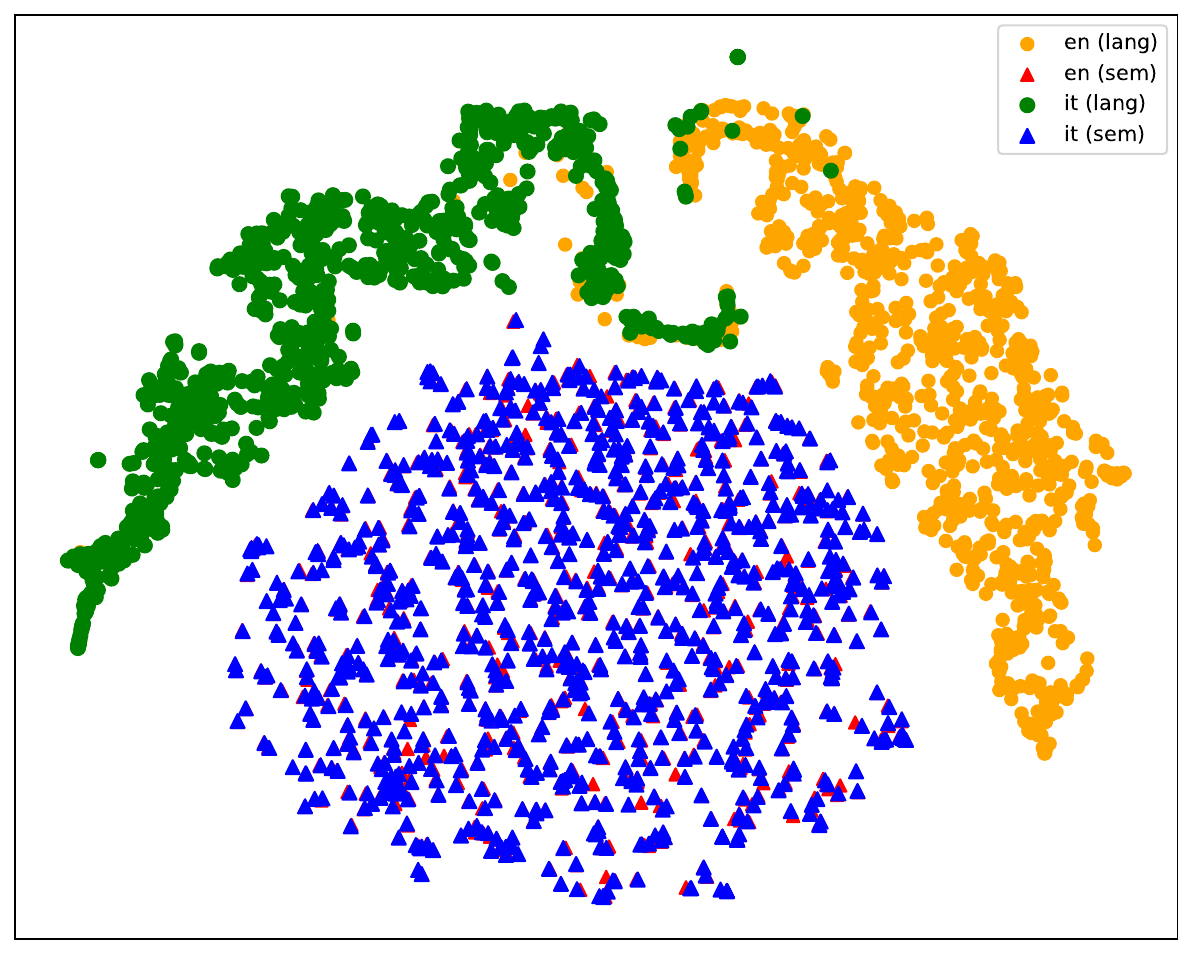}
  \caption{DREAM + \textsc{ORACLE}}
\end{subfigure}
\begin{subfigure}{.24\textwidth}
  \centering
  \includegraphics[width=\linewidth]{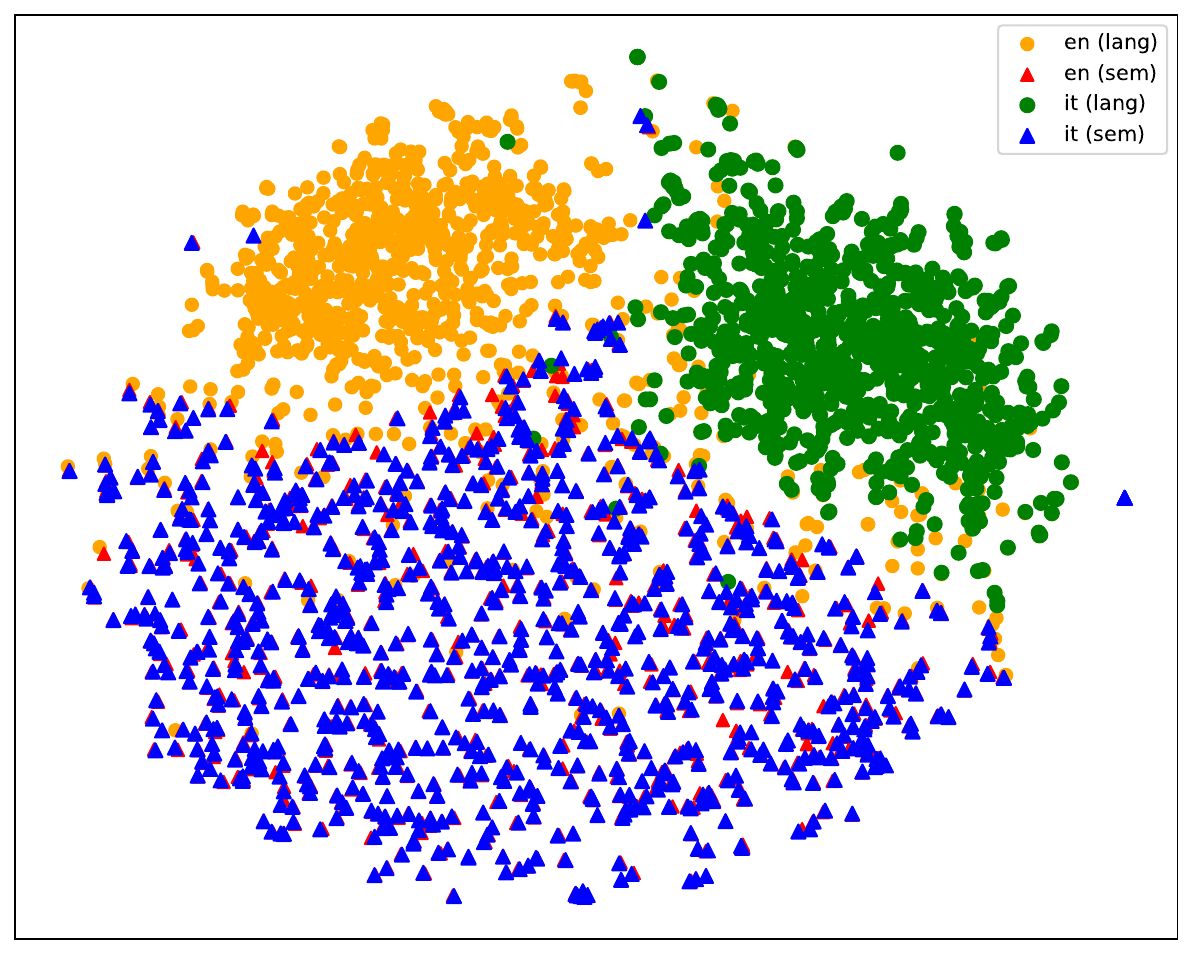}
  \caption{MEAT}
\end{subfigure}%
\begin{subfigure}{.24\textwidth}
  \centering
  \includegraphics[width=\linewidth]{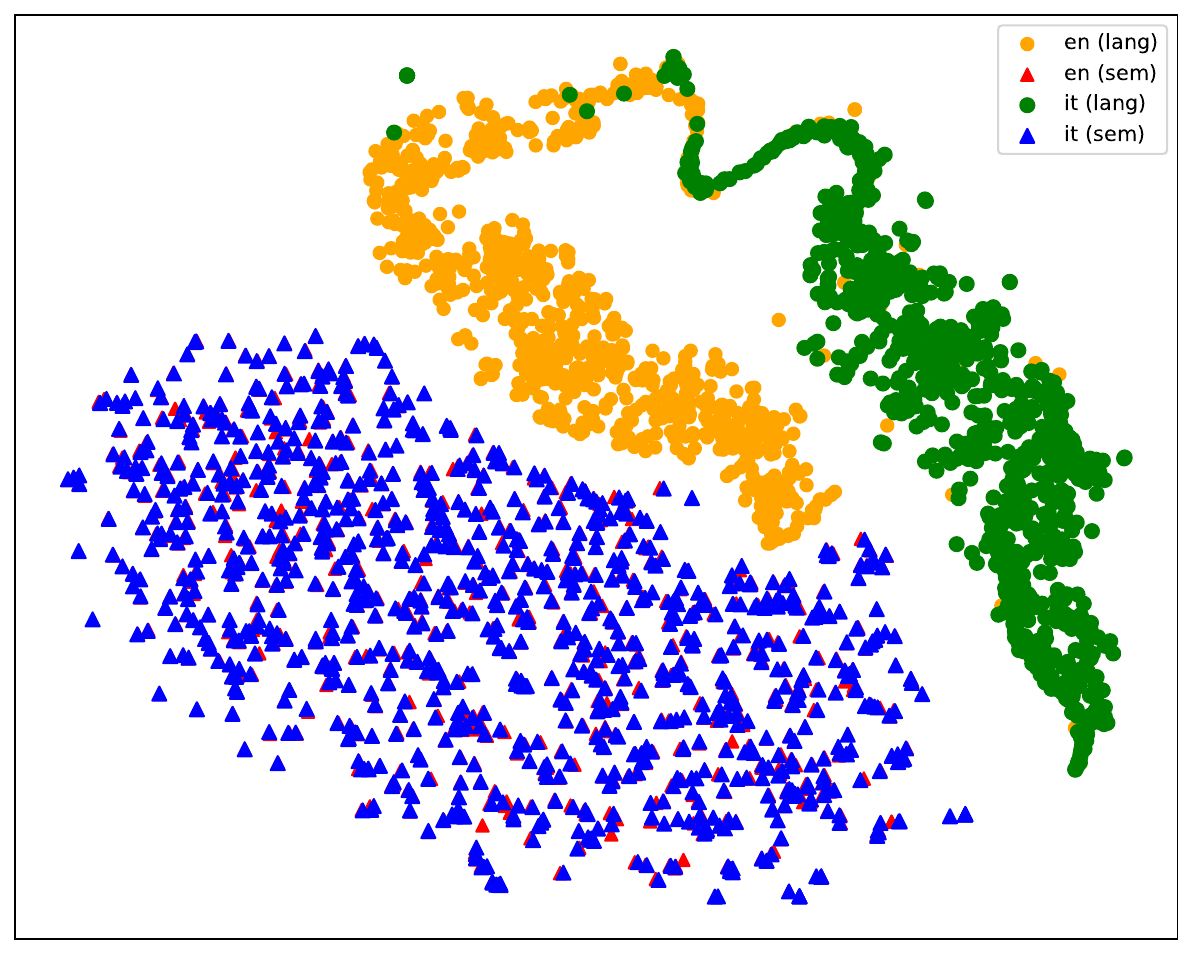}
  \caption{MEAT + \textsc{ORACLE}}
\end{subfigure}

\caption{LaBSE sentence embeddings for English-Italian sentence pair.}
\label{fig:enit}
\end{figure*}

\begin{figure*}[]
\centering

\begin{subfigure}{.24\textwidth}
  \centering
  \includegraphics[width=\linewidth]{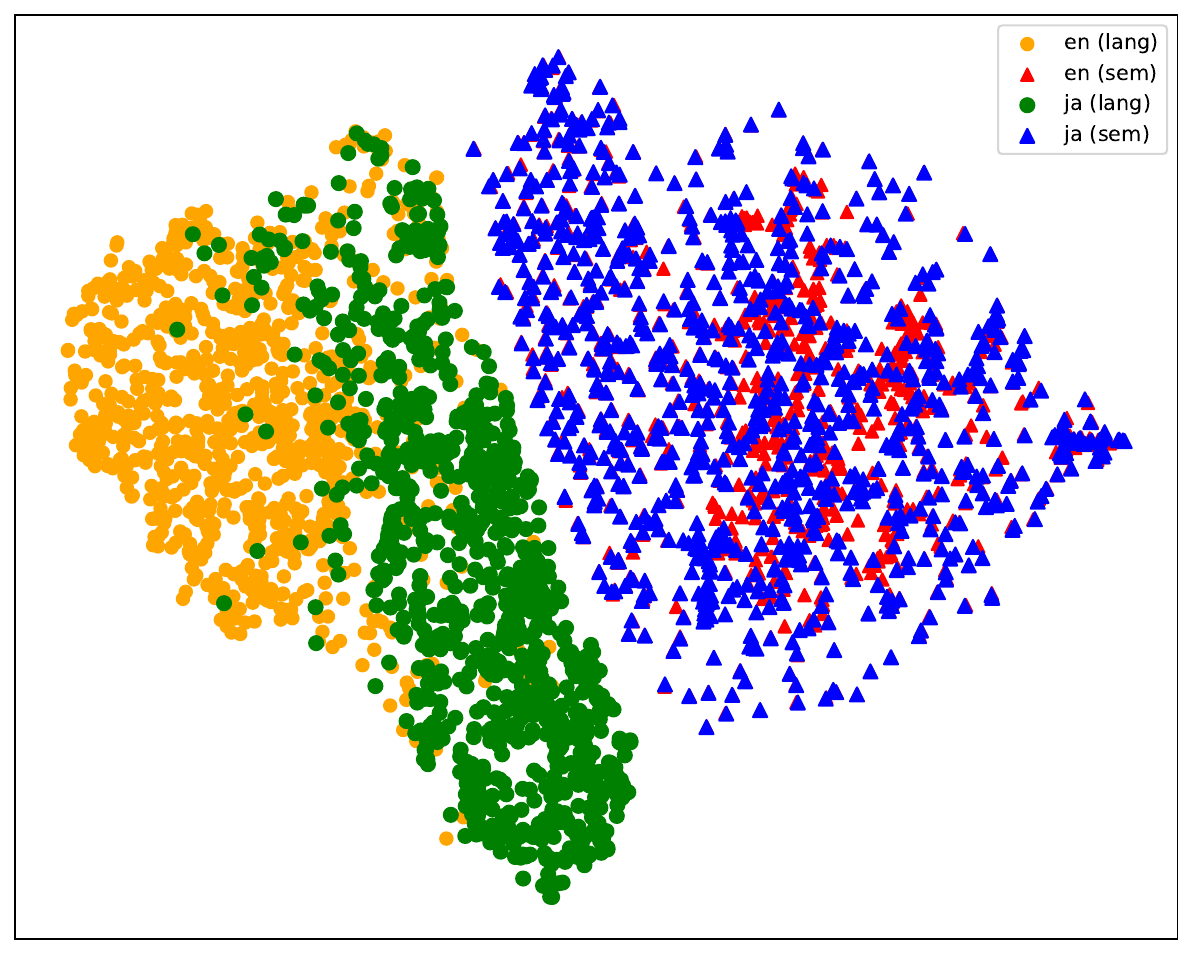}
  \caption{DREAM}
\end{subfigure}%
\begin{subfigure}{.24\textwidth}
  \centering
  \includegraphics[width=\linewidth]{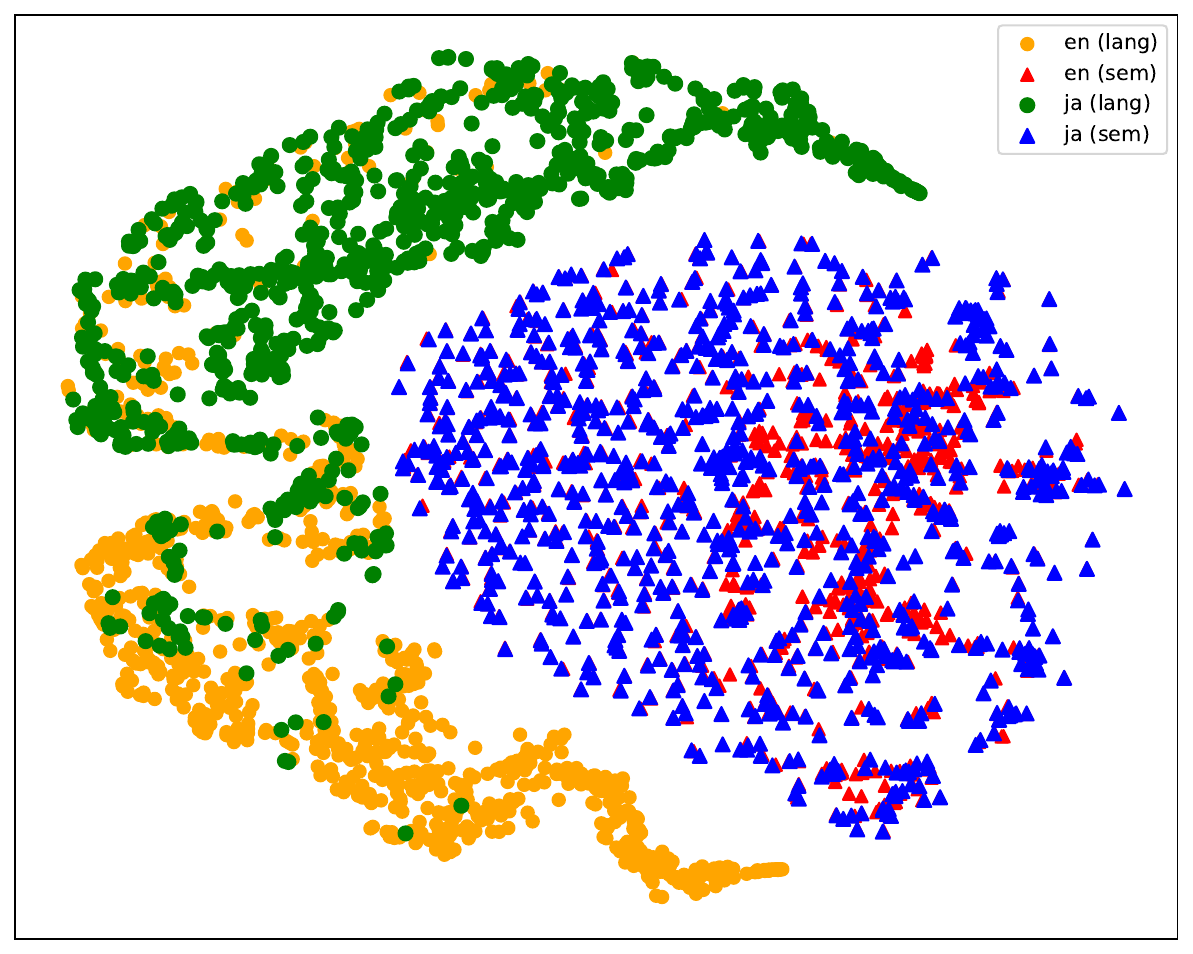}
  \caption{DREAM + \textsc{ORACLE}}
\end{subfigure}
\begin{subfigure}{.24\textwidth}
  \centering
  \includegraphics[width=\linewidth]{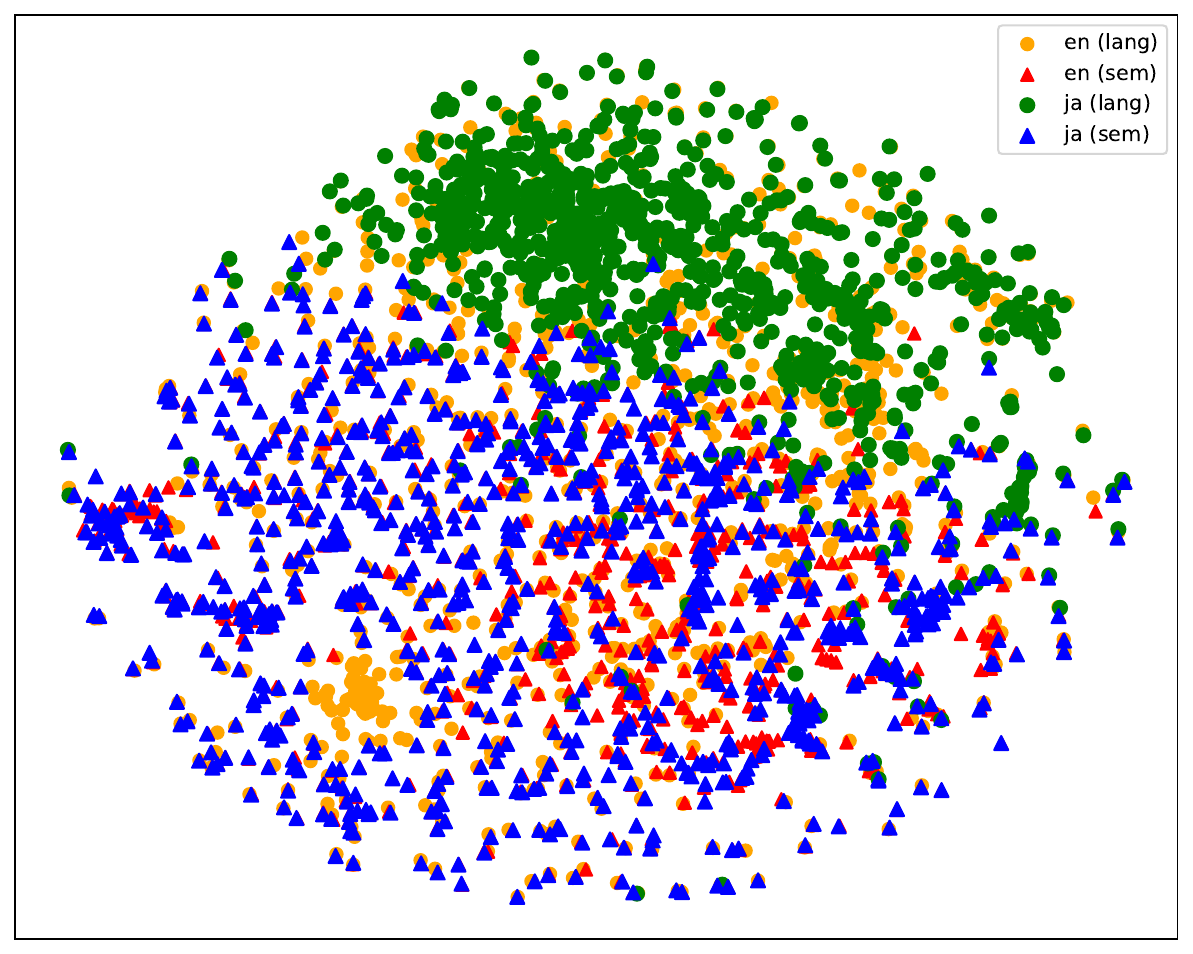}
  \caption{MEAT}
\end{subfigure}%
\begin{subfigure}{.24\textwidth}
  \centering
  \includegraphics[width=\linewidth]{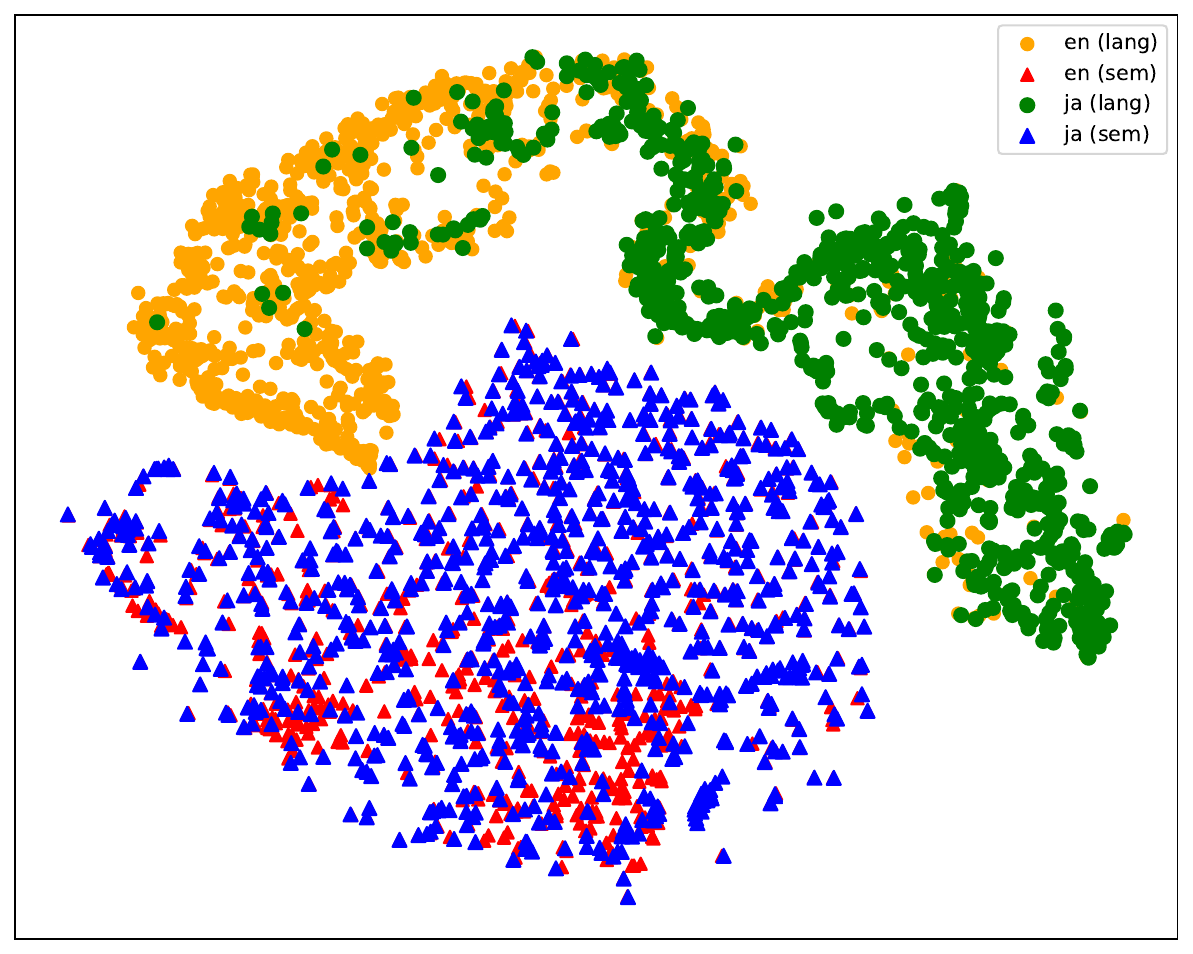}
  \caption{MEAT + \textsc{ORACLE}}
\end{subfigure}

\caption{LaBSE sentence embeddings for English-Japanese sentence pair.}
\label{fig:enja}
\end{figure*}

\begin{figure*}[]
\centering

\begin{subfigure}{.24\textwidth}
  \centering
  \includegraphics[width=\linewidth]{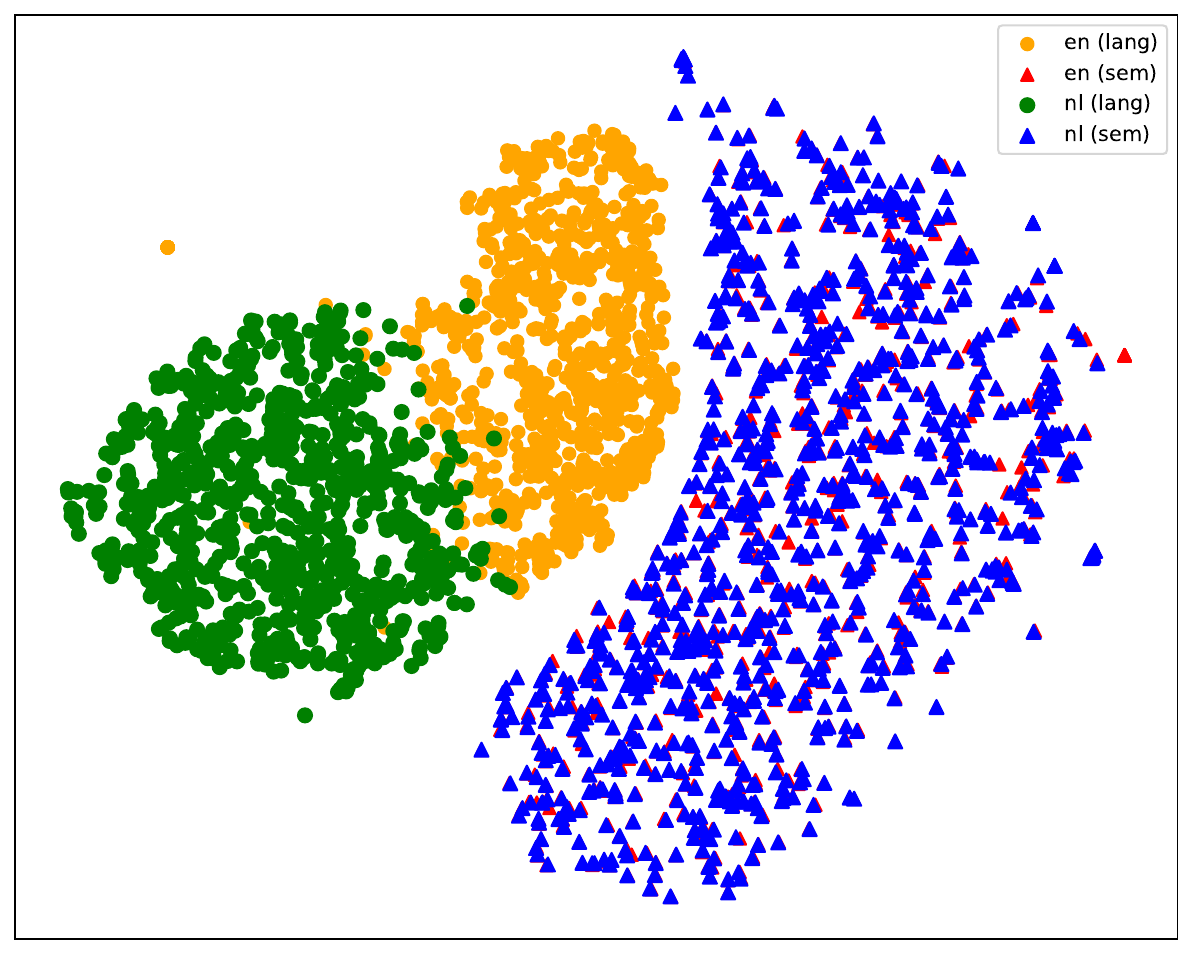}
  \caption{DREAM}
\end{subfigure}%
\begin{subfigure}{.24\textwidth}
  \centering
  \includegraphics[width=\linewidth]{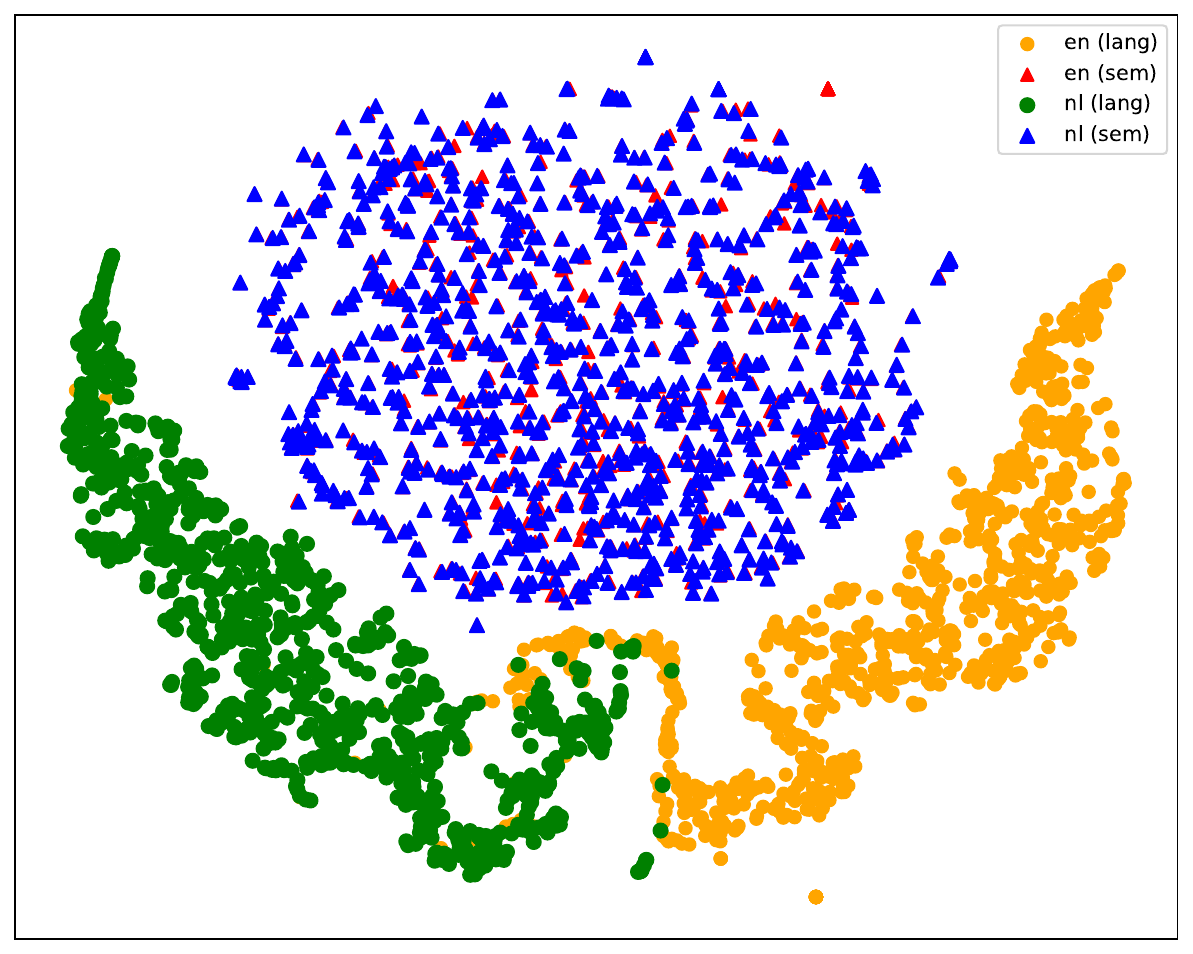}
  \caption{DREAM + \textsc{ORACLE}}
\end{subfigure}
\begin{subfigure}{.24\textwidth}
  \centering
  \includegraphics[width=\linewidth]{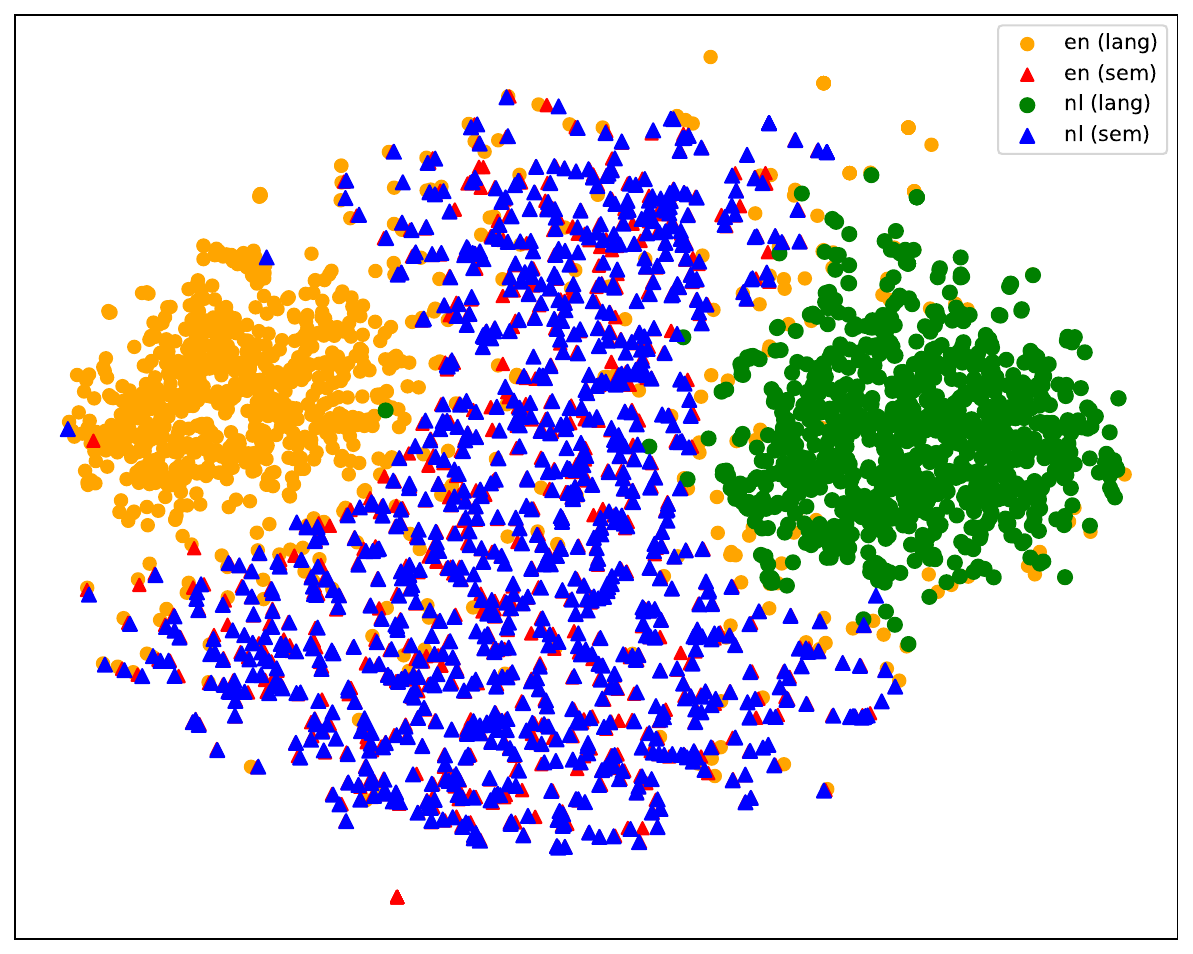}
  \caption{MEAT}
\end{subfigure}%
\begin{subfigure}{.24\textwidth}
  \centering
  \includegraphics[width=\linewidth]{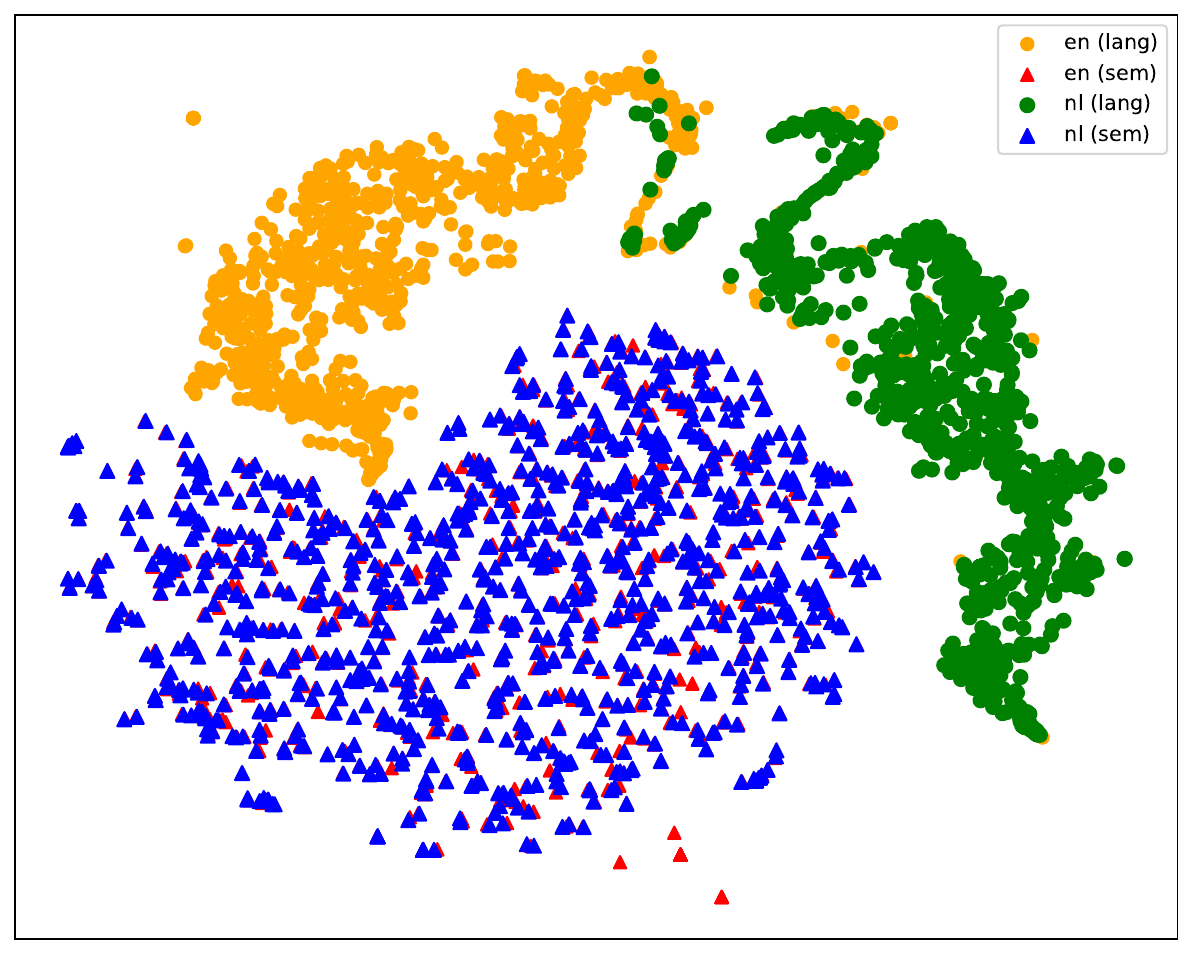}
  \caption{MEAT + \textsc{ORACLE}}
\end{subfigure}

\caption{LaBSE sentence embeddings for English-Dutch sentence pair.}
\label{fig:ennl}
\end{figure*}

\begin{figure*}[]
\centering

\begin{subfigure}{.24\textwidth}
  \centering
  \includegraphics[width=\linewidth]{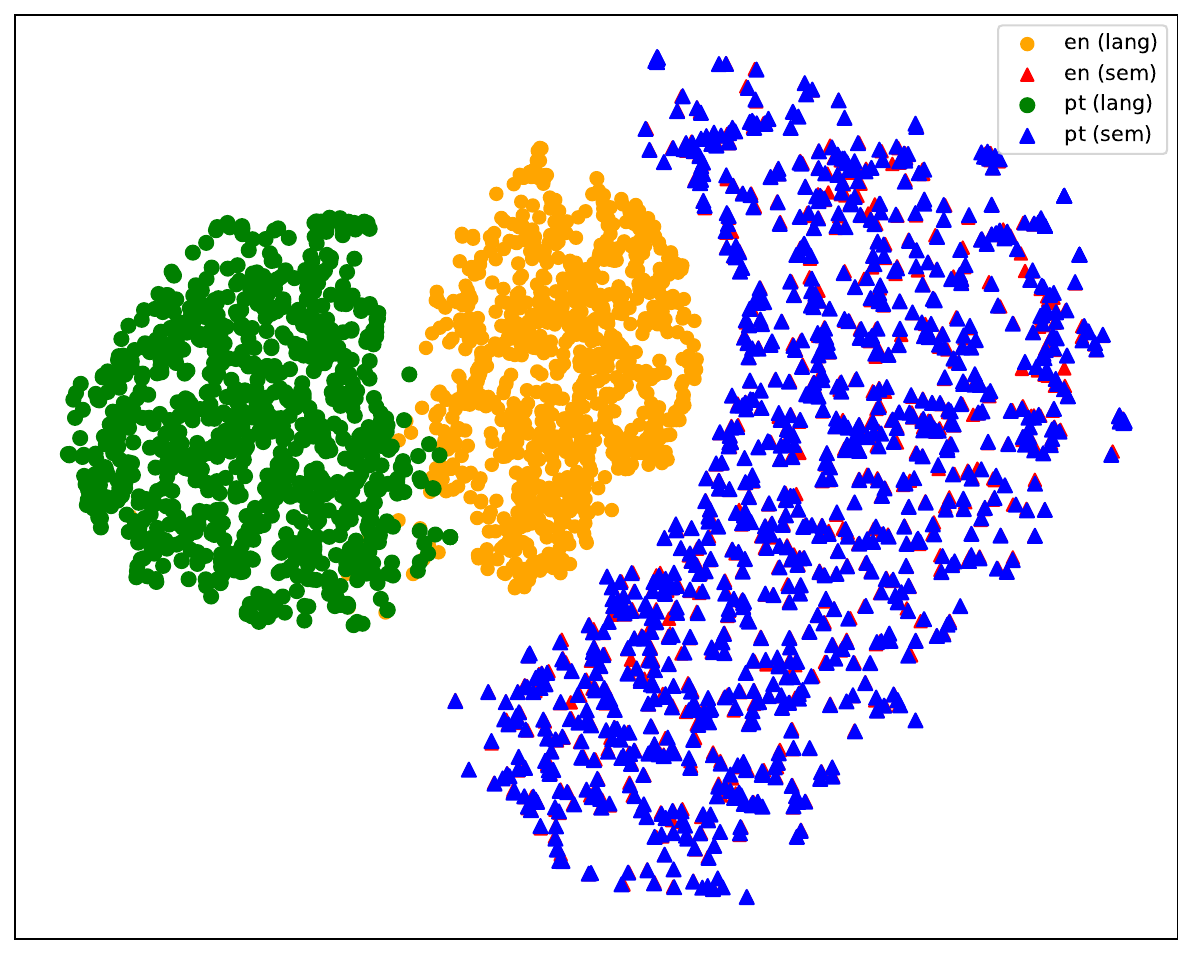}
  \caption{DREAM}
\end{subfigure}%
\begin{subfigure}{.24\textwidth}
  \centering
  \includegraphics[width=\linewidth]{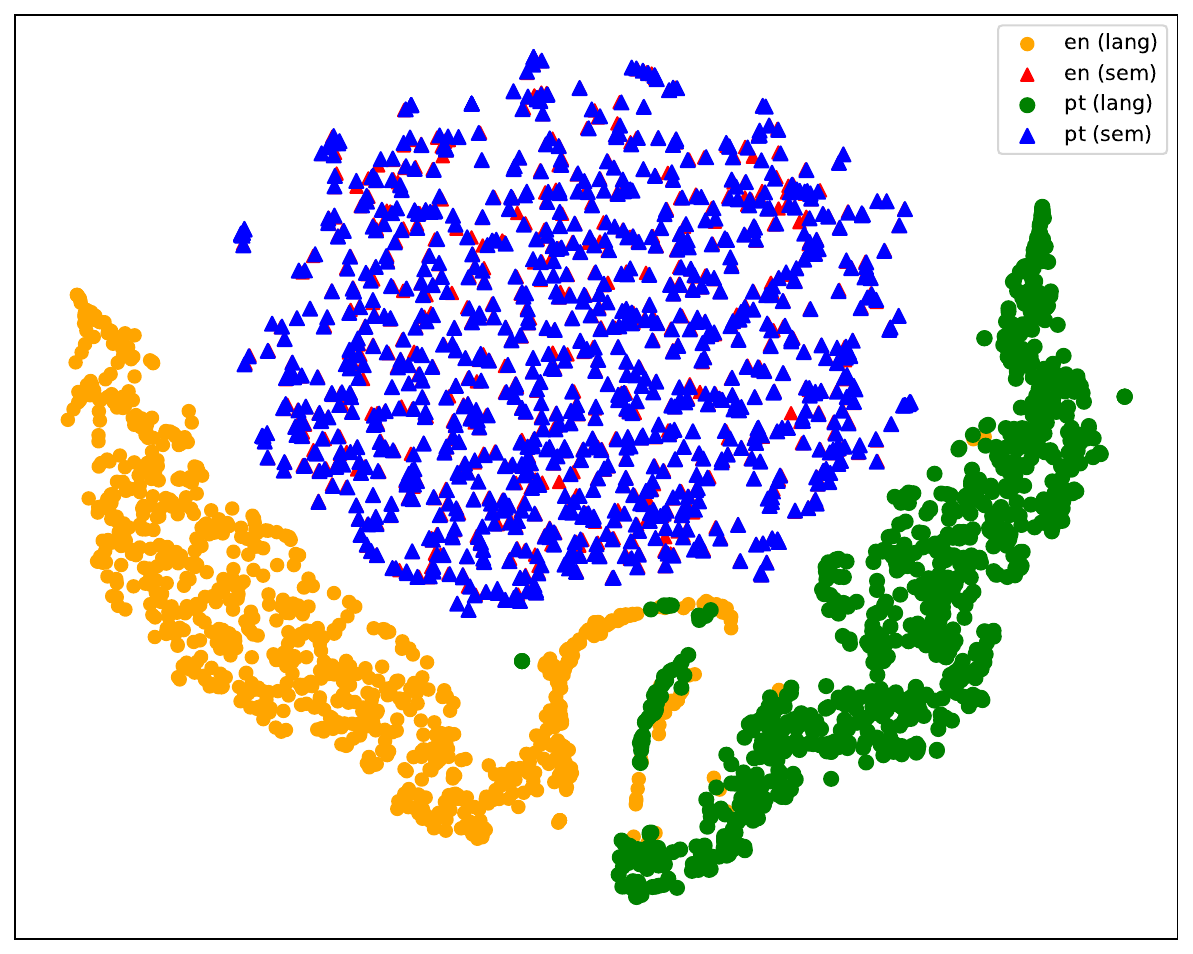}
  \caption{DREAM + \textsc{ORACLE}}
\end{subfigure}
\begin{subfigure}{.24\textwidth}
  \centering
  \includegraphics[width=\linewidth]{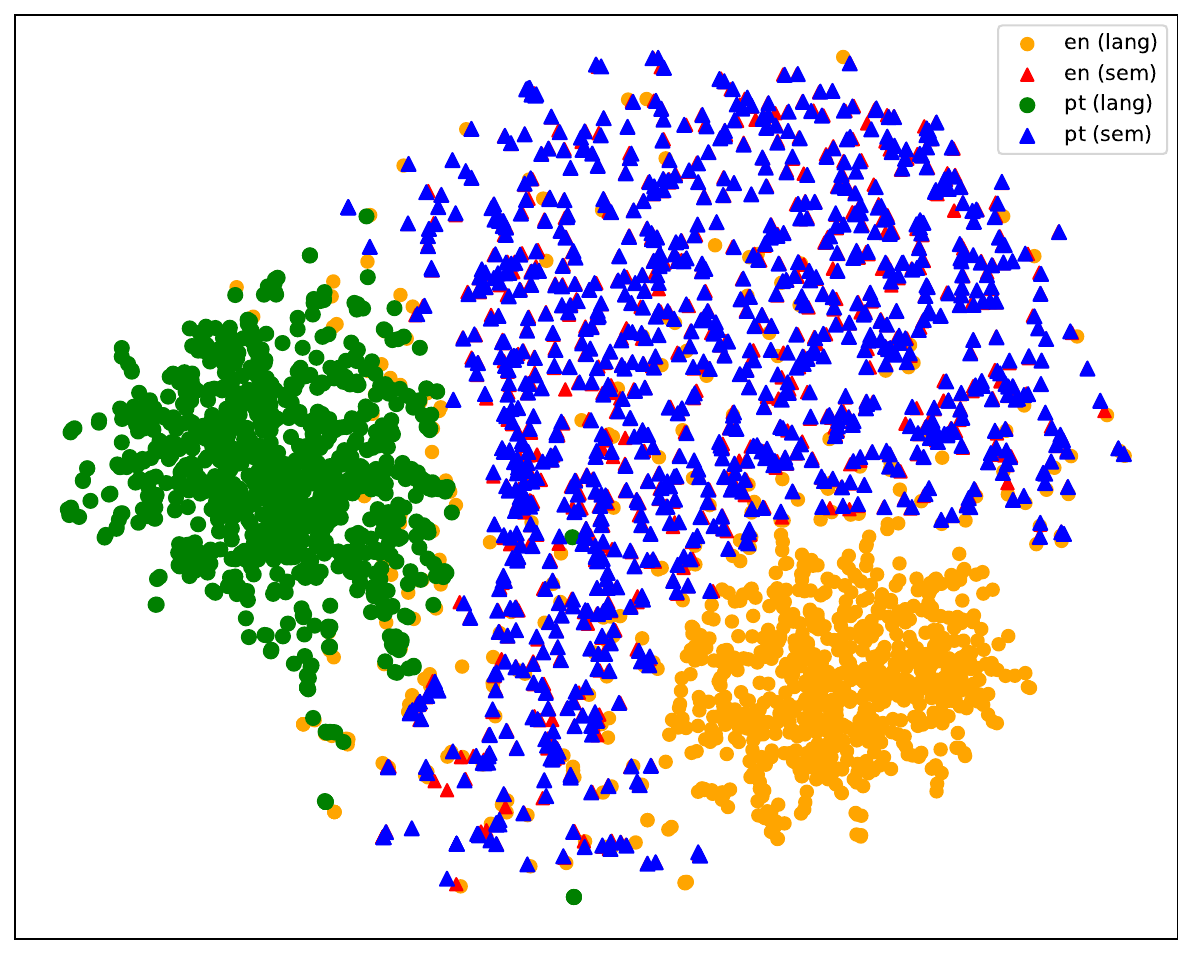}
  \caption{MEAT}
\end{subfigure}%
\begin{subfigure}{.24\textwidth}
  \centering
  \includegraphics[width=\linewidth]{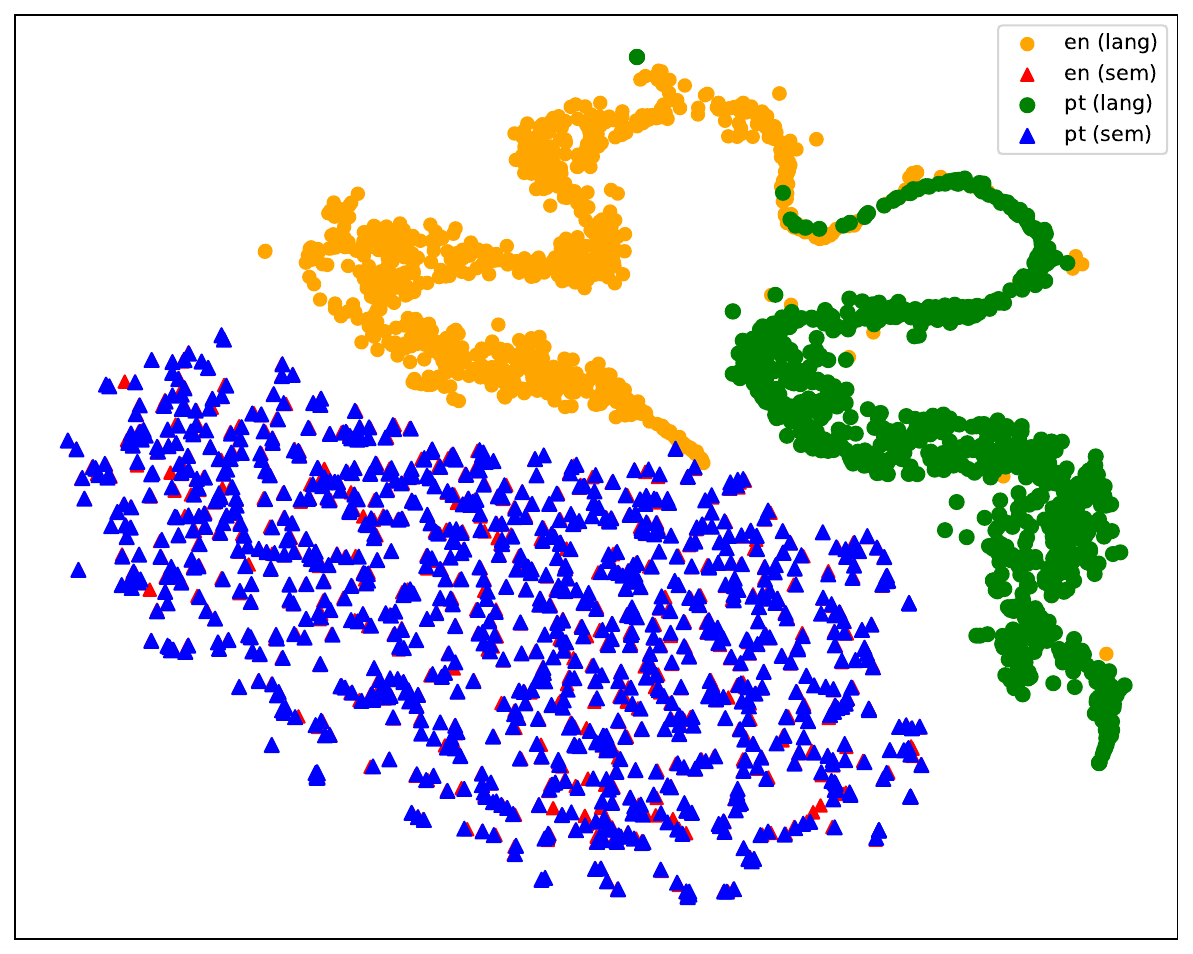}
  \caption{MEAT + \textsc{ORACLE}}
\end{subfigure}

\caption{LaBSE sentence embeddings for English-Portuguese sentence pair.}
\label{fig:enpt}
\end{figure*}

\begin{figure*}[]
\centering

\begin{subfigure}{.24\textwidth}
  \centering
  \includegraphics[width=\linewidth]{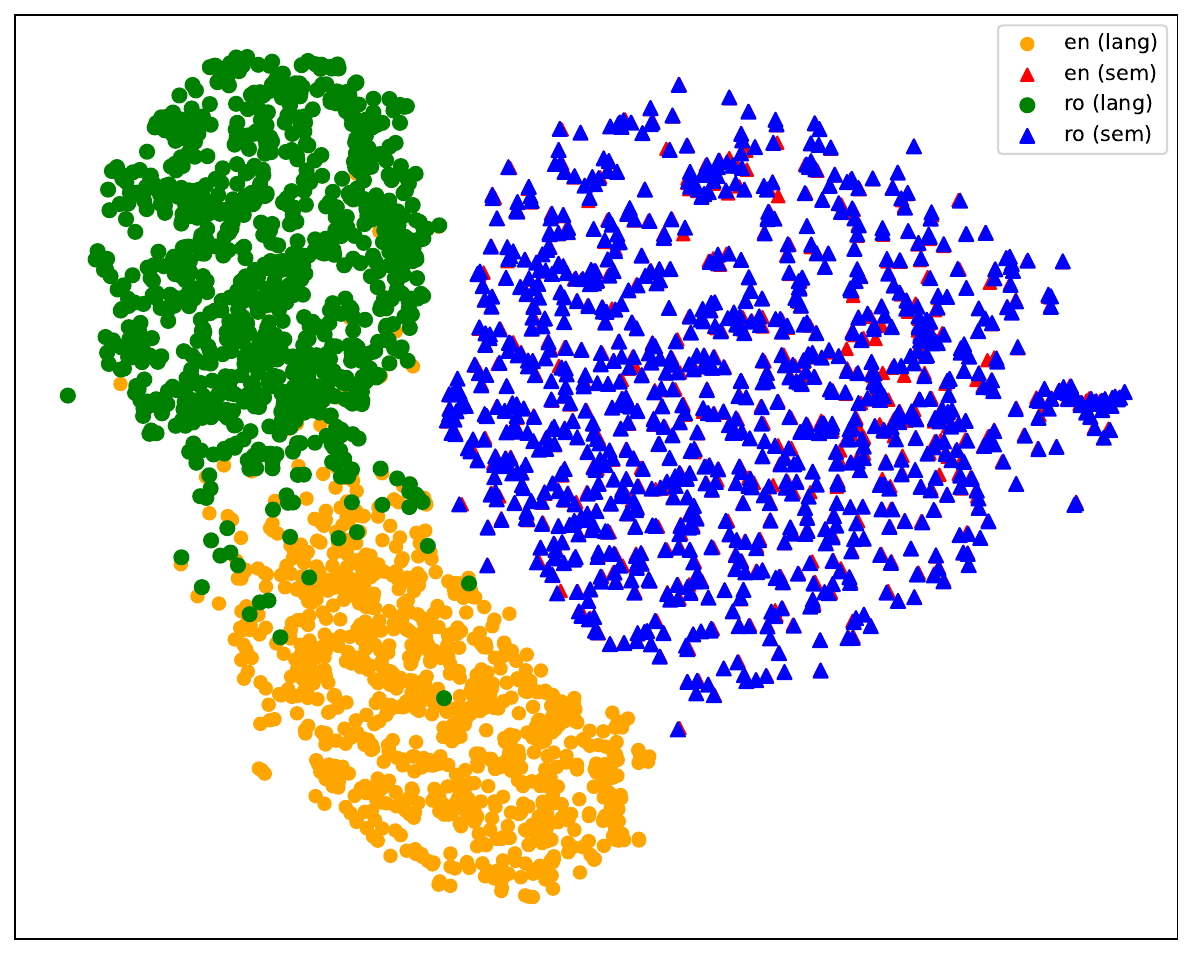}
  \caption{DREAM}
\end{subfigure}%
\begin{subfigure}{.24\textwidth}
  \centering
  \includegraphics[width=\linewidth]{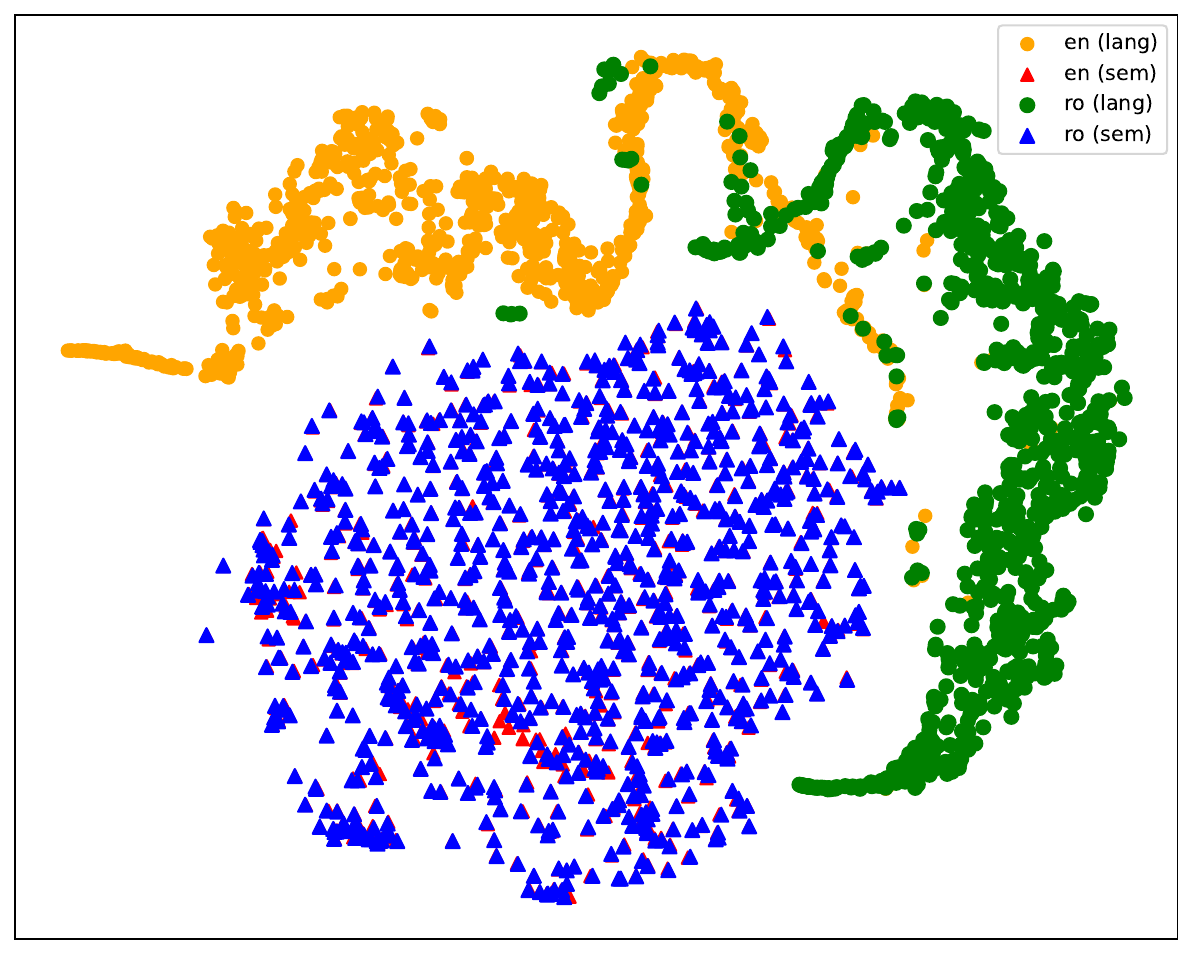}
  \caption{DREAM + \textsc{ORACLE}}
\end{subfigure}
\begin{subfigure}{.24\textwidth}
  \centering
  \includegraphics[width=\linewidth]{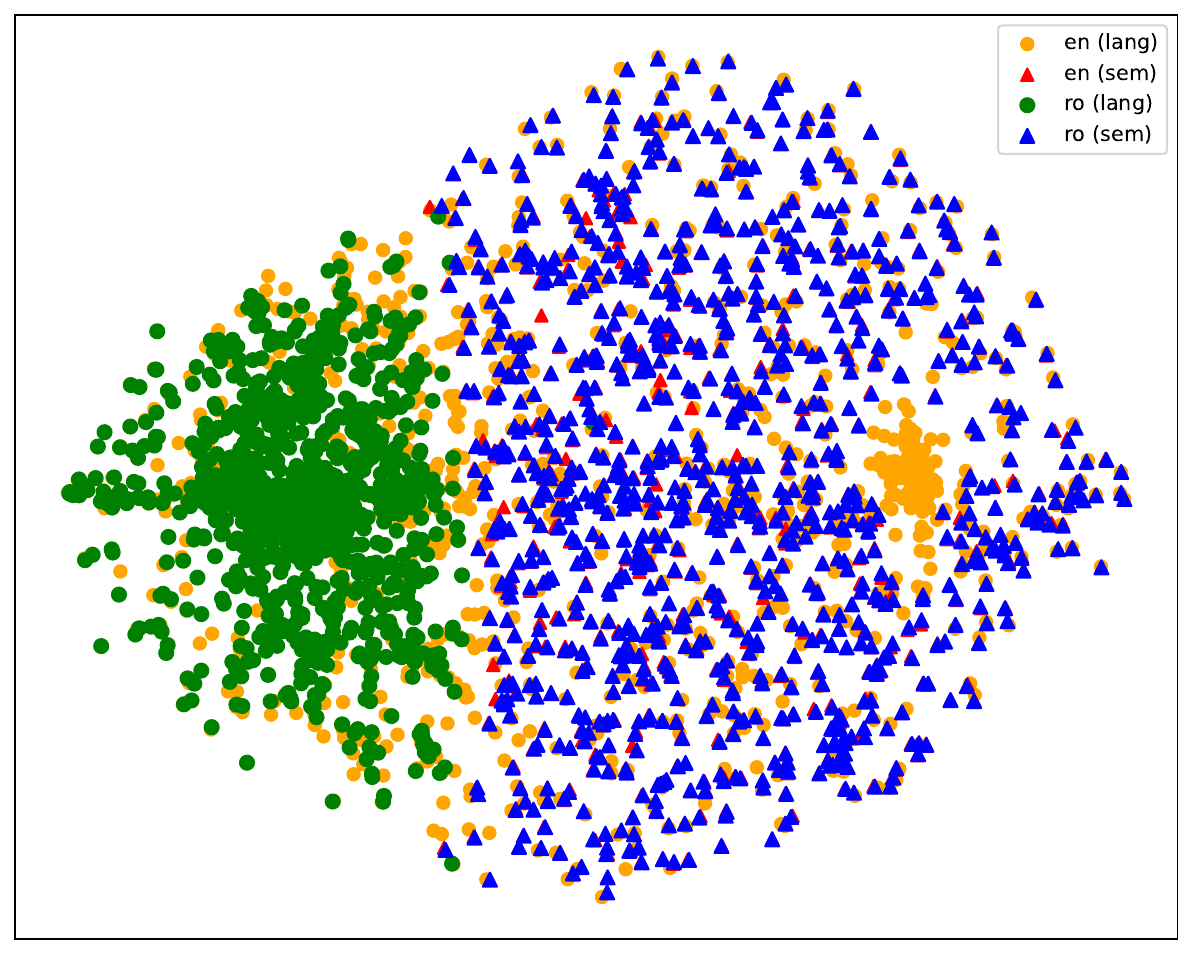}
  \caption{MEAT}
\end{subfigure}%
\begin{subfigure}{.24\textwidth}
  \centering
  \includegraphics[width=\linewidth]{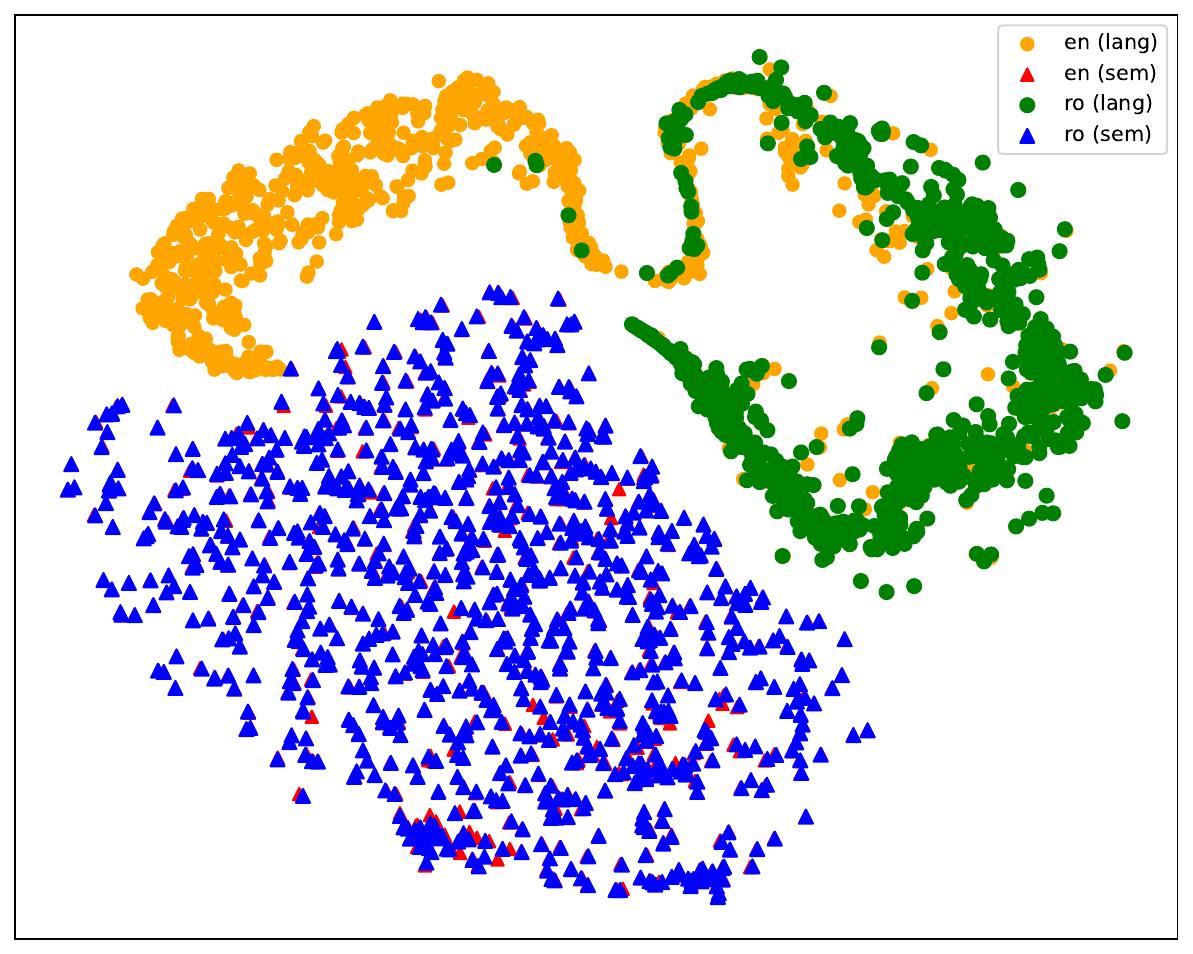}
  \caption{MEAT + \textsc{ORACLE}}
\end{subfigure}

\caption{LaBSE sentence embeddings for English-Romanian sentence pair.}
\label{fig:enro}
\end{figure*}

From Figures \ref{fig:enar} to \ref{fig:enro}, we provide visualizations of semantic and language embeddings for each language pair, complementing the discussion in Section \ref{sec:visualization}. We use LaBSE to generate the initial sentence embeddings, with 1,000 parallel sentences sampled from our held-out test set for each language pair. When solely using DREAM or MEAT (depicted in (a) and (c) for each visualization), we observe a notable amount of overlap in language embeddings between the source and target language, indicating semantic leakage. However, the integration of \textsc{ORACLE} effectively mitigates this issue, resulting in clearer separation and reduced overlap in language embeddings (depicted in (b) and (d)). This improvement is consistent across all language pairs.

\definecolor{light yellow}{rgb}{0.984, 0.996, 0.753}

\begin{table*}[!h]
\centering
\resizebox{\textwidth}{!}{%
    \begin{tabular}{l l c c c c c c c c c c c c c c}
    \toprule
    \textbf{Encoder} & \textbf{Objective} & \textbf{en-ar} & \textbf{en-ay} & \textbf{en-de} & \textbf{en-es} & \textbf{en-fr} & \textbf{en-gn} & \textbf{en-it} & \textbf{en-ja} & \textbf{en-nl} & \textbf{en-pt} & \textbf{en-ro} & \textbf{en-zh} & \textbf{Avg.} \\ \midrule\midrule

    \multicolumn{15}{c}{\textit{Original Embedding}} \\ \midrule
    
     LASER & - & 99.87 & 11.36 & 96.13 & 97.88 & 93.38 & 5.12 & 96.78 & 98.86 & 96.48 & 97.83 & 99.34 & 99.39 & 82.70 \\ 
     InfoXLM$^*$ & - & 21.24 & 1.40 & 25.53 & 29.31 & 28.41 & 0.69 & 29.07 & 10.48 & 14.72 & 23.80 & 15.59 & 13.06 & 17.78 \\ 
     LaBSE & - & 99.00 & 16.33 & 96.16 & 97.82 & 93.37 & 12.82 & 96.77 & 92.30 & 96.21 & 97.66 & 87.60 & 94.05 & 81.67 \\ \midrule\midrule

     \multicolumn{15}{c}{\textit{Semantic Embedding} (↑)} \\ \midrule

      \multirow{4}{*}{LASER} & DREAM & 94.22 & 7.27 & 93.85 & 96.81 & 92.28 & 4.33 & 95.13 & 82.94 & 93.46 & 96.39 & 77.05 & 87.10 & 76.74 \\
       & +\textsc{ORACLE} & 94.10 & 7.23 & \textbf{93.87} & 96.79 & 92.28 & 4.26 & 95.13 & \textbf{82.95} & \textbf{93.54} & 96.39 & \textbf{77.27} & \textbf{87.13} & \fcolorbox{white}{light yellow}{\textbf{76.75}} \\
       
       & MEAT & 99.58 & 13.50 & 95.91 & 97.71 & 93.18 & 8.73 & 96.29 & 97.20 & 95.95 & 97.54 & 95.82 & 98.42 & 82.49 \\
      & +\textsc{ORACLE} & \textbf{99.78} & 11.11 & \textbf{95.92} & 97.71 & \textbf{93.21} & \textbf{8.91} & \textbf{96.33} & \textbf{97.57} & \textbf{96.02} & \textbf{97.67} & \textbf{98.16} & \textbf{98.54} & \fcolorbox{white}{light yellow}{\textbf{82.58}} \\ \midrule
      
    \multirow{4}{*}{InfoXLM} & DREAM & 88.74 & 2.06 & 88.61 & 94.89 & 90.47 & 1.80 & 91.14 & 68.71 & 87.44 & 93.06 & 64.87 & 74.96 & 70.56 \\
       & +\textsc{ORACLE} & \textbf{89.11} & \textbf{2.89} & \textbf{89.20} & \textbf{95.00} & \textbf{90.50} & \textbf{1.96} & \textbf{91.49} & \textbf{69.21} & \textbf{87.85} & \textbf{93.27} & \textbf{65.23} & \textbf{75.46} & \fcolorbox{white}{light yellow}{\textbf{70.93}} \\
       
       & MEAT & 35.24 & 1.87 & 58.32 & 84.47 & 79.28 & 0.79 & 66.85 & 36.85 & 57.57 & 77.28 & 35.81 & 48.18 & 48.54 \\
      & +\textsc{ORACLE} & \textbf{37.13} & 1.87 & \textbf{60.09} & \textbf{85.46} & \textbf{80.35} & \textbf{0.87} & \textbf{68.79} & \textbf{38.02} & \textbf{59.00} & \textbf{78.80} & \textbf{36.92} & \textbf{48.96} & \fcolorbox{white}{light yellow}{\textbf{49.69}} \\ \midrule

      \multirow{4}{*}{LaBSE} & DREAM & 98.88 & 14.14 & 96.20 & 97.85 & 93.42 & 10.87 & 96.79 & 91.90 & 96.50 & 97.79 & 92.49 & 92.52 & 81.61 \\
       & +\textsc{ORACLE} & 98.87 & \textbf{15.94} & \textbf{96.22} & 97.84 & \textbf{93.44} & \textbf{11.97} & \textbf{96.86} & \textbf{92.52} & 96.46 & \textbf{97.80} & \textbf{92.66} & \textbf{93.45} & \fcolorbox{white}{light yellow}{\textbf{82.00}} \\
       
       & MEAT & 98.75 & 17.97 & 96.14 & 97.83 & 93.38 & 13.85 & 96.71 & 92.42 & 96.53 & 97.77 & 91.32 & 93.30 & 82.16 \\
      & +\textsc{ORACLE} & \textbf{99.08} & 17.18 & \textbf{96.29} & \textbf{97.87} & \textbf{93.41} & 12.96 & \textbf{96.86} & \textbf{93.17} & \textbf{96.54} & \textbf{97.79} & \textbf{92.95} & \textbf{93.99} & \fcolorbox{white}{light yellow}{\textbf{82.34}} \\ \midrule\midrule

      \multicolumn{15}{c}{\textit{Language Embedding} (↓)} \\ \midrule

      \multirow{4}{*}{LASER} & DREAM & 1.58 & 0.24 & 1.66 & 6.76 & 4.02 & 0.22 & 1.74 & 0.45 & 0.82 & 5.78 & 1.87 & 0.50 & 2.14 \\
       & +\textsc{ORACLE} & \textbf{0.53} & \textbf{0.09} & \textbf{0.08} & \textbf{0.37} & \textbf{0.07} & \textbf{0.04} & \textbf{0.07} & \textbf{0.14} & \textbf{0.04} & \textbf{0.25} & \textbf{0.48} & \textbf{0.15} & \fcolorbox{white}{light yellow}{\textbf{0.19}} \\
       & MEAT & 7.13 & 0.96 & 11.83 & 27.65 & 16.07 & 0.38 & 11.20 & 3.67 & 8.61 & 25.16 & 9.04 & 3.95 & 10.47 \\
      & +\textsc{ORACLE} & \textbf{0.76} & \textbf{0.11} & \textbf{0.21} & \textbf{1.26} & \textbf{0.25} & \textbf{0.04} & \textbf{0.18} & \textbf{0.22} & \textbf{0.12} & \textbf{0.86} & \textbf{1.04} & \textbf{0.40} & \fcolorbox{white}{light yellow}{\textbf{0.45}} \\ \midrule

       \multirow{4}{*}{InfoXLM} & DREAM & 0.22 & 0.07 & 0.00 & 0.02 & 0.01 & 0.08 & 0.05 & 0.09 & 0.03 & 0.01 & 0.31 & 0.11 & 0.08 \\
       & +\textsc{ORACLE} & \textbf{0.03} & \textbf{0.01} & 0.01 & \textbf{0.01} & \textbf{0.00} & \textbf{0.02} & 0.05 & \textbf{0.03} & \textbf{0.02} & 0.01 & \textbf{0.08} & \textbf{0.03} & \fcolorbox{white}{light yellow}{\textbf{0.03}} \\
       & MEAT & 1.30 & 1.05 & 0.11 & 0.18 & 0.14 & 0.26 & 0.20 & 0.55 & 0.15 & 0.17 & 2.10 & 0.36 & 0.55 \\
      & +\textsc{ORACLE} & \textbf{0.28} & \textbf{0.30} & \textbf{0.02} & \textbf{0.00} & \textbf{0.00} & \textbf{0.07} & \textbf{0.04} & \textbf{0.14} & \textbf{0.03} & \textbf{0.01} & \textbf{0.47} & \textbf{0.07} & \fcolorbox{white}{light yellow}{\textbf{0.12}} \\ \midrule

      \multirow{4}{*}{LaBSE} & DREAM & 7.42 & 0.66 & 1.12 & 2.73 & 1.68 & 0.69 & 2.35 & 4.65 & 1.54 & 1.21 & 3.15 & 4.59 & 2.65 \\
       & +\textsc{ORACLE} & \textbf{0.85} & \textbf{0.13} & \textbf{0.04} & \textbf{0.03} & \textbf{0.03} & \textbf{0.08} & \textbf{0.13} & \textbf{0.40} & \textbf{0.10} & \textbf{0.03} & \textbf{0.52} & \textbf{0.26} & \fcolorbox{white}{light yellow}{\textbf{0.22}} \\
       & MEAT & 60.54 & 6.84 & 18.34 & 28.36 & 21.33 & 6.27 & 23.46 & 54.32 & 20.33 & 20.38 & 44.17 & 60.28 & 30.39 \\
      & +\textsc{ORACLE} & \textbf{1.98} & \textbf{0.50} & \textbf{0.18} & \textbf{0.22} & \textbf{0.22} & \textbf{0.38} & \textbf{0.31} & \textbf{1.59} & \textbf{0.35} & \textbf{0.16} & \textbf{2.36} & \textbf{1.14} & \fcolorbox{white}{light yellow}{\textbf{0.78}} \\
      
    \bottomrule
    \end{tabular}
}
\caption{Cross-lingual sentence retrieval accuracy with our test set, comprising 0.5M pairs for each language. We expect the semantic retrieval accuracy to be higher and lower with language embedding. \textbf{Bold} represents when our method surpass the vanilla approach and \fcolorbox{white}{light yellow}{\textbf{highlight}} denotes when the average value is higher. vanilla: original DREAM or MEAT approach; \textsc{ORACLE}: incorporation of our objective. *: We use mean pooling to compute sentence embedding. All average improvements are statistically significant with $p$-value $\leq 0.001$.}
\label{tab:test_set}
\end{table*}

% For semantic embedding results, we test the statistically significance of average improvements over original DREAM or MEAT and $\dagger$ marks results with $p$-value $\leq 0.05$. 
\clearpage

\definecolor{light orange}{rgb}{0.988, 0.890, 0.761}

\begin{table*}[!h]
\centering
\resizebox{\textwidth}{!}{%
    \begin{tabular}{l l c c c c c c c c c c c c c c c c c}
    \toprule
    \textbf{Encoder} & \textbf{Objective} & \textbf{en-ar} & \textbf{en-de} & \textbf{en-es} & \textbf{en-fr} & \textbf{en-it} & \textbf{en-ja} & \textbf{en-nl} & \textbf{en-pt} & \textbf{en-ro} & \textbf{en-zh} & \textbf{Avg.} \\ \midrule\midrule

    \multicolumn{13}{c}{\textit{Original Embedding}} \\ \midrule
    
     MUSE \ding{118} & - & - & - & 95.40 & 93.50 & 94.30 & 93.80 & 94.00 & 94.90 & 30.00 & 94.30 & 86.90\\
     CRISS \ding{170} & - & - & - & 96.30 & 92.70 & 92.50 & 84.80 & 93.40 & - & - & 85.60 & 90.20\\
     DuEAM \ding{171} & - & - & - & 93.00 & 91.50 & 85.70 & 84.20 & - & 91.20 & 88.50 & 90.20 & 87.90\\
     % mBERT+DAP \ding{169} & - & 90.6 & 98.9 & 98.1 & 95.4 & 95.8 & 96.6 & 97.2 & 95.6 & - & 95.3 & 95.9\\
     % XLM-R+DAP \ding{169} & - & 92.2 & 98.9 & 98.6 & 96.3 & 96.1 & 97.3 & 97.2 & 95.6 & - & 95.3 & 96.4 \\ 
    LASER & - & 91.95 & 99.05 & 98.00 & 95.65 & 95.30 & 95.35 & 96.30 & 95.15 & 97.40 & 95.45 & 95.96\\
     InfoXLM$^*$ & - & 20.95 & 38.50 & 30.85 & 32.35 & 24.85 & 28.20 & 19.85 & 36.90 & 30.40 & 34.05 & 29.69 \\
    LaBSE & - & 89.75 & 99.20 & 98.10 & 96.05 & 94.75 & 96.40 & 96.90 & 95.55 & 97.40 & 96.20 & 96.03\\ \midrule\midrule

     \multicolumn{13}{c}{\textit{Semantic Embedding} (↑)} \\ \midrule

     \multirow{4}{*}{LASER} & DREAM & 60.35 & 89.85 & 83.40 & 76.55 & 80.95 & 71.70 & 80.10 & 82.15 & 80.60 & 74.20 & 77.99 \\
       & +\textsc{ORACLE} & 60.30 & \textbf{90.00} & 83.40 & \textbf{76.65} & \textbf{81.00} & \textbf{72.05} & \textbf{80.35} & \textbf{82.30} & 80.60 & \textbf{74.60} & \fcolorbox{white}{light orange}{\textbf{78.13}} \\
       & MEAT & 86.95 & 96.55 & 96.00 & 91.35 & 91.80 & 90.65 & 91.75 & 93.45 & 94.80 & 92.95 & 92.63 \\
      & +\textsc{ORACLE} & \textbf{87.30} & \textbf{98.05} & \textbf{96.65} & \textbf{92.75} & \textbf{92.55} & 87.95 & \textbf{93.70} & \textbf{94.25} & \textbf{95.40} & \textbf{93.80} & \fcolorbox{white}{light orange}{\textbf{93.24}} \\ \midrule

      \multirow{4}{*}{InfoXLM} & DREAM & 44.05 & 57.65 & 68.65 & 62.80 & 54.80 & 56.45 & 62.70 & 66.50 & 58.15 & 67.10 & 59.89\\
       & +\textsc{ORACLE} & \textbf{44.80} & \textbf{58.00} & \textbf{68.85} & 62.65 & 54.80 & \textbf{57.05} & 62.30 & \textbf{66.65} & 57.95 & \textbf{67.20} & \fcolorbox{white}{light orange}{\textbf{60.03}} \\
       & MEAT & 31.60 & 67.25 & 70.75 & 67.30 & 63.70 & 42.05 & 68.95 & 74.05 & 59.05 & 57.70 & 60.24 \\
      &  +\textsc{ORACLE} & \textbf{31.80} & \textbf{69.00} & \textbf{71.60} & \textbf{68.35} & \textbf{64.15} & \textbf{43.10} & \textbf{70.85} & \textbf{75.25} & \textbf{60.45} & \textbf{60.30} & \fcolorbox{white}{light orange}{\textbf{61.49}} \\ \midrule

       \multirow{4}{*}{LaBSE} & DREAM & 89.90 & 99.10 & 98.50 & 95.80 & 95.05 & 95.75 & 97.35 & 95.45 & 97.50 & 95.35 & 95.98\\
       & +\textsc{ORACLE} & 89.70 & \textbf{99.15} & 98.50 & \textbf{95.90} & 94.85 & \textbf{95.90} & 97.30 & \textbf{95.50} & \textbf{97.70} & \textbf{95.55} & \fcolorbox{white}{light orange}{\textbf{96.01}}\\
      & MEAT & 90.30 & 99.20 & 98.15 & 98.90 & 94.55 & 96.15 & 97.30 & 95.55 & 97.55 & 95.50 & 96.32 \\
      &  +\textsc{ORACLE} & \textbf{90.95} & \textbf{99.40} & \textbf{98.50} & \textbf{96.30} & \textbf{95.20} & \textbf{96.40} & \textbf{97.40} & \textbf{95.75} & \textbf{97.85} & \textbf{95.80} & \fcolorbox{white}{light orange}{\textbf{96.36}} \\ \midrule\midrule

      \multicolumn{13}{c}{\textit{Language Embedding} (↓)} \\ \midrule

      \multirow{4}{*}{LASER} & DREAM & 1.20 & 1.50 & 2.95 & 1.50 & 4.45 & 1.15 & 2.00 & 3.70 & 1.85 & 1.70 & 2.20\\
       & +\textsc{ORACLE} & \textbf{0.25} & \textbf{0.10} & \textbf{0.30} & \textbf{0.20} & \textbf{0.25} & \textbf{0.10} & \textbf{0.35} & \textbf{0.30} & \textbf{0.20} & \textbf{0.05} & \fcolorbox{white}{light orange}{\textbf{0.21}}\\
       & MEAT & 19.40 & 9.75 & 13.55 & 6.70 & 14.30 & 5.65 & 9.90 & 16.15 & 16.20 & 10.85 & 12.25\\
      & +\textsc{ORACLE} & \textbf{0.60} & \textbf{0.20} & \textbf{0.45} & \textbf{0.30} & \textbf{0.55} & \textbf{0.30} & \textbf{0.65} & \textbf{0.40} & \textbf{0.75} & \textbf{0.45} & \fcolorbox{white}{light orange}{\textbf{0.47}}\\ \midrule

      \multirow{4}{*}{InfoXLM} & DREAM & 0.10 & 0.10 & 0.25 & 0.15 & 0.45 & 0.20 & 0.40 & 0.20 & 0.15 & 0.20 & 0.22\\
       & +\textsc{ORACLE} & 0.10 & 0.10 & \textbf{0.10} & \textbf{0.10} & \textbf{0.10} & \textbf{0.10} & \textbf{0.10} & \textbf{0.10} & \textbf{0.10} & \textbf{0.10} & \fcolorbox{white}{light orange}{\textbf{0.10}}\\
       & MEAT & 0.35 & 0.55 & 2.00 & 0.90 & 2.75 & 0.30 & 3.35 & 0.85 & 1.15 & 0.40 & 1.26\\
      & +\textsc{ORACLE} & \textbf{0.10} & \textbf{0.15} & \textbf{0.25} & \textbf{0.15} & \textbf{0.10} & \textbf{0.10} & \textbf{0.20} & \textbf{0.20} & \textbf{0.15} & \textbf{0.15} & \fcolorbox{white}{light orange}{\textbf{0.16}}\\ \midrule
      
       \multirow{4}{*}{LaBSE} & DREAM & 24.50 & 11.50 & 18.95 & 17.85 & 24.25 & 11.20 & 14.20 & 9.70 & 12.35 & 24.70 & 16.92\\
       & +\textsc{ORACLE} & \textbf{2.15} & \textbf{1.20} & \textbf{1.00} & \textbf{0.30} & \textbf{1.45} & \textbf{1.15} & \textbf{1.30} & \textbf{9.30} & \textbf{0.70} & \textbf{1.00} & \fcolorbox{white}{light orange}{\textbf{1.96}}\\
       & MEAT & 64.25 & 55.30 & 64.10 & 64.30 & 64.50 & 65.45 & 60.25 & 58.15 & 57.60 & 60.45 & 61.44\\
      & +\textsc{ORACLE} & \textbf{8.30} & \textbf{7.25} & \textbf{5.05} & \textbf{5.00} & \textbf{8.15} & \textbf{9.70} & \textbf{7.20} & \textbf{3.90} & \textbf{6.75} & \textbf{5.65} & \fcolorbox{white}{light orange}{\textbf{6.70}}\\
    \bottomrule
    \end{tabular}
}
\caption{Cross-lingual retrieval accuracy with Tatoeba task. For each language pair, we report the average accuracy of both directions (from English and into English). \textbf{Bold} represents when our method surpass the vanilla approach and \fcolorbox{white}{light orange}{\textbf{highlight}} denotes when the average value is higher. \ding{118}: results from \citet{lee-chen-2017-muse} (\textit{supervised}) ; \ding{170}: results from \citet{tran2020crosslingual} (\textit{weakly supervised}); \ding{171}: results from \citet{goswami-etal-2021-cross} (\textit{self-supervised}). *: We use mean pooling to compute sentence embedding. All average improvements are statistically significant with $p$-value $\leq 0.001$.}
\label{tab:tatoeba_detailed}
\end{table*}

% For semantic embedding results, we test the statistically significance of improvements over original DREAM or MEAT and $\dagger$ marks results with $p$-value $\leq 0.05$.}
\clearpage

\definecolor{light blue}{rgb}{0.678,0.847,0.902}
\definecolor{light green}{rgb}{0.906, 0.996, 0.753}
\definecolor{light purple}{rgb}{0.871, 0.753, 0.996}
\definecolor{light aqua}{rgb}{0.753, 0.996, 0.961}
\definecolor{light peach}{rgb}{0.996, 0.918, 0.753}
\definecolor{light yellow}{rgb}{0.984, 0.996, 0.753}
\definecolor{light lavendar}{rgb}{0.867, 0.835, 1.0}

\begin{table*}[!h]
\centering
\resizebox{\textwidth}{!}{%
    \begin{tabular}{l l c c c c c c c c c c c}
    \toprule
    \textbf{Encoder} & \textbf{Objective} & \fcolorbox{white}{pink}{\textbf{ar-ar}} & \fcolorbox{white}{pink}{\textbf{en-en}} & \fcolorbox{white}{pink}{\textbf{es-es}} & \fcolorbox{white}{light blue}{\textbf{en-ar}} & \fcolorbox{white}{light blue}{\textbf{en-de}} & \fcolorbox{white}{light blue}{\textbf{en-tr}} & \fcolorbox{white}{light blue}{\textbf{en-es}} & \fcolorbox{white}{light blue}{\textbf{en-fr}} & \fcolorbox{white}{light blue}{\textbf{en-it}} & \fcolorbox{white}{light blue}{\textbf{en-nl}} & \textbf{Avg.}\\ \midrule\midrule

    \multicolumn{13}{c}{\textit{Original Embedding}} \\ \midrule
     
     LASER & - & 68.85 & 66.55 & 57.93 & 77.62 & 79.68 & 64.20 & 71.98 & 69.05 & 70.83 & 68.68 & 69.54 \\
     InfoXLM$^*$ & - & 19.11 & 50.20 & 36.17 & 12.89 & 16.31 & 24.86 & 9.10 & 25.12 & 28.10 & 30.55 & 25.24 \\
    mSimCSE & - & 69.06 & 74.50 & 65.71 & 79.45 & 80.83 & 73.85 & 72.07 & 76.98 & 76.98 & 75.22 & 74.47 \\
     \midrule\midrule

     \multicolumn{13}{c}{\textit{Semantic Embedding} (↑)} \\ \midrule

     \multirow{4}{*}{LASER} & DREAM & 57.10 & 53.98 & 46.36 & 43.22 & 41.92 & 40.60 & 32.88 & 48.58 & 49.94 & 47.47 & 46.21 \\
       & +\textsc{ORACLE} & \textbf{57.16} & \textbf{54.14} & \textbf{46.64} & \textbf{43.55} & \textbf{42.11} & \textbf{40.67} & 32.83 & \textbf{48.70} & \textbf{49.98} & \textbf{47.80} & \fcolorbox{white}{light orange}{\textbf{46.36}} \\
       
       & MEAT & 66.87 & 71.95 & 79.16 & 62.41 & 60.44 & 65.06 & 54.63 & 61.94 & 66.27 & 63.90 & 65.26 \\
      &  +\textsc{ORACLE} & \textbf{67.14} & \textbf{72.69} & 78.75 & \textbf{63.59} & 60.03 & \textbf{66.19} & \textbf{55.20} & 61.38 & 65.80 & 63.76 & \fcolorbox{white}{light orange}{\textbf{65.45}} \\ \midrule

        \multirow{4}{*}{InfoXLM} & DREAM & 50.28 & 56.39 & 56.16 & 43.35 & 39.54 & 42.71 & 38.42 & 48.02 & 47.80 & 50.18 & 47.29 \\
       & +\textsc{ORACLE} & 50.25 & 56.38 & 56.16 & 43.35 & \textbf{49.55} & 42.61 & 38.40 & 47.98 & \textbf{47.82} & \textbf{50.19} & \fcolorbox{white}{light orange}{\textbf{48.27}} \\
       
       & MEAT & 35.83 & 61.23 & 51.14 & 11.09 & 25.58 & 33.32 & 20.04 & 29.38 & 41.58 & 37.50 & 34.67 \\
      &  +\textsc{ORACLE} & \textbf{35.87} & \textbf{61.40} & 50.73 & \textbf{11.26} & \textbf{28.15} & \textbf{34.83} & \textbf{21.41} & \textbf{31.55} & \textbf{42.72} & \textbf{38.77} & \fcolorbox{white}{light orange}{\textbf{35.67}} \\ \midrule
      
       \multirow{4}{*}{LaBSE} & DREAM & 69.84 & 74.78 & 79.82 & 70.97 & 70.82 & 71.30 & 64.22 & 75.67 & 76.28 & 75.56 & 72.93 \\
       & +\textsc{ORACLE} & \textbf{70.65} & \textbf{76.03} & \textbf{81.06} & \textbf{72.37} & \textbf{72.49} & \textbf{73.33} & \textbf{66.18} & \textbf{76.13} & \textbf{76.76} & \textbf{76.29} & \fcolorbox{white}{light orange}{\textbf{74.13}} \\
      
      & MEAT & 72.03 & 80.34 & 83.66 & 74.71 & 75.40 & 73.59 & 70.48 & 77.82 & 78.18 & 77.43 & 76.36 \\
      &  +\textsc{ORACLE} & \textbf{72.05} & \textbf{80.41} & \textbf{83.86} & \textbf{75.09} & \textbf{75.67} & \textbf{74.56} & \textbf{70.98} & 77.75 & \textbf{78.57} & \textbf{77.44} & \fcolorbox{white}{light orange}{\textbf{76.64}} \\ \midrule\midrule

      \multicolumn{13}{c}{\textit{Language Embedding} (↓)} \\ \midrule

      \multirow{4}{*}{LASER} & DREAM & 45.12 & 34.87 & 39.95 & 18.94 & 11.89 & 21.50 & 17.64 & 8.29 & 14.91 & 10.81 & 22.39 \\
       & +\textsc{ORACLE} & \textbf{21.43} & \textbf{12.77} & \textbf{18.40} & 20.85 & \textbf{11.65} & \textbf{17.12} & \textbf{15.30} & \textbf{6.14} & 16.87 & 11.51 & \fcolorbox{white}{light orange}{\textbf{15.20}} \\
       & MEAT & 52.51 & 40.35 & 55.29 & 35.78 & 22.93 & 27.26 & 27.89 & 22.73 & 30.74 & 25.99 & 34.15 \\
      &  +\textsc{ORACLE} & \textbf{21.26} & \textbf{14.97} & \textbf{21.75} & \textbf{21.48} & \textbf{6.09} & \textbf{15.71} & \textbf{15.51} & \textbf{4.86} & \textbf{15.37} & \textbf{8.49} & \fcolorbox{white}{light orange}{\textbf{14.55}} \\ \midrule

      \multirow{4}{*}{InfoXLM} & DREAM & 39.62 & 51.03 & 49.74 & 3.87 & 2.66 & 4.53 & 12.95 & 9.94 & 9.37 & 12.21 & 19.59 \\
       & +\textsc{ORACLE} & \textbf{24.62} & \textbf{31.07} & \textbf{36.01} & \textbf{-10.84} & \textbf{-7.12} & \textbf{-9.51} & \textbf{2.85} & \textbf{-3.18} & \textbf{-2.06} & \textbf{2.67} & \fcolorbox{white}{light orange}{\textbf{6.45}} \\
       & MEAT & 33.00 & 50.01 & 51.02 & -6.10 & 8.21 & -2.50 & -1.97 & 8.44 & 13.58 & 10.87 & 16.46 \\
      &  +\textsc{ORACLE} & 33.13 & \textbf{46.96} & 51.63 & \textbf{-11.63} & \textbf{0.14} & \textbf{-8.14} & \textbf{-7.91} & \textbf{1.57} & \textbf{7.25} & \textbf{3.59} & \fcolorbox{white}{light orange}{\textbf{11.66}} \\ \midrule

    \multirow{4}{*}{LaBSE} & DREAM & 44.32 & 40.35 & 50.81 & 24.12 & 22.56 & 28.29 & 18.76 & 22.86 & 20.38 & 22.02 & 29.45 \\
       & +\textsc{ORACLE} & \textbf{33.10} & \textbf{19.88} & \textbf{28.60} & \textbf{1.59} & \textbf{1.14} & \textbf{17.39} & \textbf{15.57} & \textbf{12.17} & \textbf{8.36} & \textbf{12.53} & \fcolorbox{white}{light orange}{\textbf{15.03}} \\
       & MEAT & 52.11 & 68.57 & 68.18 & 38.57 & 28.94 & 35.87 & 31.27 & 28.66 & 29.40 & 27.21 & 40.88 \\
      &  +\textsc{ORACLE} & \textbf{37.25} & \textbf{27.30} & \textbf{33.61} & \textbf{0.08} & \textbf{0.79} & \textbf{16.94} & \textbf{16.60} & \textbf{10.98} & \textbf{8.01} & \textbf{13.12} & \fcolorbox{white}{light orange}{\textbf{16.47}} \\
    \bottomrule
    \end{tabular}
}
\caption{Spearman's rank correlation coefficients ($\rho$) of \fcolorbox{white}{pink}{monolingual} and \fcolorbox{white}{light blue}{cross-lingual} STS task. \textbf{Bold} represents when our method surpass the vanilla approach and \fcolorbox{white}{light orange}{\textbf{highlight}} indicates when the average value is higher. *: We use mean pooling to compute sentence embedding. All average improvements are statistically significant with $p$-value $\leq 0.001$.}
\label{tab:sts}
\end{table*}

% For semantic embedding results, we test the statistically significance of improvements over original DREAM or MEAT and $\dagger$ marks results with $p$-value $\leq 0.05$.
\clearpage

\definecolor{light yellow}{rgb}{0.984, 0.996, 0.753}
\definecolor{light pink}{rgb}{0.9609375, 0.87109375, 0.93359375}

\begin{table*}[!h]
\centering
\resizebox{\textwidth}{!}{%
    \begin{tabular}{l l c c c c c c c c c}
    \toprule
    \textbf{Encoder} & \textbf{Objective} & \textbf{en-ar} & \textbf{en-de} & \textbf{en-es} & \textbf{en-fr} & \textbf{en-it} & \textbf{en-nl} & \textbf{en-pt} & \textbf{en-ro} & \textbf{Avg.} \\ \midrule\midrule

    \multicolumn{11}{c}{\textit{Original Embedding}} \\ \midrule
    
     LASER & - & 90.82 & 98.75 & 98.26 & 95.17 & 94.63 & 95.22 & 95.91 & 98.37 & 95.89 \\ 
     InfoXLM$^*$ & - & 13.71 & 39.04 & 37.05 & 36.20 & 29.44 & 29.54 & 41.44 & 31.37 & 32.22 \\ 
     LaBSE & - & 90.06 & 99.48 & 98.49 & 95.60 & 93.46 & 97.02 & 96.56 & 98.24 & 96.11 \\ \midrule\midrule

     \multicolumn{11}{c}{\textit{Semantic Embedding} (↑)} \\ \midrule

      \multirow{4}{*}{LASER} & DREAM & 59.75 & 88.58 & 84.67 & 73.90 & 84.46 & 80.66 & 82.02 & 82.43 & 79.56 \\
       & +\textsc{ORACLE} & 59.62 & \textbf{88.79} & \textbf{85.02} & 73.90 & 84.46 & 80.45 & \textbf{82.24} & 82.18 & \fcolorbox{white}{light pink}{\textbf{79.58}} \\
       
       & MEAT & 84.91 & 97.82 & 96.28 & 92.70 & 93.34 & 93.84 & 95.05 & 96.86 & 93.85 \\
      & +\textsc{ORACLE} & \textbf{86.54} & 97.09 & \textbf{97.79} & 92.59 & 93.22 & 92.67 & \textbf{95.26} & 96.11 & \fcolorbox{white}{light pink}{\textbf{93.91}} \\ \midrule
      
    \multirow{4}{*}{InfoXLM} & DREAM & 28.05 & 55.14 & 58.07 & 59.72 & 54.09 & 53.99 & 54.90 & 56.46 & 52.55 \\
       & +\textsc{ORACLE} & 27.80 & \textbf{56.39} & \textbf{58.19} & \textbf{60.15} & \textbf{54.67} & \textbf{54.84} & \textbf{56.62} & 56.34 & \fcolorbox{white}{light pink}{\textbf{53.13}} \\
       
       & MEAT & 15.35 & 42.26 & 60.51 & 58.22 & 56.43 & 56.43 & 57.70 & 50.69 & 49.70 \\
       & +\textsc{ORACLE} & \textbf{16.35} & 42.26 & \textbf{61.32} & \textbf{59.94} & \textbf{58.06} & \textbf{57.17} & \textbf{59.31} & \textbf{52.07} & \fcolorbox{white}{light pink}{\textbf{50.81}} \\ \midrule

      \multirow{4}{*}{LaBSE} & DREAM & 87.78 & 99.27 & 97.33 & 95.38 & 92.87 & 96.60 & 96.45 & 97.99 & 95.46 \\
       & +\textsc{ORACLE} & \textbf{88.30} & 99.27 & \textbf{98.14} & 95.38 & \textbf{93.22} & \textbf{96.81} & \textbf{96.66} & \textbf{98.11} & \fcolorbox{white}{light pink}{\textbf{95.74}} \\
       
       & MEAT & 88.43 & 99.38 & 98.03 & 94.95 & 92.87 & 96.49 & 96.12 & 97.74 & 95.50 \\
      & +\textsc{ORACLE} & \textbf{89.56} & \textbf{99.69} & \textbf{98.37} & \textbf{96.60} & \textbf{93.34} & \textbf{97.13} & \textbf{96.34} & \textbf{98.24} & \fcolorbox{white}{light pink}{\textbf{96.16}} \\ \midrule\midrule

      \multicolumn{11}{c}{\textit{Language Embedding} (↓)} \\ \midrule

      \multirow{4}{*}{LASER} & DREAM & 2.01 & 4.15 & 7.08 & 3.01 & 7.36 & 6.06 & 9.36 & 3.76 & 5.35 \\
       & +\textsc{ORACLE} & \textbf{0.25} & \textbf{0.83} & \textbf{1.39} & \textbf{0.43} & \textbf{0.93} & \textbf{0.85} & \textbf{2.48} & \textbf{0.75} & \fcolorbox{white}{light pink}{\textbf{0.99}} \\
       
       & MEAT & 35.72 & 23.88 & 35.31 & 17.72 & 27.57 & 31.77 & 49.62 & 22.84 & 30.55 \\
      & +\textsc{ORACLE} & \textbf{1.51} & \textbf{1.66} & \textbf{3.02} & \textbf{1.07} & \textbf{2.45} & \textbf{2.34} & \textbf{5.06} & \textbf{1.25} & \fcolorbox{white}{light pink}{\textbf{2.30}} \\ \midrule

       \multirow{4}{*}{InfoXLM} & DREAM & 0.13 & 0.52 & 0.93 & 0.64 & 0.93 & 2.34 & 0.54 & 0.53 & 0.82 \\
       & +\textsc{ORACLE} & 0.13 & \textbf{0.10} & \textbf{0.46} & \textbf{0.11} & \textbf{0.23} & \textbf{0.85} & \textbf{0.11} & \textbf{0.13} & \fcolorbox{white}{light pink}{\textbf{0.27}} \\
       
       & MEAT & 1.89 & 10.38 & 20.44 & 13.64 & 20.56 & 36.03 & 23.90 & 15.93 & 17.85 \\
      & +\textsc{ORACLE} & \textbf{0.38} & \textbf{1.04} & \textbf{3.37} & \textbf{1.40} & \textbf{3.15} & \textbf{8.93} & \textbf{3.12} & \textbf{1.76} & \textbf{2.89} \\ \midrule

      \multirow{4}{*}{LaBSE} & DREAM & 11.57 & 14.54 & 21.24 & 23.52 & 27.69 & 27.21 & 19.38 & 17.44 & 20.32 \\
       & +\textsc{ORACLE} & \textbf{1.26} & \textbf{1.25} & \textbf{1.39} & \textbf{2.69} & \textbf{1.99} & \textbf{2.98} & \textbf{0.75} & \textbf{0.88} & \fcolorbox{white}{light pink}{\textbf{1.65}} \\
       & MEAT & 48.81 & 37.80 & 53.31 & 51.34 & 56.54 & 53.35 & 43.27 & 34.00 & 47.30 \\
      & +\textsc{ORACLE} & \textbf{6.42} & \textbf{7.06} & \textbf{6.04} & \textbf{7.63} & \textbf{9.11} & \textbf{10.52} & \textbf{6.14} & \textbf{6.40} & \fcolorbox{white}{light pink}{\textbf{7.42}} \\
      
    \bottomrule
    \end{tabular}
}
\caption{Retrieval accuracy with our code-switching dataset. \textbf{Bold} represents when our method surpasses the vanilla approach and \fcolorbox{white}{light pink}{\textbf{highlight}} denotes when the average value is higher. vanilla: original DREAM or MEAT approach; \textsc{ORACLE}: incorporation of our objective. *: We use mean pooling to compute sentence embedding. All average improvements are statistically significant with $p$-value $\leq 0.001$.}
\label{tab:code-switch-results}
\end{table*}

% For semantic embedding results, we test the statistically significance of average improvements over original DREAM or MEAT and $\dagger$ marks results with $p$-value $\leq 0.05$. 
\clearpage

\clearpage

\end{document}